%% file: iclr2027_conference.tex
\pdfoutput=1
\documentclass{article}
\usepackage{iclr2027_conference,times}

\input{math_commands.tex}

\usepackage{hyperref}
\usepackage{url}
\usepackage{amsmath}
\usepackage{amssymb}
\usepackage{algorithm}
\usepackage{algpseudocode}
\usepackage{booktabs}
\usepackage{multirow}
\usepackage{array}
\usepackage{caption}
\usepackage{subcaption}
\usepackage{amsthm}
\usepackage{graphicx}
\usepackage{tikz}
\usetikzlibrary{positioning,calc,arrows.meta,fit,backgrounds,decorations.pathmorphing,shapes.geometric}
\graphicspath{{figures/}}
\newtheorem{definition}{Definition}

\newtheorem{proposition}{Proposition}

\newtheorem{lemma}{Lemma}[section]
\newtheorem{corollary}[lemma]{Corollary}
\newtheorem{aproposition}[lemma]{Proposition}
\theoremstyle{remark}
\newtheorem*{remark}{Remark}
\theoremstyle{plain}
\newcommand{\TV}{d_{\mathrm{TV}}}

\newcommand{\vpi}{\boldsymbol{\pi}}

\newcommand{\vell}{\boldsymbol{\ell}}
\newcommand{\vtau}{\boldsymbol{\tau}}
\newcommand{\vgamma}{\boldsymbol{\gamma}}
\newcommand{\vphi}{\boldsymbol{\phi}}
\newcommand{\vomega}{\boldsymbol{\omega}}

\input{figures/dig2_palette}

\input{figures/assets/method_data}

\input{figures/assets/fig2_prior_data}

\input{figures/assets/fig2_traj_data}

\makeatletter
\newif\ifalgdiff
\newcommand\algdiffnext{\global\algdifftrue}
\algrenewcommand\alglinenumber[1]{%
  \ifalgdiff\global\algdifffalse
    \edef\algdiff@ht{\the\dimexpr\baselineskip\relax}\edef\algdiff@dp{\the\dp\strutbox}%
    \footnotesize #1:%
    \rlap{\hspace*{\labelsep}\textcolor{black!15}{\rule[-\dimexpr\algdiff@dp+1pt\relax]{\linewidth}{\dimexpr\algdiff@ht+2pt\relax}}}%
  \else\footnotesize #1:\fi}
\makeatother

\definecolor{revblue}{HTML}{0030C0}

\newcommand{\CE}{\mathrm{CE}}
\newcommand{\Fro}{F_{\vtheta}}
\def\Tabref#1{Table~\ref{#1}}

\newcommand{\coderepo}{https://github.com/revolins/DiG2}

\title{Multi-Domain Discrete Graph Generation}

\author{%
  Jay Revolinsky$^{1}$,\;
  Harry Shomer$^{2}$,\;
  Haoyu Han$^{1}$,\;
  Jiliang Tang$^{1}$ \\
  $^{1}$Michigan State University, East Lansing, MI, USA \\
  $^{2}$University of Texas - Arlington, Arlington, TX, USA \\
  \texttt{revolins@msu.edu}
}

\iclrfinalcopy 
\begin{document}

\maketitle
\lhead{Preprint. Under review.}

\begin{abstract}
Current graph generative models are overwhelmingly trained on a single input distribution and then sampled to reproduce it. In this work, we propose extending graph generation into multiple structural domains. Under our proposed multi-domain setting, existing single-domain models, even when given a domain label, sample every domain from a single shared internal distribution. In preliminary benchmarking, we demonstrate the existing model paradigm fails, even when state-of-the-art models are trained with multi-domain awareness. To address this, we propose Discrete-interpolative Graph Generation or DiG$^{2}$, a graph flow-matching model specially-designed to internalize domain-conditioning along a bridged discrete path. Benchmarking demonstrates that DiG$^{2}$ achieves at least $85\%$ sampling accuracy across all tested domains, while current domain-conditioned models suffer mode-collapse. Additionally, of the domain-conditioned generators we test, only DiG$^{2}$'s embedding blend moves molecular properties monotonically from QM9 to ZINC at fixed size and extends past the ZINC row. This interpolation mechanism uncovers targeted molecule generation for achieving comparable performance relative to model $38$ times larger. Source Code: \href{\coderepo}{Available Here}
\end{abstract}

\input{sections/introduction}
\input{sections/background}

\input{sections/formulation}
\input{sections/benchmark}
\input{sections/ablation}

\input{sections/scaling}
\input{sections/interpolation}
\input{sections/conclusion}

\subsection*{AI use statement}
In this work, we used generative AI tools to assist in the writing of proofs. We have not used generative AI tools for: idea formulation, help developing theoretical models or conceptual frameworks, formulating mathematical claims, providing critical ingredients for proving mathematical claims, proposing or refining hypotheses, designing feedback on research methodology or experiments, implementing methods, assisting with translation, cleaning and reformatting data, supporting qualitative and thematic data analysis, and interpreting results. Using an LLM to generate synthetic data sets is not applicable to this work. Additionally, we used generative AI tools for: drafting parts of this research paper, editing software code, and modifying scientific figures. We have reviewed all AI-assisted work. LLM-generated code was verified and tested for correctness by the first author, mathematical proofs were refined by 2 authors and checked by all authors. We take responsibility for the final content of this work, including text, claims or artifacts produced with the aid of generative AI.

\subsection*{Ethics statement}
Given the nature of graph generation research, we acknowledge that advancing this subfield could have unintended consequences for generative models to be applied for dual-use.

\subsection*{Reproducibility statement}
We have included the following in order to promote the reproducibility of our work: implementation details indicating exact hyperparameters (Appendix~\ref{app:impl}), \href{\coderepo}{source code here} and within the abstract, and deeper details on our theoretical proof (Appendix~\ref{app:proofs}).

\bibliography{iclr2027_conference}
\bibliographystyle{iclr2027_conference}

\appendix
\section{Appendix}
\input{sections/appendix}

\end{document}

%% file: math_commands.tex
\usepackage{amsmath,amsfonts,bm}

\def\Figref#1{Figure~\ref{#1}}

\def\Secref#1{Section~\ref{#1}}

\def\eqref#1{equation~\ref{#1}}
\def\Eqref#1{Equation~\ref{#1}}

\def\1{\bm{1}}

\def\rh{{\textnormal{h}}}

\def\rr{{\textnormal{r}}}

\def\rx{{\textnormal{x}}}

\def\vone{{\bm{1}}}
\def\vmu{{\bm{\mu}}}
\def\vtheta{{\bm{\theta}}}

\def\vc{{\bm{c}}}

\def\ve{{\bm{e}}}

\def\vm{{\bm{m}}}

\def\vp{{\bm{p}}}
\def\vq{{\bm{q}}}
\def\vr{{\bm{r}}}
\def\vs{{\bm{s}}}

\def\vw{{\bm{w}}}

\def\vy{{\bm{y}}}
\def\vz{{\bm{z}}}

\def\mA{{\bm{A}}}
\def\mB{{\bm{B}}}

\def\mE{{\bm{E}}}

\def\mI{{\bm{I}}}

\def\mK{{\bm{K}}}

\def\mQ{{\bm{Q}}}

\def\mV{{\bm{V}}}

\def\mX{{\bm{X}}}

\DeclareMathAlphabet{\mathsfit}{\encodingdefault}{\sfdefault}{m}{sl}
\SetMathAlphabet{\mathsfit}{bold}{\encodingdefault}{\sfdefault}{bx}{n}
\newcommand{\tens}[1]{\bm{\mathsfit{#1}}}

\def\tE{{\tens{E}}}

\def\gG{{\mathcal{G}}}

\def\sD{{\mathbb{D}}}

\def\sI{{\mathbb{I}}}

\def\sK{{\mathbb{K}}}

\def\sV{{\mathbb{V}}}

\newcommand{\E}{\mathbb{E}}

\newcommand{\R}{\mathbb{R}}

\newcommand{\softmax}{\mathrm{softmax}}

%% file: figures/dig2_palette.tex
\usepackage{xcolor}

\definecolor{digPlanar}{HTML}{0072B2}
\definecolor{digSBM}{HTML}{56B4E9}
\definecolor{digTree}{HTML}{009E73}
\definecolor{digQM9}{HTML}{E69F00}
\definecolor{digZINC}{HTML}{D55E00}

\definecolor{dig2cond}{HTML}{CC79A7}

\definecolor{digInk}{HTML}{333333}
\definecolor{digRule}{HTML}{8A8A8A}
\definecolor{digFill}{HTML}{E8E8E8}
\definecolor{digFillSoft}{HTML}{F4F4F4}
\definecolor{digFail}{HTML}{9A9A9A}

\tikzset{
  digfont/.style        = {font=\sffamily\scriptsize, text=digInk},
  digtiny/.style        = {font=\sffamily\tiny, text=digInk},
  digtinier/.style      = {font=\sffamily\fontsize{5}{6}\selectfont, text=digInk},
  digblock/.style       = {draw=digRule, line width=0.4pt, fill=digFillSoft,
                           rounded corners=1.4pt, inner sep=3pt, digtiny,
                           align=center},
  digcondblock/.style   = {digblock, draw=dig2cond, line width=0.6pt,
                           fill=dig2cond!8, text=digInk},
  digarrow/.style       = {-{Straight Barb[length=2.4pt,width=2.6pt]},
                           draw=digInk, line width=0.5pt},
  digcondarrow/.style   = {-{Straight Barb[length=2.4pt,width=2.6pt]},
                           draw=dig2cond, line width=0.65pt, dash pattern=on 1.6pt off 1.1pt},
  digjunction/.style    = {circle, fill=digInk, inner sep=0pt, minimum size=1.7pt},
  digcondjunction/.style= {circle, fill=dig2cond, inner sep=0pt, minimum size=1.9pt},
  digpanel/.style       = {draw=digRule, line width=0.4pt, rounded corners=2pt,
                           fill=white},
  diglabel/.style       = {font=\sffamily\scriptsize\bfseries, text=digInk},
}

%% file: figures/assets/method_data.tex
\def\mCondDim{64}

\def\mGainQKV{0.05}
\def\mGainGate{0.05}
\def\mGainFiLM{0.30}
\def\mGainTime{0.25}

%% file: figures/assets/fig2_prior_data.tex
\def\cdOwnNTree{6}

\def\cdOwnGapsTree{0}
\def\cdStepTree{1.000000/0.000000/0/0,0.966530/1.000000/0/0,0.475221/0.966530/0/0,0.153580/0.475221/0/0,0.034445/0.153580/0/0,0.007095/0.034445/0/1}

\def\cdOwnNZinc{14}

\def\cdOwnGapsZinc{6}
\def\cdStepZinc{0.197225/0.000000/0/0,1.000000/0.197225/0/0,0.663098/1.000000/0/0,0.022411/0.663098/0/1,0.016993/0.000000/1/0,0.187387/0.016993/1/0,0.107607/0.187387/1/1,0.178136/0.000000/2/0,0.076764/0.178136/2/1,0.035137/0.000000/3/1,0.029796/0.000000/4/0,0.012368/0.029796/4/1,0.018961/0.000000/5/1,0.005647/0.000000/6/1}

%% file: figures/assets/fig2_traj_data.tex
\def\trajTreeBlack{(0.0000,0.0817) -- (0.0208,0.0101) -- (0.0417,0.1639) -- (0.0625,0.1605) -- (0.0833,0.3179) -- (0.1042,0.3098) -- (0.1250,0.1483) -- (0.1458,0.2293) -- (0.1667,0.1189) -- (0.1875,0.1414) -- (0.2083,0.2567) -- (0.2292,0.3169) -- (0.2500,0.3295) -- (0.2708,0.1749) -- (0.2917,0.1411) -- (0.3125,0.0258) -- (0.3333,0.1490) -- (0.3542,0.1829) -- (0.3750,0.2485) -- (0.3958,0.3392) -- (0.4167,0.3553) -- (0.4375,0.2800) -- (0.4583,0.2420) -- (0.4792,0.2038) -- (0.5000,0.1077) -- (0.5208,0.0912) -- (0.5417,-0.0157) -- (0.5625,-0.0338) -- (0.5833,-0.0878) -- (0.6042,-0.1181) -- (0.6250,0.0593) -- (0.6458,-0.0055) -- (0.6667,0.0363) -- (0.6875,-0.0856) -- (0.7083,-0.0781) -- (0.7292,-0.0442) -- (0.7500,-0.0545) -- (0.7708,-0.0882) -- (0.7917,-0.0832) -- (0.8125,-0.0445) -- (0.8333,0.0728) -- (0.8542,-0.0223) -- (0.8750,-0.0315) -- (0.8958,0.0026) -- (0.9167,0.0079) -- (0.9375,0.0392) -- (0.9583,0.0536) -- (0.9792,0.0275) -- (1.0000,0.0000)}
\def\trajTreeGreyA{(0.0000,0.4083) -- (0.0208,0.2589) -- (0.0417,0.1806) -- (0.0625,0.1054) -- (0.0833,0.3329) -- (0.1042,0.4635) -- (0.1250,0.3663) -- (0.1458,0.3990) -- (0.1667,0.4131) -- (0.1875,0.3821) -- (0.2083,0.3036) -- (0.2292,0.2258) -- (0.2500,0.4641) -- (0.2708,0.2635) -- (0.2917,0.0671) -- (0.3125,0.2715) -- (0.3333,0.3053) -- (0.3542,0.2860) -- (0.3750,0.3703) -- (0.3958,0.3100) -- (0.4167,0.2975) -- (0.4375,0.1263) -- (0.4583,-0.0307) -- (0.4792,0.1170) -- (0.5000,0.2381) -- (0.5208,0.2344) -- (0.5417,0.1971) -- (0.5625,0.2881) -- (0.5833,0.3541) -- (0.6042,0.3710) -- (0.6250,0.3817) -- (0.6458,0.3172) -- (0.6667,0.3198) -- (0.6875,0.3141) -- (0.7083,0.2978) -- (0.7292,0.2358) -- (0.7500,0.3044) -- (0.7708,0.2511) -- (0.7917,0.2439) -- (0.8125,0.2968) -- (0.8333,0.3179) -- (0.8542,0.3889) -- (0.8750,0.4706) -- (0.8958,0.5626) -- (0.9167,0.5824) -- (0.9375,0.5826) -- (0.9583,0.5872) -- (0.9792,0.5861) -- (1.0000,0.5800)}
\def\trajTreeGreyB{(0.0000,-0.2450) -- (0.0208,-0.0358) -- (0.0417,0.0157) -- (0.0625,0.1800) -- (0.0833,0.3656) -- (0.1042,0.2963) -- (0.1250,0.2612) -- (0.1458,0.1147) -- (0.1667,0.0921) -- (0.1875,-0.0223) -- (0.2083,0.2859) -- (0.2292,0.2902) -- (0.2500,0.1507) -- (0.2708,0.1363) -- (0.2917,0.2456) -- (0.3125,0.2795) -- (0.3333,0.0208) -- (0.3542,0.2919) -- (0.3750,0.3361) -- (0.3958,0.2676) -- (0.4167,0.1746) -- (0.4375,0.0955) -- (0.4583,0.0621) -- (0.4792,0.1215) -- (0.5000,0.1823) -- (0.5208,0.1995) -- (0.5417,0.1331) -- (0.5625,0.2289) -- (0.5833,0.2388) -- (0.6042,0.2121) -- (0.6250,-0.0208) -- (0.6458,-0.0112) -- (0.6667,0.0329) -- (0.6875,0.0925) -- (0.7083,-0.0277) -- (0.7292,-0.0245) -- (0.7500,-0.0507) -- (0.7708,-0.0681) -- (0.7917,-0.0917) -- (0.8125,-0.1109) -- (0.8333,-0.1607) -- (0.8542,-0.2065) -- (0.8750,-0.2987) -- (0.8958,-0.3195) -- (0.9167,-0.3357) -- (0.9375,-0.3965) -- (0.9583,-0.4153) -- (0.9792,-0.4026) -- (1.0000,-0.4400)}
\def\trajTreeGreyC{(0.0000,0.2450) -- (0.0208,0.0671) -- (0.0417,0.3564) -- (0.0625,0.2827) -- (0.0833,0.0544) -- (0.1042,-0.0344) -- (0.1250,0.0881) -- (0.1458,-0.0187) -- (0.1667,-0.0884) -- (0.1875,-0.2816) -- (0.2083,-0.1700) -- (0.2292,-0.2764) -- (0.2500,-0.3964) -- (0.2708,-0.3192) -- (0.2917,-0.4764) -- (0.3125,-0.2822) -- (0.3333,-0.2382) -- (0.3542,-0.2297) -- (0.3750,-0.1274) -- (0.3958,-0.2176) -- (0.4167,-0.2854) -- (0.4375,-0.1368) -- (0.4583,-0.0925) -- (0.4792,-0.2555) -- (0.5000,-0.0700) -- (0.5208,-0.0700) -- (0.5417,-0.0182) -- (0.5625,0.0216) -- (0.5833,0.1215) -- (0.6042,0.1699) -- (0.6250,0.0861) -- (0.6458,0.1772) -- (0.6667,0.0817) -- (0.6875,0.1319) -- (0.7083,0.1469) -- (0.7292,0.1281) -- (0.7500,0.1005) -- (0.7708,0.0892) -- (0.7917,0.0415) -- (0.8125,0.1103) -- (0.8333,0.1781) -- (0.8542,0.2188) -- (0.8750,0.1767) -- (0.8958,0.2399) -- (0.9167,0.1744) -- (0.9375,0.2276) -- (0.9583,0.2921) -- (0.9792,0.3036) -- (1.0000,0.3300)}
\def\trajTreeFaintA{(0.0000,-0.5717) -- (0.0208,-0.4769) -- (0.0417,-0.4011) -- (0.0625,-0.5925) -- (0.0833,-0.5384) -- (0.1042,-0.5727) -- (0.1250,-0.5650) -- (0.1458,-0.7426) -- (0.1667,-0.7031) -- (0.1875,-0.5062) -- (0.2083,-0.2012) -- (0.2292,-0.2067) -- (0.2500,-0.2853) -- (0.2708,-0.2165) -- (0.2917,-0.0701) -- (0.3125,-0.1431) -- (0.3333,-0.3647) -- (0.3542,-0.3443) -- (0.3750,-0.3700) -- (0.3958,-0.4427) -- (0.4167,-0.3857) -- (0.4375,-0.3455) -- (0.4583,-0.3301) -- (0.4792,-0.4269) -- (0.5000,-0.4739) -- (0.5208,-0.3098) -- (0.5417,-0.4114) -- (0.5625,-0.4315) -- (0.5833,-0.4724) -- (0.6042,-0.4620) -- (0.6250,-0.5836) -- (0.6458,-0.5685) -- (0.6667,-0.5509) -- (0.6875,-0.4610) -- (0.7083,-0.3600) -- (0.7292,-0.4579) -- (0.7500,-0.6039) -- (0.7708,-0.6059) -- (0.7917,-0.6442) -- (0.8125,-0.7035) -- (0.8333,-0.7780) -- (0.8542,-0.8231) -- (0.8750,-0.7609) -- (0.8958,-0.7847) -- (0.9167,-0.6961) -- (0.9375,-0.7369) -- (0.9583,-0.7256) -- (0.9792,-0.7273) -- (1.0000,-0.7200)}
\def\trajTreeFaintB{(0.0000,-0.4083) -- (0.0208,-0.1432) -- (0.0417,-0.0661) -- (0.0625,0.1152) -- (0.0833,0.1761) -- (0.1042,0.1514) -- (0.1250,0.1833) -- (0.1458,0.0957) -- (0.1667,0.1236) -- (0.1875,0.2595) -- (0.2083,0.2021) -- (0.2292,0.4572) -- (0.2500,0.1726) -- (0.2708,0.2622) -- (0.2917,0.2721) -- (0.3125,0.2206) -- (0.3333,0.1724) -- (0.3542,0.2783) -- (0.3750,0.3144) -- (0.3958,0.4221) -- (0.4167,0.6155) -- (0.4375,0.4124) -- (0.4583,0.1828) -- (0.4792,0.1004) -- (0.5000,0.1305) -- (0.5208,0.1297) -- (0.5417,0.0095) -- (0.5625,-0.0260) -- (0.5833,-0.0362) -- (0.6042,-0.0345) -- (0.6250,-0.0872) -- (0.6458,0.0466) -- (0.6667,-0.1184) -- (0.6875,-0.1170) -- (0.7083,-0.1417) -- (0.7292,-0.1983) -- (0.7500,-0.2833) -- (0.7708,-0.3900) -- (0.7917,-0.3848) -- (0.8125,-0.4428) -- (0.8333,-0.5009) -- (0.8542,-0.5790) -- (0.8750,-0.5967) -- (0.8958,-0.6378) -- (0.9167,-0.5681) -- (0.9375,-0.5777) -- (0.9583,-0.5898) -- (0.9792,-0.5693) -- (1.0000,-0.5800)}
\def\trajTreeFaintC{(0.0000,-0.0817) -- (0.0208,-0.0374) -- (0.0417,-0.0048) -- (0.0625,-0.0613) -- (0.0833,-0.1439) -- (0.1042,-0.1705) -- (0.1250,-0.1034) -- (0.1458,0.0355) -- (0.1667,-0.1458) -- (0.1875,-0.1518) -- (0.2083,-0.1137) -- (0.2292,-0.1660) -- (0.2500,-0.0517) -- (0.2708,-0.1112) -- (0.2917,-0.2086) -- (0.3125,-0.3274) -- (0.3333,-0.5701) -- (0.3542,-0.4335) -- (0.3750,-0.5166) -- (0.3958,-0.3801) -- (0.4167,-0.2891) -- (0.4375,-0.3594) -- (0.4583,-0.2936) -- (0.4792,-0.4444) -- (0.5000,-0.6040) -- (0.5208,-0.5305) -- (0.5417,-0.5862) -- (0.5625,-0.5734) -- (0.5833,-0.6291) -- (0.6042,-0.5770) -- (0.6250,-0.6308) -- (0.6458,-0.5757) -- (0.6667,-0.4351) -- (0.6875,-0.3809) -- (0.7083,-0.3269) -- (0.7292,-0.3537) -- (0.7500,-0.4568) -- (0.7708,-0.4953) -- (0.7917,-0.5866) -- (0.8125,-0.6047) -- (0.8333,-0.5067) -- (0.8542,-0.4249) -- (0.8750,-0.4065) -- (0.8958,-0.4012) -- (0.9167,-0.3867) -- (0.9375,-0.3714) -- (0.9583,-0.3564) -- (0.9792,-0.3302) -- (1.0000,-0.2900)}
\def\trajTreeFaintD{(0.0000,0.5717) -- (0.0208,0.3926) -- (0.0417,0.1051) -- (0.0625,0.1332) -- (0.0833,-0.0329) -- (0.1042,-0.0847) -- (0.1250,0.0005) -- (0.1458,0.0114) -- (0.1667,0.1003) -- (0.1875,0.1061) -- (0.2083,-0.0524) -- (0.2292,-0.0299) -- (0.2500,0.1008) -- (0.2708,0.1381) -- (0.2917,-0.0544) -- (0.3125,0.0519) -- (0.3333,0.1368) -- (0.3542,0.1922) -- (0.3750,0.2501) -- (0.3958,0.1697) -- (0.4167,0.1832) -- (0.4375,0.2417) -- (0.4583,0.3127) -- (0.4792,0.3731) -- (0.5000,0.4220) -- (0.5208,0.3624) -- (0.5417,0.2796) -- (0.5625,0.3207) -- (0.5833,0.3038) -- (0.6042,0.3389) -- (0.6250,0.2402) -- (0.6458,0.1897) -- (0.6667,0.2735) -- (0.6875,0.3904) -- (0.7083,0.4067) -- (0.7292,0.4056) -- (0.7500,0.3295) -- (0.7708,0.3775) -- (0.7917,0.5093) -- (0.8125,0.5014) -- (0.8333,0.6423) -- (0.8542,0.6725) -- (0.8750,0.6433) -- (0.8958,0.6640) -- (0.9167,0.7069) -- (0.9375,0.6653) -- (0.9583,0.6651) -- (0.9792,0.6824) -- (1.0000,0.7200)}
\def\trajZincBlack{(0.0000,0.1867) -- (0.0208,0.2053) -- (0.0417,-0.0546) -- (0.0625,-0.2200) -- (0.0833,-0.4685) -- (0.1042,-0.4140) -- (0.1250,-0.2503) -- (0.1458,-0.0965) -- (0.1667,-0.0351) -- (0.1875,-0.0938) -- (0.2083,0.0039) -- (0.2292,-0.1164) -- (0.2500,0.0312) -- (0.2708,0.0272) -- (0.2917,0.1278) -- (0.3125,0.0158) -- (0.3333,0.0385) -- (0.3542,-0.0510) -- (0.3750,-0.0467) -- (0.3958,-0.1714) -- (0.4167,-0.0798) -- (0.4375,-0.1093) -- (0.4583,-0.1300) -- (0.4792,-0.1379) -- (0.5000,-0.1936) -- (0.5208,0.0334) -- (0.5417,-0.1510) -- (0.5625,-0.1028) -- (0.5833,0.0165) -- (0.6042,0.1230) -- (0.6250,0.0360) -- (0.6458,0.1784) -- (0.6667,0.2279) -- (0.6875,0.1459) -- (0.7083,0.1629) -- (0.7292,0.0614) -- (0.7500,0.0219) -- (0.7708,-0.1114) -- (0.7917,-0.0771) -- (0.8125,-0.0770) -- (0.8333,-0.0389) -- (0.8542,-0.0455) -- (0.8750,-0.0158) -- (0.8958,0.0291) -- (0.9167,0.0688) -- (0.9375,0.0242) -- (0.9583,0.0535) -- (0.9792,0.0176) -- (1.0000,0.0000)}
\def\trajZincGreyA{(0.0000,-0.1867) -- (0.0208,-0.1916) -- (0.0417,-0.2698) -- (0.0625,-0.4975) -- (0.0833,-0.2756) -- (0.1042,-0.0915) -- (0.1250,-0.1591) -- (0.1458,-0.1641) -- (0.1667,-0.3598) -- (0.1875,-0.2482) -- (0.2083,-0.3684) -- (0.2292,-0.4562) -- (0.2500,-0.5881) -- (0.2708,-0.6162) -- (0.2917,-0.6489) -- (0.3125,-0.7245) -- (0.3333,-0.6472) -- (0.3542,-0.6996) -- (0.3750,-0.6672) -- (0.3958,-0.4930) -- (0.4167,-0.2519) -- (0.4375,-0.2618) -- (0.4583,-0.3829) -- (0.4792,-0.3669) -- (0.5000,-0.4473) -- (0.5208,-0.6459) -- (0.5417,-0.5887) -- (0.5625,-0.4641) -- (0.5833,-0.4924) -- (0.6042,-0.5397) -- (0.6250,-0.3758) -- (0.6458,-0.5397) -- (0.6667,-0.4926) -- (0.6875,-0.3930) -- (0.7083,-0.5082) -- (0.7292,-0.5272) -- (0.7500,-0.4979) -- (0.7708,-0.6026) -- (0.7917,-0.5184) -- (0.8125,-0.5006) -- (0.8333,-0.5186) -- (0.8542,-0.5077) -- (0.8750,-0.5280) -- (0.8958,-0.5071) -- (0.9167,-0.5388) -- (0.9375,-0.5996) -- (0.9583,-0.5640) -- (0.9792,-0.5650) -- (1.0000,-0.5500)}
\def\trajZincGreyB{(0.0000,0.3733) -- (0.0208,0.2869) -- (0.0417,0.2962) -- (0.0625,0.4712) -- (0.0833,0.5277) -- (0.1042,0.6396) -- (0.1250,0.3969) -- (0.1458,0.1251) -- (0.1667,-0.2417) -- (0.1875,-0.2630) -- (0.2083,-0.3388) -- (0.2292,-0.3159) -- (0.2500,-0.5589) -- (0.2708,-0.4311) -- (0.2917,-0.2541) -- (0.3125,-0.1558) -- (0.3333,-0.1781) -- (0.3542,-0.0707) -- (0.3750,-0.2604) -- (0.3958,-0.3185) -- (0.4167,-0.2300) -- (0.4375,-0.1513) -- (0.4583,-0.1642) -- (0.4792,-0.1954) -- (0.5000,-0.2978) -- (0.5208,-0.0417) -- (0.5417,0.0771) -- (0.5625,0.0186) -- (0.5833,-0.0192) -- (0.6042,0.0980) -- (0.6250,0.0448) -- (0.6458,0.1307) -- (0.6667,0.0529) -- (0.6875,0.1190) -- (0.7083,0.0810) -- (0.7292,0.0076) -- (0.7500,0.1857) -- (0.7708,0.2049) -- (0.7917,0.0885) -- (0.8125,0.2128) -- (0.8333,0.2674) -- (0.8542,0.3381) -- (0.8750,0.3630) -- (0.8958,0.3628) -- (0.9167,0.3835) -- (0.9375,0.4475) -- (0.9583,0.4578) -- (0.9792,0.4353) -- (1.0000,0.4700)}
\def\trajZincGreyC{(0.0000,0.0000) -- (0.0208,0.0017) -- (0.0417,-0.0239) -- (0.0625,0.0169) -- (0.0833,-0.0175) -- (0.1042,0.0811) -- (0.1250,0.0854) -- (0.1458,0.0266) -- (0.1667,0.1527) -- (0.1875,-0.0830) -- (0.2083,0.0282) -- (0.2292,-0.2353) -- (0.2500,-0.2613) -- (0.2708,-0.4258) -- (0.2917,-0.4485) -- (0.3125,-0.6068) -- (0.3333,-0.5566) -- (0.3542,-0.4787) -- (0.3750,-0.3737) -- (0.3958,-0.4484) -- (0.4167,-0.4949) -- (0.4375,-0.2888) -- (0.4583,-0.4235) -- (0.4792,-0.4709) -- (0.5000,-0.4650) -- (0.5208,-0.4509) -- (0.5417,-0.4169) -- (0.5625,-0.3602) -- (0.5833,-0.3255) -- (0.6042,-0.2970) -- (0.6250,-0.1728) -- (0.6458,-0.1983) -- (0.6667,-0.1546) -- (0.6875,-0.1850) -- (0.7083,-0.2557) -- (0.7292,-0.2878) -- (0.7500,-0.3448) -- (0.7708,-0.2847) -- (0.7917,-0.1653) -- (0.8125,-0.1984) -- (0.8333,-0.1437) -- (0.8542,-0.1708) -- (0.8750,-0.2532) -- (0.8958,-0.2178) -- (0.9167,-0.2734) -- (0.9375,-0.2800) -- (0.9583,-0.2382) -- (0.9792,-0.3015) -- (1.0000,-0.3100)}
\def\trajZincFaintA{(0.0000,0.5600) -- (0.0208,0.5297) -- (0.0417,0.4027) -- (0.0625,0.5121) -- (0.0833,0.6310) -- (0.1042,0.6620) -- (0.1250,0.7696) -- (0.1458,0.7806) -- (0.1667,0.6167) -- (0.1875,0.5052) -- (0.2083,0.4386) -- (0.2292,0.3973) -- (0.2500,0.4813) -- (0.2708,0.4606) -- (0.2917,0.4288) -- (0.3125,0.6545) -- (0.3333,0.7469) -- (0.3542,0.7192) -- (0.3750,0.6231) -- (0.3958,0.5882) -- (0.4167,0.6522) -- (0.4375,0.6532) -- (0.4583,0.6321) -- (0.4792,0.4800) -- (0.5000,0.5387) -- (0.5208,0.5272) -- (0.5417,0.7519) -- (0.5625,0.8055) -- (0.5833,0.8037) -- (0.6042,0.8209) -- (0.6250,0.7550) -- (0.6458,0.7476) -- (0.6667,0.7976) -- (0.6875,0.7109) -- (0.7083,0.7309) -- (0.7292,0.6154) -- (0.7500,0.7371) -- (0.7708,0.7508) -- (0.7917,0.7414) -- (0.8125,0.7304) -- (0.8333,0.7722) -- (0.8542,0.7459) -- (0.8750,0.7523) -- (0.8958,0.7185) -- (0.9167,0.7675) -- (0.9375,0.7255) -- (0.9583,0.7123) -- (0.9792,0.7058) -- (1.0000,0.7200)}
\def\trajZincFaintB{(0.0000,-0.5600) -- (0.0208,-0.5984) -- (0.0417,-0.6338) -- (0.0625,-0.4695) -- (0.0833,-0.5813) -- (0.1042,-0.5705) -- (0.1250,-0.4356) -- (0.1458,-0.3438) -- (0.1667,-0.3601) -- (0.1875,-0.2237) -- (0.2083,-0.4021) -- (0.2292,-0.3242) -- (0.2500,-0.1834) -- (0.2708,-0.3675) -- (0.2917,-0.4108) -- (0.3125,-0.4182) -- (0.3333,-0.4443) -- (0.3542,-0.5036) -- (0.3750,-0.4154) -- (0.3958,-0.5655) -- (0.4167,-0.4149) -- (0.4375,-0.4363) -- (0.4583,-0.4282) -- (0.4792,-0.4146) -- (0.5000,-0.5280) -- (0.5208,-0.4838) -- (0.5417,-0.5046) -- (0.5625,-0.4710) -- (0.5833,-0.4733) -- (0.6042,-0.6000) -- (0.6250,-0.6314) -- (0.6458,-0.7110) -- (0.6667,-0.7323) -- (0.6875,-0.7881) -- (0.7083,-0.7784) -- (0.7292,-0.7534) -- (0.7500,-0.7274) -- (0.7708,-0.7064) -- (0.7917,-0.7046) -- (0.8125,-0.7505) -- (0.8333,-0.7283) -- (0.8542,-0.7204) -- (0.8750,-0.7735) -- (0.8958,-0.7668) -- (0.9167,-0.8070) -- (0.9375,-0.8083) -- (0.9583,-0.8548) -- (0.9792,-0.8430) -- (1.0000,-0.8400)}
\def\trajZincFaintC{(0.0000,-0.3733) -- (0.0208,-0.0955) -- (0.0417,-0.0909) -- (0.0625,-0.0035) -- (0.0833,-0.2539) -- (0.1042,0.0195) -- (0.1250,-0.1294) -- (0.1458,-0.1756) -- (0.1667,-0.2355) -- (0.1875,-0.2556) -- (0.2083,-0.3703) -- (0.2292,-0.5133) -- (0.2500,-0.4874) -- (0.2708,-0.5232) -- (0.2917,-0.6113) -- (0.3125,-0.8136) -- (0.3333,-0.8491) -- (0.3542,-0.7266) -- (0.3750,-0.6871) -- (0.3958,-0.5650) -- (0.4167,-0.4849) -- (0.4375,-0.5025) -- (0.4583,-0.5849) -- (0.4792,-0.5672) -- (0.5000,-0.5098) -- (0.5208,-0.4356) -- (0.5417,-0.3507) -- (0.5625,-0.3739) -- (0.5833,-0.3863) -- (0.6042,-0.5478) -- (0.6250,-0.4104) -- (0.6458,-0.4265) -- (0.6667,-0.3881) -- (0.6875,-0.4923) -- (0.7083,-0.5071) -- (0.7292,-0.5457) -- (0.7500,-0.4371) -- (0.7708,-0.5340) -- (0.7917,-0.5042) -- (0.8125,-0.4564) -- (0.8333,-0.5219) -- (0.8542,-0.5569) -- (0.8750,-0.6142) -- (0.8958,-0.6571) -- (0.9167,-0.6311) -- (0.9375,-0.6490) -- (0.9583,-0.6836) -- (0.9792,-0.6861) -- (1.0000,-0.7000)}

%% file: sections/introduction.tex
\section{Introduction}
\label{sec:intro}

\begin{figure}[!ht]
\centering
\input{figures/fig1_hero}
\caption{
\textbf{One model per domain versus DiG$^{2}$.} \textcolor{digPlanar}{\rule{0.55em}{0.55em}}\,Planar, \textcolor{digSBM}{\rule{0.55em}{0.55em}}\,SBM, \textcolor{digTree}{\rule{0.55em}{0.55em}}\,Tree, \textcolor{digQM9}{\rule{0.55em}{0.55em}}\,QM9, \textcolor{digZINC}{\rule{0.55em}{0.55em}}\,ZINC.
\textbf{(1)}~Today, separate models $\theta_{1},\dots,\theta_{5}$ learn one
distribution to re-create their input graphs.
\textbf{(2)}~DiG$^{2}$ reads multiple domains at once, generating from a single model,
as tagged by a learned token $\vz_{d}$.}
\label{fig:hero}
\end{figure}
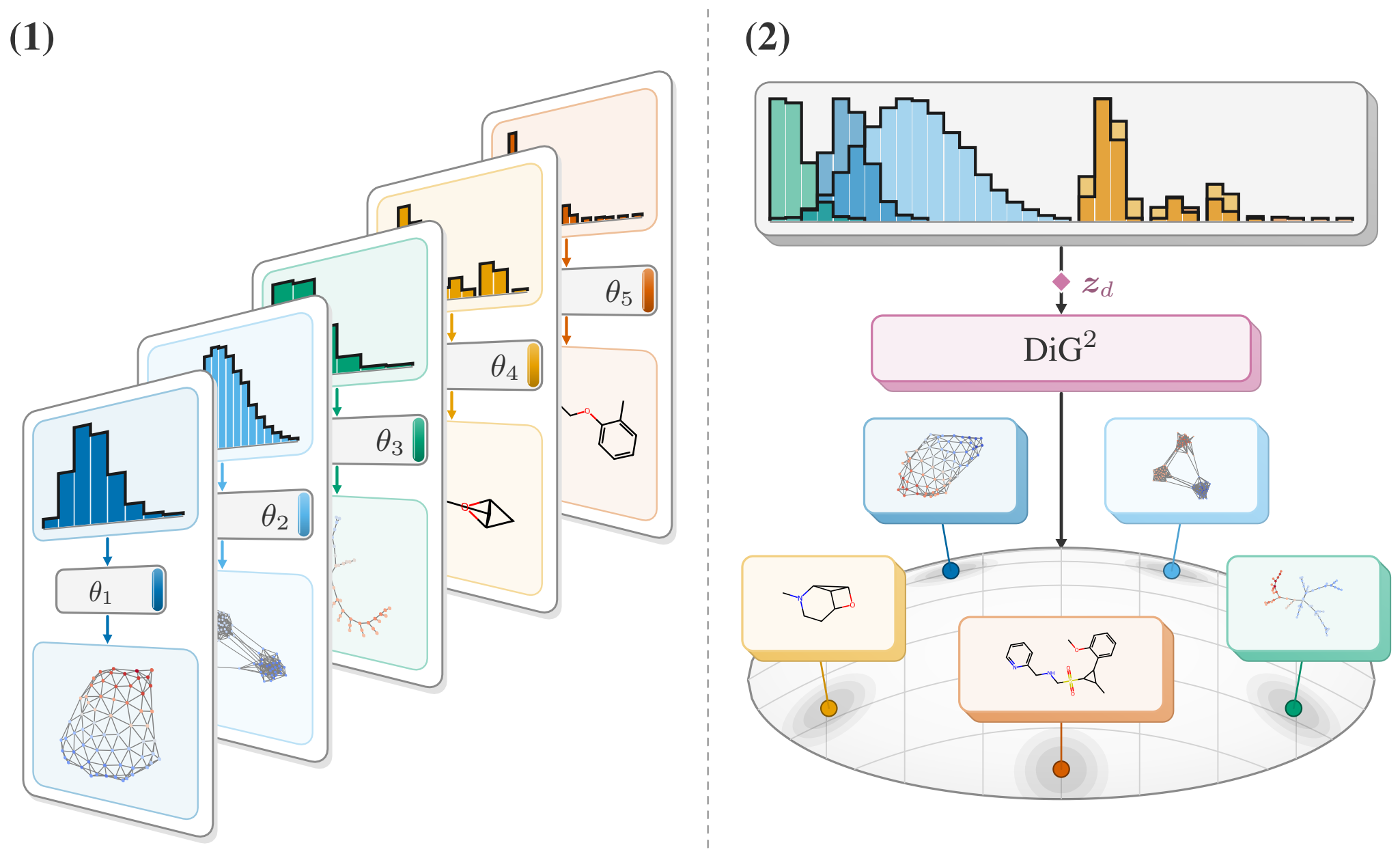

Graph generative models are overwhelmingly \emph{single-domain}, current research is focused on developing models trained to reproduce one dataset per model~\citep{vignac2023digress,qin2024defog,huang2026hog}. This modeling paradigm produces strong results within a target domain, enabling reconstruction of difficult structures, such as planar graphs or drug-like
molecules~\citep{lee1980two,brown2019guacamol}. Outside of generation, graph learning is moving towards one model for many domains. Multi-domain graph foundation models train a single network across heterogeneous graph families and transfer between them~\citep{sun2026multi,wang2025mdgfm,wang2026generative}. However, graph generation has no such counterpart yet. We find that building a multi-domain graph generative model has two hurdles not present for generative models within the image and text domain.

The first challenge is proper \emph{size and ordering}. Image and text generation operate on fixed-length output representations, such as pixel grids or token sequences, enabling consistent learning objectives and memory pressure~\citep{austin2021d3pm,radford2018improving}. In graphs, node counts vary across and within domains. So, any model must manage differing combinatorial \emph{and} canonical orderings between datasets~\citep{you2018graphrnn}.
The second challenge concerns \emph{vocabulary}. Image and text modalities share pixel values or token vocabularies~\citep{yu2015lsun,devlin2019bert}, reducing risk of leakage within transfer settings. In contrast, graph domains are semantically \emph{disjoint}. A molecular double-bond has no counterpart among the edge classes of a stochastic block model~\citep{ramakrishnan2014qm9,abbe2018community, irwin2012zinc, polykovskiy2020molecular}.

Existing multi-domain graph generation work sidesteps node and edge vocabulary entirely, learning directly from adjacency matrix structure~\citep{wang2025large}. Separately, multi-domain learning without graph generation indicates promise applying domain labels~\citep{yuan2025much} or graphon structure~\citep{yuan2026graver} to transfer performance across graph domains. Despite this success, our initial investigation found key limitations (App.~\ref{app:initial}). %
A generative model conditioned on graph-level domain labels makes the model domain-\emph{aware}, but remains incapable of generating domain-\emph{separable} samples.
The lack of vocabulary-aware, multi-domain graph generative models leaves an open question. \emph{Can a graph model successfully generate several structured domains at once?}

In order to answer this question, we propose extending the graph generation problem to consist of \emph{multiple structural domains with distinct vocabularies}. This is currently unprecedented, given that LGGM is structure only and state-of-the-art graph generative models have only been tested on single domains, as shown in the left-half of Figure~\ref{fig:hero}. To overcome this, we introduce the Discrete-Interpolative Graph Generation model, or DiG$^{2}$, as shown in the right-half of  Figure~\ref{fig:hero}. DiG$^{2}$ achieves successful multi-domain graph generation through a specialized domain conditioning block coupled with bridge denoising between explicit per-domain priors.

\paragraph{Contributions.}
\begin{enumerate}
\item We develop the first structurally-distinct multi-domain graph generation benchmarking protocol, enabling evaluation of current graph generative models under the multi-domain graph generation problem (\Secref{sec:benchmark}). This protocol involves previously-untested challenges, including: disjoint per-domain vocabularies, mixing domains under a shared budget, and generating structurally-distinctive targets (App.~\ref{app:protocol}).
\item We propose DiG$^{2}$, a discrete  flow-matching model integrating a structured per-domain priors for the first successful demonstration of structurally-distinct multi-domain graph generation (\Secref{sec:formulation}), .
\item We analyze the dynamics of multi-domain graph generation, centered around: benchmarking the capabilities of existing models and DiG$^{2}$ on difficult domains (\Secref{sec:benchmark}), determining which components make DiG$^{2}$ effective at multi-domain graph generation (\Secref{sec:ablation}), and uncovering DiG$^{2}$'s ability for de novo molecular generation (\Secref{sec:interp}).
\end{enumerate}

%% file: figures/fig1_hero.tex
{\rmfamily\fontsize{6.5}{7.8}\selectfont%
\def\svgwidth{\linewidth}%
\IfFileExists{figures/fig1_hero_C.pdf_tex}%
  {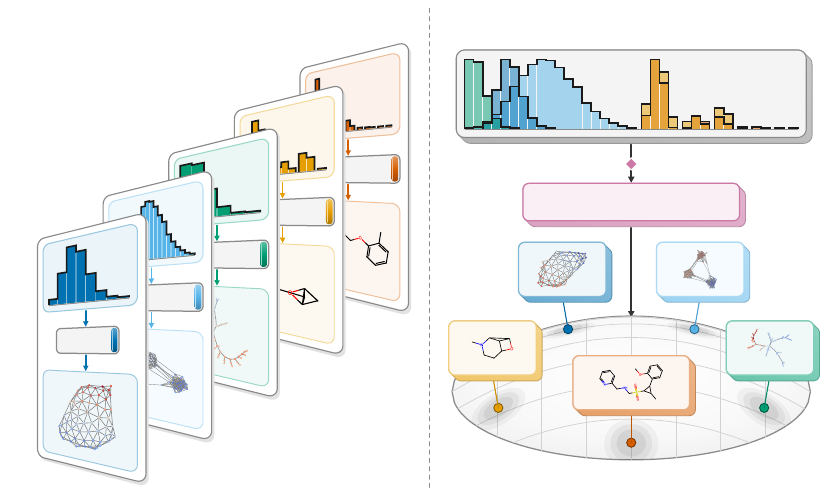}%
  {\input{fig1_hero_C.pdf_tex}}%
}%

%% file: figures/fig1_hero_C.pdf_tex
\begingroup%
  \makeatletter%
  \providecommand\color[2][]{%
    \errmessage{(Inkscape) Color is used for the text in Inkscape, but the package 'color.sty' is not loaded}%
    \renewcommand\color[2][]{}%
  }%
  \providecommand\transparent[1]{%
    \errmessage{(Inkscape) Transparency is used (non-zero) for the text in Inkscape, but the package 'transparent.sty' is not loaded}%
    \renewcommand\transparent[1]{}%
  }%
  \providecommand\rotatebox[2]{#2}%
  \newcommand*\fsize{\dimexpr\f@size pt\relax}%
  \newcommand*\lineheight[1]{\fontsize{\fsize}{#1\fsize}\selectfont}%
  \ifx\svgwidth\undefined%
    \setlength{\unitlength}{396bp}%
    \ifx\svgscale\undefined%
      \relax%
    \else%
      \setlength{\unitlength}{\unitlength * \real{\svgscale}}%
    \fi%
  \else%
    \setlength{\unitlength}{\svgwidth}%
  \fi%
  \global\let\svgwidth\undefined%
  \global\let\svgscale\undefined%
  \makeatother%
  \begin{picture}(1,0.60101010)%
    \lineheight{1}%
    \setlength\tabcolsep{0pt}%
    \put(0,0){\includegraphics[width=\unitlength,page=1]{fig1_hero_C.pdf}}%
    \put(0.46966408,0.38830554){\color[rgb]{0.2000,0.2000,0.2000}\scalebox{0.9702}[0.9612]{\makebox(0,0)[rt]{\lineheight{1.25}\smash{\begin{tabular}[t]{r}{\fontsize{9}{10.5}\selectfont $\theta_{5}$}\end{tabular}}}}}%
    \put(0.39011863,0.33653786){\color[rgb]{0.2000,0.2000,0.2000}\scalebox{0.9702}[0.9612]{\makebox(0,0)[rt]{\lineheight{1.25}\smash{\begin{tabular}[t]{r}{\fontsize{9}{10.5}\selectfont $\theta_{4}$}\end{tabular}}}}}%
    \put(0.31057317,0.28477018){\color[rgb]{0.2000,0.2000,0.2000}\scalebox{0.9702}[0.9612]{\makebox(0,0)[rt]{\lineheight{1.25}\smash{\begin{tabular}[t]{r}{\fontsize{9}{10.5}\selectfont $\theta_{3}$}\end{tabular}}}}}%
    \put(0.23102772,0.23300251){\color[rgb]{0.2000,0.2000,0.2000}\scalebox{0.9702}[0.9612]{\makebox(0,0)[rt]{\lineheight{1.25}\smash{\begin{tabular}[t]{r}{\fontsize{9}{10.5}\selectfont $\theta_{2}$}\end{tabular}}}}}%
    \put(0.09974550,0.18107588){\color[rgb]{0.2000,0.2000,0.2000}\scalebox{0.8179}[0.8826]{\makebox(0,0)[t]{\lineheight{1.25}\smash{\begin{tabular}[t]{c}{\fontsize{9}{10.5}\selectfont $\theta_{1}$}\end{tabular}}}}}%
    \put(0.03535354,0.56313131){\color[rgb]{0.2000,0.2000,0.2000}\makebox(0,0)[lt]{\lineheight{1.25}\smash{\begin{tabular}[t]{l}\fontsize{11}{13}\selectfont\textbf{(1)}\end{tabular}}}}%
    \put(0.77878788,0.39469697){\color[rgb]{0.6000,0.3569,0.4902}\makebox(0,0)[lt]{\lineheight{1.25}\smash{\begin{tabular}[t]{l}{\fontsize{9}{10.5}\selectfont $\vz_{d}$}\end{tabular}}}}%
    \put(0.76515152,0.34747475){\color[rgb]{0.2000,0.2000,0.2000}\makebox(0,0)[t]{\lineheight{1.25}\smash{\begin{tabular}[t]{c}{\fontsize{9.5}{11}\selectfont DiG$^{2}$}\end{tabular}}}}%
    \put(0.54545455,0.56313131){\color[rgb]{0.2000,0.2000,0.2000}\makebox(0,0)[lt]{\lineheight{1.25}\smash{\begin{tabular}[t]{l}\fontsize{11}{13}\selectfont\textbf{(2)}\end{tabular}}}}%
  \end{picture}%
\endgroup%

%% file: sections/background.tex
\section{Background}
\label{sec:background}

\textbf{Adjacency-based graph generation.}
Seminal work with GraphRNN~\citep{you2018graphrnn} applied state transitions to autoregressively generate graphs from adjacency representations. More recently, G$^{2}$PT incorporated tokenized graph representations for powerful autoregressive generation~\citep{g2pt2025}. In parallel, diffusion emerged as the dominant paradigm for graph generation. GDSS~\citep{jo2022score} introduced continuous-time score-based generation by jointly perturbing node features and the adjacency matrix through a system of stochastic differential equations. DiGress~\citep{vignac2023digress} instead formulated graph generation directly in the discrete domain, denoising categorical node and edge features under marginal transition chains derived from discrete diffusion~\citep{austin2021d3pm}. DeFoG~\citep{qin2024defog} subsequently replaced the discrete diffusion chain with discrete flow matching~\citep{lipman2023flow,campbell2024generative}, parameterizing transitions through a continuous-time rate matrix. SPECTRE~\citep{martinkus2022spectre}, LGD~\citep{zhou2024unifying}, and GraphBSI~\citep{petersen2026discrete} all demonstrate promise for graph generation.

\textbf{Structured and hybrid graph diffusion.}
Recent approaches introduce structural factorizations beyond one-shot representations. PARD~\citep{zhao2024pard} combines autoregressive and diffusion by decomposing generation into blocks, where a shared model learns the conditional distribution over said blocks. HOG-Diff~\citep{huang2026hog} integrates higher-order graph skeletons into bridge diffusion, which guides generation from coarse structure toward pairwise connectivity.  Despite advancements, prior approaches are predominantly domain-specific and do not train a single model to jointly capture heterogeneous graph domains.


\textbf{Multi-domain graph generation.} Domain-conditioned generation spans drug discovery~\citep{hu2025domain}, vision~\citep{khan2026frozen,belal2025attention} and decision-making~\citep{ajay2022conditional}. Conditional priors are likewise established for continuous 
models~\citep{lee2021priorgrad,popov2021grad,zhang2023shiftddpms,issachar2025designing},
 where all conditions share one space and the class-wise priors overlap. To the best of our knowledge, other multi-domain graph generators either drop node and edge features~\citep{tang2026cross, wang2025large} or share one vocabulary and prior across related chemical domains~\citep{pmlr-v267-joshi25a,liu2026graph}. Our proposed multi-domain graph setting is distinct from prior work, restricting each domain's prior to its own 
 vocabulary, especially when node- and edge-vocabularies are \emph{disjoint} across 
 domains.

\textbf{De novo molecule generation.} Masked diffusion
language models~\citep{sahoo2024simple} and image diffusion~\citep{rombach2022high} obtain broad coverage over a shared vocabularies, enabling generalization transfer. Despite the difficulty for transfer in graphs, discrete diffusion has shown strong in-context learning~\citep{lee2025genmol,liu2026graph} or even OOD-exploration~\citep{lee2023exploring} for \emph{de novo} molecule generation with property predictor supplying guidance.

\textbf{Discrete Diffusion for Multiple Domains.} Current graph diffusion is trained from a shared noise distribution, predicting over one vocabulary. LGGM is multi-domain, but discards vocabulary to operate purely on the adjacency matrix. DiG$^{2}$ integrates domain conditioning at every denoiser layer, scoping predictions across each domain's vocabulary.  Relative to LGGM, DiG$^{2}$ is novel in two respects. 
 It models node and edge features instead of adjacency structure alone, 
 and it restricts each domain's prior to that domain's vocabulary, 
 so the requested domain stays identifiable throughout denoising (Lemma~\ref{lem:domain}). 
Single-domain models fix one prior per dataset~\citep{qin2024defog}.
Like HOG-Diff, DiG$^{2}$ samples along a bridge. 
However, DiG$^{2}$'s sampler is specialized to disambiguate between a domain-\emph{separable} graph and its prior, overcoming the difficulty posed by a lack of domain-awareness (App.~\ref{app:initial}).


%% file: sections/formulation.tex
\section{Model Formulation}
\label{sec:formulation}

\subsection{Preliminaries and overview}
\label{sec:form-overview}

\textbf{Notation.} A graph $\gG = (\mX, \tE)$ with $n$ nodes has one-hot node classes
$\mX \in \{0,1\}^{n \times C_X}$ and one-hot edge classes $\tE \in \{0,1\}^{n \times n \times C_E}$,
where ``no edge'' is itself an edge class; $x^i$ is the class of entry $i \in \sI_n$ (a node or node
pair). Every graph belongs to a domain $d \in \sD$; here
$\sD = \{$Planar, SBM, Tree, QM9, ZINC$\}$. All domains share one global vocabulary of node and
edge classes, but domain $d$ only uses its subset $\sV_d$ (i.e.\ bond types exist only in
the molecular domains), $\vq_d$, $\sV_d$ and the embedding $\vz_d$ are rows selected by the domain
one-hot $\ve_d$. Time runs from noise at $t=0$ to data at $t=1$, and $\gG_t$ is a partially corrupted
graph. The prior, the corruption of a given clean graph and the prediction $\vpi$ are categorical and sampled independently per entry. We define a domain as the triple, $d = (P_{\mathrm{data}}(\cdot \mid d), \sV_d, \ve_d)$, which combines: a dataset with it's respective feature distribution, or $P_{\mathrm{data}}(\cdot \mid d)$ along with the declared vocabulary subset and label. Two domains can share a vocabulary, where the structural regime (i.e. Planar, SBM, and Tree) use a binary edge vocabulary. QM9 and ZINC form the molecular regime with distinct molecular vocabularies. The \emph{domain prior} $\vq_d$ is domain $d$'s class marginal over its own training graphs, restricted to $\sV_d$ and selected by $\ve_d$.

\begin{figure}[h!]
\centering
\input{figures/fig2a_ribbon}
\caption{\textbf{DiG$^{2}$ generation.} A unique prior, $\vq_d$, seeds a given domain (\Secref{sec:form-bridge}).
A denoiser $\theta$ is domain-conditioned through every layer, a gate $\vw$ mixes readouts into
one prediction (\Secref{sec:form-denoiser}). Each adjacency is masked over $\sV_d$, restricting learning to relevant domains (\Secref{sec:form-train}). Entries are resolved from the masked $\vpi$ via a shared bridge toward a target domain (\Secref{sec:form-sample}).}
\label{fig:ribbon}
\end{figure}

\subsection{Domain-conditioned corruption and its bridge posterior}
\label{sec:form-bridge}

We define the denoising process over multiple graph domains, where corruption explicitly isolates a domain. To start, DiG$^{2}$ corrupts a clean graph $\gG_1$ toward a
\emph{domain-specific} prior $\vq_d$, rather than one shared endpoint. So, the amount of noise present at time $t$ already reflects the domain $d$.

\begin{definition}[Domain-conditioned corruption process]
\label{def:forward}
For a clean graph $\gG_1$ and domain $d \in \sD$ with prior $\vq_d$, define the corruption
process $\{P_t(\cdot \mid \gG_1, d)\}_{t}$ by the entrywise linear interpolation,
\begin{equation}
P_t(\gG_t \mid \gG_1, d) \;=\; \prod\nolimits_{i \in \sI_n}\big[\,t\,\delta_{x_1^i} + (1-t)\,\vq_d^i\,\big](x_t^i), \qquad \gG_t \sim P_t(\cdot \mid \gG_1, d).
\label{eq:forward}
\end{equation}
At $t=0$, $P_0(\cdot \mid \gG_1, d) = \otimes_i\vq_d^i$ is the domain prior; at $t=1$,
$P_1(\cdot \mid \gG_1, d) = \delta_{\gG_1}$ is the clean graph.
\end{definition}

This generalizes domain-agnostic interpolation paths to $|\sD|$ distinct endpoints indexed by domain.

\textbf{Reverse sampling by direct interpolation.} Generation runs \Eqref{eq:forward} backward from $\vq_d$ to a clean graph. As a linear path, \Eqref{eq:forward}'s reverse transition is available in closed form:

\begin{proposition}[Domain-conditional bridge sampling]
\label{prop:bridge}
Fix $d$, $n$ and $0 \le t < s \le 1$, let $a = \frac{s-t}{1-t} \in (0,1]$, and let
$\pi_d^i(x \mid \gG_t) = P(x_1^i{=}x \mid \gG_t, d)$ under Definition~\ref{def:forward}.
\textbf{(i) Shared kernel:} $P(\gG_s \mid \gG_t, \gG_1, d) = \prod_i \big[a\,\delta_{x_1^i} +
(1-a)\,\delta_{x_t^i}\big](x_s^i)$, which depends on neither $d$ nor $\vq_d$.
\textbf{(ii) Exact step:} averaged over $\gG_1 \sim P(\cdot \mid \gG_t, d)$, maps $P_t(\cdot \mid d)$ to
$P_s(\cdot \mid d)$ for any step size, so iterating it from $\gG_0 \sim \otimes_i \vq_d^i$ returns
$P_{\mathrm{data}}(\cdot \mid d)$; entrywise,
\begin{equation}
P(x_s^i \mid \gG_t, d) \;=\; a\,\pi_d^i(\cdot \mid \gG_t) + (1-a)\,\delta_{x_t^i}.
\label{eq:reverse}
\end{equation}
\textbf{(iii) Closure:} if $\mathrm{supp}\,\vq_d^i \subseteq \sV_d^i$, drawing entries from
$a\,\vpi^i + (1-a)\,\delta_{x_t^i}$ with any $\vpi^i$ on $\sV_d^i$, masked or guided, keeps every iterate in
$\prod_i \sV_d^i$.
\end{proposition}

Per entry, \Eqref{eq:reverse} is the row-stochastic matrix
$\mB^{(a)} = (1-a)\mI + a\,\vone(\vpi^i)^\top$, with no free parameters beyond $a$ and $\vpi^i$.
The domain thus acts only through $\pi_d^i$ and $\sV_d$, and the noisy state identifies the target domain.
Precisely, Definition~\ref{def:forward} is a mixture path of discrete flow matching from~\citep{gat2024discrete,campbell2024generative} with a domain-specific source $\vq_d$, and Proposition~\ref{prop:bridge} is its exact finite-step reverse kernel. We call the resulting sampler a bridge because each domain's path is pinned to its own prior. Proof in App.~\ref{app:bridge-proof}.

\subsection{Multi-depth, domain-conditioned denoiser}
\label{sec:form-denoiser}

Given that a 12-node molecule can be parametrized with fewer message-passing layers relative to a 144-node SBM graph~\citep{gilmer2017neural}, there is no one-size-fits-all architecture to balance learning on the two domains. This motivates development for a per-depth mechanism to optimize the required depth of denoiser prediction layers, activating deeper predictions only when necessary.

\textbf{Multi-depth denoising.} A shared denoiser $f(\cdot\,;\vtheta)$ is
trained to invert Definition~\ref{def:forward}, conditioned on $\gG_t$, $t$, and $d$. Instead of single predictions at the final layer, $f$ exposes candidate reconstructions of $\gG_1$ at several depths simultaneously,
\begin{equation}
\big\{\hat\gG_1^{(\ell)}\big\}_{\ell \in \sK} \;=\; f(\gG_t, t, d\,;\vtheta), \qquad \sK \subset \{1,\dots,L\},
\label{eq:denoiser}
\end{equation}
where $L$ is the number of transformer layers and $\sK$ the set of readout depths. Readouts are combined within a lightweight gate $f_{\mathrm{gate}}(\cdot\,;\vphi)$ conditioned on
the same domain signal,
\begin{equation}
\hat\gG_1 \;=\; \sum_{\ell \in \sK} w_\ell(\vc)\, \hat\gG_1^{(\ell)}, \qquad
\vw(\vc) = \softmax\big(f_{\mathrm{gate}}(\vc\,;\vphi)\big).
\label{eq:combine}
\end{equation}
$f_{\mathrm{gate}}$ is zero-initialized, so every domain starts from the uniform average $\vw = \tfrac{1}{|\sK|}\vone$ and learns its own weighting over depths during training. The gate reads the same conditioning vector $\vc$ as input layers, so weighting differs by domain across each denoising prediction. A per-depth analysis of what the gate learns is in App.~\ref{app:readout}.

\textbf{Domain-conditioning block.} Since our initial investigation determined that graph-level conditioning was too weak for multi-domain graph generation (Appendix~\ref{app:initial}). We introduce a specialized domain-conditioning block, capable of discerning complicated graph structure and learning separation between input domains. The block maximizes domain-separable information within \Eqref{eq:denoiser} entering through a conditioning vector $\vc = [\vz_d \Vert \vs]$ in two parts. The first, $\vz_d$ is a learned embedding table of domain labels, with a row reserved, $\vz_{\varnothing}$, as the null label for classifier-free guidance (\Secref{sec:form-sample}). A label states \emph{which} domain is requested, but not \emph{what its graphs look like}, and $\vs$ supplies the latter. The second mechanism, $\vs$ is a structural summary, obtained via a centered EMA prototype and message-passing. A small GNN $f_{\mathrm{struct}}(\cdot\,;\vomega)$ encodes the current noisy graph into
$\vs_{\mathrm{cur}}$. The same encoder, applied to domain $d$'s clean training graphs, maintains a
prototype $\vs_{\mathrm{proto},d}$: an exponential moving average of their mean embedding, centered on
the mean over domains and $\ell_2$-normalized. Centering shared structure amongst domains, so the
prototype encodes what distinguishes $d$. The two are then fused as
$\vs = f_{\mathrm{fuse}}\big([\vs_{\mathrm{proto},d} \Vert \vs_{\mathrm{cur}}]\big)$. Critically, $\vc$ is
injected into the node and edge representations at \emph{every} layer, rather than as a single graph-level vector alongside them. We provide an illustration of the conditioning block in Figure~\ref{fig:cnd_layer}.

\subsection{Training}
\label{sec:form-train}

Node vocabularies are often disjoint between domains. A molecule sample can have a carbon atom, which can join a triple bond. Whereas, nodes from a planar graph are connected exclusively by single edges. Meaning, a model trained on carbon atoms is likely to predict triple-bonds when attempting to construct a planar graph.

\textbf{Domain-constrained training objective.} Given this risk of vocabulary leakage, we restrict the
cross-entropy objective for each graph to its own domain's valid class support $\sV_d$,
\begin{equation}
\mathcal{L}(\vtheta, \vphi) \;=\; \E_{(\gG_1, d) \sim P_{\mathrm{data}}}
  \Big[\, \CE\big(\hat\gG_1 \!\upharpoonright_{\sV_d},\; \gG_1\big) \Big],
\label{eq:loss}
\end{equation}
where $\upharpoonright_{\sV_d}$ masks the logits of classes outside $\sV_d$ before the softmax.
In effect, the model foregoes a gradient penalty while also avoiding probability assignment to classes outside a target domain. This constraint has no current analogue in single-domain discrete diffusion, all of which train with one shared vocabulary.

\textbf{Training step.} Domain-restriction upgrades training, where each step draws a graph and its domain, $(\gG_1, d) \sim P_{\mathrm{data}}$,
a time $t = \psi(u)$ with $u \sim \mathrm{Unif}(0,1)$ and $\psi(u) = 2u - u^2$, and a corrupted graph
$\gG_t$ from \Eqref{eq:forward}. With probability $p_{\mathrm{drop}}$ the domain conditioning is
replaced by the null conditioning $\vc_{\varnothing}$, so the same network also learns the
unconditional prediction used for guidance. Besides \Eqref{eq:loss} on the gated prediction, each
non-terminal readout depth in $\sK$ receives the same masked cross-entropy, averaged into a
deep-supervision term $\mathcal{L}_{\mathrm{deep\text{-}sup}}$ (App.~\ref{app:impl}).

\subsection{Sampling}
\label{sec:form-sample}

A natural strength of diffusion models is classifier-free guidance (CFG)~\citep{ho2022cfg}. Currently, discrete graph diffusion is rightly-centered around CFG for property-targeting. Within this study, we treat target domains as a property, enabling domain-directed guidance exclusively via a 1-hot label.

\textbf{Domain-directed guidance.} The model evaluates once with the true domain label and once with a null label, giving masked
logits $\vell^{c} = \Fro(\gG_t, t, \vc)\!\upharpoonright_{\sV_d}$ and
$\vell^{u} = \Fro(\gG_t, t, \vc_{\varnothing})\!\upharpoonright_{\sV_d}$. Both
predictions are linearly extrapolated with a domain-specific strength $g(d)$,
\begin{equation}
\vpi \;=\; \softmax\big(\vell^{u} + g(d)\,(\vell^{c} - \vell^{u})\big).
\label{eq:cfg}
\end{equation}
Here $g(d) \in \mathbb{R}$ is a tuned scalar read from a per-domain lookup table, as determined by $\ve_d$. The per-domain extrapolation  within \Eqref{eq:cfg} is an expanded linear form of classifier-free discrete samplers, grounded in domain-specific prior distributions.

\textbf{Sampling loop.} This domain-awareness then allows generation to start from $\gG_0 \sim \otimes_i \vq_d^i$ and take $T$ steps along the grid
$t_k = \psi(k/T)$. At each step it computes $\vpi$ from \Eqref{eq:cfg} and draws the next graph from
\Eqref{eq:reverse}. Every entry is resolved to a draw from $\vpi^i$ with probability $a$ and otherwise
kept. The prediction stands in for $\pi_d^i$, with marginals in place of the joint posterior and a
guidance tilt for $k < T-1$ (App.~\ref{app:proofs}). The final step ($s=1$) reads
out the conditional prediction $\softmax(\vell^{c})$ rather than the guided $\vpi$, so guidance
shapes the trajectory but not the last draw.

\textbf{Training and sampling procedures.} DiG$^{2}$ is trained with Algorithm~\ref{alg:train}, which calls Algorithm~\ref{alg:proto} once per
batch, and sampled with Algorithm~\ref{alg:sample}. Each domain $d$ supplies: a prior ($\vq_d$) as the normalized training class counts over nodes and node pairs, a size
distribution ($P(n \mid d)$), and vocabulary mask
($\upharpoonright_{\sV_d}$). The network $\Fro$ (\Secref{sec:form-denoiser}) is trained via cross-entropy loss ($\CE = \CE_X + \lambda_E \CE_E$) and returns one
logit vector per node and per node pair. The logit vector is combined with the per-depth readouts, which is optimized with the network by
$\mathcal{L}_{\mathrm{deep\text{-}sup}}$. At sampling time, the graph is updated in place and the conditioning $\vc$ is rebuilt from it before every network call. Every step depends on the domain except the final redraws on each entry from $\vpi_i$ with probability $a_k$ (Proposition~\ref{prop:bridge}(i)). We represent null domain-conditioning as $\vc_{\varnothing}$ ( Equation~\ref{eq:cnull}).

\begin{figure}[h!]
\centering
\begin{minipage}[t]{0.48\textwidth}
\begin{algorithm}[H]
\caption{DiG$^{2}$ training}
\label{alg:train}
\begin{algorithmic}[1]
\Require data $P_{\mathrm{data}}$; weights $\vtheta, \vphi, \vomega$
\Ensure trained $\vtheta, \vphi, \vomega$; prototypes $\vs_{\mathrm{proto},d}$
\While{not converged}
    \State draw a batch $B \sim P_{\mathrm{data}}$
    \For{each $(\gG_1, d) \in B$}
        \State $u \sim \mathrm{Unif}(0,1)$;\; $t \leftarrow \psi(u)$
        \State $\gG_t \sim P_t(\cdot \mid \gG_1, d)$ \Comment{noise to $\vq_d$}
        \State $\vs \leftarrow f_{\mathrm{fuse}}([\vs_{\mathrm{proto},d} \Vert f_{\mathrm{struct}}(\gG_t)])$
        \State $\vc \leftarrow [\vz_d \Vert \vs]$, or $\vc_{\varnothing}$ w.p.\ $p_{\mathrm{drop}}$
        \State $\hat\gG_1 \leftarrow \Fro(\gG_t,t,\vc) \!\upharpoonright_{\sV_d}$
    \EndFor
    \State $\mathcal{L} \leftarrow \CE(\hat\gG_1, \gG_1) + \mathcal{L}_{\mathrm{deep\text{-}sup}}$
    \State write prototypes from $B$ (Alg.~\ref{alg:proto})
    \State $\vtheta, \vphi, \vomega \leftarrow$ AdamW step on $\nabla\mathcal{L}$
\EndWhile
\end{algorithmic}
\end{algorithm}
\end{minipage}%
\hfill
\begin{minipage}[t]{0.48\textwidth}
\begin{algorithm}[H]
\caption{DiG$^{2}$ sampling}
\label{alg:sample}
\begin{algorithmic}[1]
\Require domain $d$; steps $T$; $g(d)$; $\Fro$
\State $n \sim P(n \mid d)$ \Comment{graph size}
\State $\gG \leftarrow \gG_0$: $x^i \sim \vq_d^i$ for $i \in \sI_n$ \Comment{prior}
\For{$k = 0, \dots, T-1$}
    \State $t_k, t_{k+1} \leftarrow \psi(k/T),\, \psi((k{+}1)/T)$
    \State $\vell^{c} \leftarrow \Fro(\gG,t_k,\vc) \!\upharpoonright_{\sV_d}$
    \If{$k < T-1$}
        \State $\vell^{u} \leftarrow \Fro(\gG,t_k,\vc_{\varnothing}) \!\upharpoonright_{\sV_d}$
        \State $\vpi \leftarrow \softmax\big(\vell^{u} + g(d)(\vell^{c} - \vell^{u})\big)$
    \Else
        \State $\vpi \leftarrow \softmax(\vell^{c})$ \Comment{unguided}
    \EndIf
    \State $a_k \leftarrow (t_{k+1} - t_k)/(1 - t_k)$
    \State redraw each $x^i \sim \vpi^i$ w.p.\ $a_k$
\EndFor
\State \Return $\gG$
\end{algorithmic}
\end{algorithm}
\end{minipage}
\end{figure}

%% file: figures/fig2a_ribbon.tex
\definecolor{digSil}{HTML}{1A1A1A}
\begin{tikzpicture}[
  x=1cm, y=1cm,
  lbl/.style   = {font=\rmfamily\fontsize{6.4}{7.4}\selectfont, text=digInk},
  flow/.style  = {-{Straight Barb[length=2.6pt,width=3.0pt]}, draw=black, line width=0.55pt},
  cflow/.style = {-{Straight Barb[length=2.2pt,width=2.6pt]}, draw=black,
                  line width=0.65pt, dash pattern=on 1.8pt off 1.2pt},
  gdot/.style  = {circle, inner sep=0pt, minimum size=2.3pt},
]

\def\rowT{5.55}   %
\def\rowZ{3.45}   %
\def\ymid{4.50}
\def\hh{0.85}     %
\def\matX{6.34}   %
\def\matP{0.32}   %
\def\cardX{11.56} %

\def\card#1#2#3#4#5{%
  \fill[#5!35, rounded corners=2.4pt] ({#1+0.06},{#2-0.06}) rectangle ({#3+0.06},{#4-0.06});
  \filldraw[draw=black, line width=0.5pt, fill=#5!8, rounded corners=2.4pt]
    ({#1},{#2}) rectangle ({#3},{#4});
}
\def\qdcard#1#2#3#4#5#6{%
  \card{0.10}{#1-\hh}{2.20}{#1+\hh}{#2}
  \pgfmathsetmacro\pit{(1.86-0.0775*#5)/#4}
  \pgfmathsetmacro\ins{min(0.00705,\pit*0.12)}
  \pgfmathsetmacro\bs{#1-\hh+0.20}
  \draw[black, line width=0.3pt] (0.22,\bs) -- (2.08,\bs);
  \foreach \f/\fp/\g/\e [count=\i from 0] in #3{
    \pgfmathsetmacro\xl{0.22+\i*\pit+\g*0.0775}
    \pgfmathsetmacro\xr{\xl+\pit}
    \pgfmathsetmacro\bh{max(\f*1.16,0.0247)}
    \pgfmathsetmacro\bp{\fp>0 ? max(\fp*1.16,0.0247) : 0}
    \fill[#2] ({\xl+\ins},\bs) rectangle ({\xr-\ins},{\bs+\bh});
    \draw[digSil, line width=0.8pt, line join=miter]
      (\xl,{\bs+\bp}) -- (\xl,{\bs+\bh}) -- (\xr,{\bs+\bh});
    \ifnum\e=1
      \draw[digSil, line width=0.8pt, line join=miter] (\xr,{\bs+\bh}) -- (\xr,\bs);
    \fi
  }
  \node[lbl, anchor=north east] at (2.12,{#1+\hh-0.04}) {$#6$};
}
\def\tri#1#2#3#4#5#6{%
  \foreach \v [count=\i from 0] in {#5}{
    \pgfmathtruncatemacro\rr{\i<3 ? 0 : (\i<5 ? 1 : 2)}
    \pgfmathtruncatemacro\cc{\i<3 ? \i : (\i<5 ? \i-2 : 2)}
    \pgfmathsetmacro\cx{#1+\cc*#3+0.03*#3}
    \pgfmathsetmacro\cy{#2+0.5*#3-\rr*#3+0.03*#3}
    \ifnum\v=1
      \filldraw[draw=black, line width=0.5pt, fill=digInk]
        (\cx,\cy) rectangle ({\cx+0.94*#3},{\cy+0.94*#3});
    \else
      \filldraw[draw=black, line width=0.5pt, fill=#4!58]
        (\cx,\cy) rectangle ({\cx+0.94*#3},{\cy+0.94*#3});
    \fi
  }
  \node[font=\fontsize{8}{9}\selectfont, text=digInk, anchor=south west, inner sep=0.8pt]
    at ({#1+0.03*#3},{#2+1.47*#3+0.02}) {$#6$};
}
\def\strip#1#2{%
  \begin{scope}[transparency group, opacity=0.45]
    \card{6.14}{#1-\hh}{13.62}{#1+\hh}{#2}
  \end{scope}
}
\def\paths#1#2#3#4{%
  \pgfmathsetmacro\px{\matX+2.97*\matP}
  \pgfmathsetmacro\pw{\cardX-\px}
  \begin{scope}[shift={(\px,#1)}, xscale=\pw, yscale=0.72]
    \foreach \g in {#3}{
      \draw[black!20, line width=0.35pt, line join=round] \g;
    }
    \foreach \g in {#2}{
      \draw[black!38, line width=0.45pt, line join=round] \g;
    }
    \draw[black, line width=0.9pt, line join=round] #4;
  \end{scope}
}

\qdcard{\rowT}{digTree}{\cdStepTree}{\cdOwnNTree}{\cdOwnGapsTree}{\vq_1}
\qdcard{\rowZ}{digZINC}{\cdStepZinc}{\cdOwnNZinc}{\cdOwnGapsZinc}{\vq_n}

\card{2.56}{2.60}{5.76}{6.40}{dig2cond}
\node[lbl, anchor=north west, text=digInk,
      font=\fontsize{7.4}{8.4}\selectfont] at (2.64,6.34) {$\vtheta$};

\node[draw=black, fill=dig2cond!10, line width=0.6pt, rounded corners=1.5pt,
      inner sep=2.0pt, font=\fontsize{6.6}{7.6}\selectfont] (cvec) at (4.16,5.98)
     {$\vc=[\,\vz_d\,\Vert\,\vs\,]$};
\draw[black, line width=0.65pt] (4.16,5.78) -- (4.16,5.58);
\draw[black, line width=0.65pt] (3.00,5.58) -- (5.46,5.58);
\foreach \x in {3.00,3.82,4.64}{ \draw[cflow] (\x,5.58) -- (\x,4.92); }
\draw[cflow] (5.46,5.58) -- (5.46,4.76);

\foreach \k/\lx in {1/3.00, 2/3.82, 3/4.64}{
  \node[draw=black, line width=0.5pt, fill=digFillSoft, rounded corners=1.8pt,
        minimum width=0.58cm, minimum height=0.74cm, inner sep=1pt,
        font=\fontsize{7.2}{8.2}\selectfont, text=digInk] (L\k) at (\lx,\ymid) {$L_{\k}$};
}

\def\readout#1#2{%
  \filldraw[draw=black, line width=0.5pt, fill=digFillSoft]
    ({#1-0.26},{#2+0.22}) -- ({#1+0.26},{#2+0.22}) -- ({#1+0.13},{#2-0.22})
    -- ({#1-0.13},{#2-0.22}) -- cycle;
}
\readout{3.00}{3.40}
\readout{4.64}{3.40}
\node[circle, draw=black, fill=white, line width=0.75pt, inner sep=0.9pt,
      font=\fontsize{6.8}{7.6}\selectfont, text=digInk] (gate) at (5.46,\ymid) {$\vw$};
\draw[black, line width=0.5pt] (3.00,3.18) -- (3.00,2.92) -- (5.46,2.92) -- (5.46,4.28);
\draw[black, line width=0.5pt] (4.64,3.18) -- (4.64,3.04) -- (5.46,3.04);
\node[gdot, fill=black, minimum size=1.9pt] at (5.46,3.04) {};

\strip{\rowT}{digTree}
\strip{\rowZ}{digZINC}
\tri{\matX}{\rowT}{\matP}{digTree}{0,1,0,0,0,0}{\sV_1}
\tri{\matX}{\rowZ}{\matP}{digZINC}{0,0,0,1,0,0}{\sV_n}
\foreach \c/\yy/\f in {digTree/\rowT/tree, digZINC/\rowZ/zinc}{
  \card{\cardX}{\yy-0.72}{13.50}{\yy+0.72}{\c}
  \node[inner sep=0pt] at (12.53,\yy)
    {\includegraphics[width=1.84cm, height=1.36cm, keepaspectratio]{assets/code_fig2_\f.pdf}};
}
\paths{\rowT}{\trajTreeGreyA,\trajTreeGreyB,\trajTreeGreyC}
  {\trajTreeFaintA,\trajTreeFaintB,\trajTreeFaintC,\trajTreeFaintD}{\trajTreeBlack}
\paths{\rowZ}{\trajZincGreyA,\trajZincGreyB,\trajZincGreyC}
  {\trajZincFaintA,\trajZincFaintB,\trajZincFaintC}{\trajZincBlack}

\foreach \ellx in {1.15, 9.88}{%
  \foreach \dy in {-0.09, 0, 0.09}{ \fill[digInk] (\ellx,{4.47+\dy}) circle (0.028); }
}

\draw[black, line width=0.55pt] (2.20,\rowT) -- (2.34,\rowT) -- (2.34,\rowZ) -- (2.20,\rowZ);
\draw[flow] (2.34,\ymid) -- (L1.west);
\draw[flow] (L1.east) -- (L2.west);
\draw[flow] (L2.east) -- (L3.west);
\draw[flow] (3.00,{\ymid-0.37}) -- (3.00,3.62);
\draw[flow] (4.64,{\ymid-0.37}) -- (4.64,3.62);
\draw[black, line width=0.55pt] (5.64,\ymid) -- (5.94,\ymid);
\draw[black, line width=0.55pt] (5.94,\rowZ) -- (5.94,\rowT);
\draw[flow] (5.94,\rowT) -- (6.30,\rowT);
\draw[flow] (5.94,\rowZ) -- (6.30,\rowZ);
\draw[flow, rounded corners=3pt]
  (13.62,\rowT) -- (13.78,\rowT) -- (13.78,2.20) -- (-0.08,2.20) -- (-0.08,\rowT) -- (0.10,\rowT);
\draw[black, line width=0.55pt] (13.62,\rowZ) -- (13.78,\rowZ);
\draw[flow] (-0.08,\rowZ) -- (0.10,\rowZ);
\node[lbl, fill=white, inner sep=1.4pt] at (6.85,2.20) {$\times T$};

\end{tikzpicture}

%% file: sections/benchmark.tex
\section{Experiments}
\label{sec:experiments}

In order to assess the difficulty of the multi-domain graph generation problem and determine the effectiveness of DiG$^{2}$, we conduct experiments to answer the following research questions: \textbf{(RQ1):} Are domain-aware graph generative models and DiG$^{2}$ capable of generating multiple graph domains at once? \textbf{(RQ2):} What components of DiG$^{2}$ are necessary for successful multi-domain graph generation? \textbf{(RQ3):} Can DiG$^{2}$ interpolate for successful \emph{de novo} molecule generation?

\subsection{Main Results}
\label{sec:benchmark}

Within the following section, we answer \textbf{RQ1} by benchmarking the capabilities of current graph generation models and DiG$^{2}$ to assess the difficulty of the multi-domain graph generation problem.

\textbf{Setup} We evaluate on five domains with  disjoint semantics families: structural domains (\textsc{Planar}, \textsc{SBM}, \textsc{Tree}) and molecular domains
(\textsc{QM9}, \textsc{ZINC}). All domains are encoded into a global vocabulary of $11$ distinct node and
$5$ edge classes. This
shared encoding makes single jointly-trained models possible, and makes domain conditioning failures decidable by inspection.

\textbf{Arms.} All benchmarked models are trained on the five dataset corpus under one shared schedule of $2.5$M steps (10-day max training budget, Appendix~\ref{app:baselines}). All models are anchor fine-tuned, $25$K steps per wave in ordered waves for each structural domain: Tree $\rightarrow$ SBM $\rightarrow$ Planar, for $75$K steps total (Algorithm~\ref{alg:anchor}, Appendix~\ref{app:impl}). Every number in Table~\ref{tab:benchmark} is reported after fine-tuning, and every arm is fine-tuned on the same randomly-generated pools of $8{,}192$-graphs for each structural domain (App.~\ref{app:finetune}). 1-hot embeddings are projected into node- and edge-channels of single-domain models. The $\sV_d$ mask is applied to every arm at every training and sampling step. Per-domain priors are provided for DiGress and DeFoG, which share DiG$^{2}$'s denoising Graph Transformer backbone. Classifier-free guidance is given to DiGress, DeFoG, LGGM and LGD at the same weights as DiG$^{2}$ ($g=3.0$ on SBM, $2.0$ elsewhere). \textsc{LGGM} integrates 1-hot labels and native text-prompt conditioning~\citep{devlin2019bert}. \textsc{G2PT} conditions natively on a per-domain prefix token. Every arm is re-parented onto a single configuration, so the dataset, splits, optimizer schedule and evaluation harness are identical across generative models. Backbones, parameter counts and method-specific hyperparameters are in Appendix~\ref{app:baselines}.

\textbf{Metrics.} For the structural regime, we follow SPECTRE's standard~\citep{martinkus2022spectre} and report: \textit{V.U.N.} on planarity, acyclicity, or block-structure and \textit{average MMD Ratios} across degree, clustering, orbit, spectral and wavelet distributions. For the molecular regime, we report Validity and Fr\'echet ChemNet Distance (FCD)~\citep{preuer2018fcd}. Models are chosen based on the best accuracy metrics during validation. Uniqueness and Novelty are reported in Table~\ref{tab:novelty}.

\begin{table}[h!]
\centering
\small
\begin{tabular}{lccccc}
\toprule
&  \multicolumn{3}{c}{Structural (V.U.N.\ $\uparrow$ / Ratio $\downarrow$)} & \multicolumn{2}{c}{Molecular (Valid. $\uparrow$\,/\,FCD $\downarrow$)} \\
\cmidrule(lr){2-4}\cmidrule(lr){5-6}
Model  & Planar & SBM & Tree & QM9  & ZINC \\
\midrule
DiG$^{2}$             & \textbf{100}/ \textbf{2.3} & \textbf{85} / \textbf{1.8} & \textbf{97.5} / \textbf{1.56} & \textbf{99.92} / \textbf{0.29} & \textbf{95.46} / \textbf{2.7} \\
\midrule
G$^2$PT              & 2.5/ 7.5 & \underline{73.1}/4.92 & \underline{87.4}/\underline{1.57} & 90.6 / 0.55 & 72.6 / 6.4 \\
DeFoG$^{\dagger}$    & \underline{95.0}/\underline{4.3} & 5.0/6.9 & 62.5/2.35 & 90.21 / 0.49 & \underline{92.93} / \underline{3.4} \\
DiGress$^{\dagger}$      & 10.0/7.8 & 37.5/ \underline{3.04} & 67.5/1.81 & \underline{95.43} / \underline{0.4} & 87.1 / 4.7 \\
HOG-Diff$^{\dagger}$  & 0.0/ 29.5 & 25.0/ 4.72 & 52.5/3.73 & 93.20 / 1.5 & 86.38 / 4.0 \\
GraphBSI$^{\dagger}$  & 0.0/ 30.1 & 22.5/ 3.56 & 72.5/1.95 & 43.50 / 9.4 & 42.19 / 25.2 \\
LGGM      & 0.0/ 76 & 12.5/ 9.1 & 0.0 / 8.78 & 86.3 / 0.75 & 85.2 / 8.55 \\
Pard$^{\dagger}$  & 0.0/ 298 & -- & 47.5 / 4.96 & 47.48 / 10.1 & 31.4 / 25.2 \\
GraphRNN$^{\dagger}$  & 0.0/ 156 & 10.0 / 22.58 & 22.5 / 2.76 & 47.53 / 10.5 & 32.83 / 25.6 \\
LGD$^{\dagger}$  & 0.0/ 2591 & 0.0 / 253 & 0.0 / 375 & 34.2 / 19.1 & 37.89 / 26.1 \\
SPECTRE$^{\dagger}$  & 0.0/ 2766 & 0.0 / 374 & 0.0 / 461 & 17.3 / -- & 9.51 / -- \\

\bottomrule
\end{tabular}
\vskip -0.75em
\caption{Multi-domain generation results across two distinct structural families. $^{\dagger}$ denotes a baseline given multi-domain conditioning uplift. Sampling is fixed to $T{=}200$ steps for all diffusion baselines.}
\label{tab:benchmark}
\end{table}

\textbf{DiG$^2$ succeeds at multi-domain generation.}
As shown in Table~\ref{tab:benchmark}, DiG$^2$ achieves 100.0\%, 85.0\%, and 97.5\% V.U.N. on
PLANAR, SBM, and TREE, and reaches 99.92\% Validity on QM9 and 95.46\% on ZINC, \textbf{performing the best on every metric}. The current benchmark therefore indicates generation across multiple domains
requires sampling paths and internal representations to
remain coordinated with the requested domain.

\textbf{Domain conditioning does not solve multi-domain generation.}
In contrast, Table~\ref{tab:benchmark} also exposes a substantial gap for single-domain models, even after the models are provided domain-conditioning. For example, DeFoG retains 95.0\% planarity V.U.N., but falls to
5.0\% on SBM.
G$^2$PT is substantially stronger on SBM and TREE (73.1\% and 87.4\%,
respectively), yet falls to 2.5\% on PLANAR. More specialized graph
generators are brittle: HOG-Diff, GraphBSI, PARD, GraphRNN, and LGD
all fail to recover at least one structural family. SPECTRE suffers mode collapse under the joint training protocol. Therefore, domain conditioning alone is insufficient for enforcing domain-separable generative dynamics.

%% file: sections/ablation.tex
\subsection{Ablation Study}
\label{sec:ablation}

To answer \textbf{(RQ2)} and isolate where separate DiG$^2$ components are necessary for multi-domain graph generation. We conduct a leave-one-component-out ablation beginning from the full DiG$^2$ configuration. All results can be found in Table~\ref{tab:ablation}.

\textbf{Setup} We ablate the following model components: (i) the
EMA prototype embedding~\citep{morales2024exponential}, forcing the denoiser to rely on 1-hot
conditioning alone; (ii) per-domain corruption priors, replacing
$q_d$ with one pooled prior while keeping the label, prototype and $\sV_d$ mask; (iii) domain-restricted cross-entropy support,
allowing logits over the full global vocabulary; and (iv) multi-depth
readout, retaining only terminal-layer predictions. This isolates representation, endpoint, vocabulary-support,
and computation-depth effects, respectively.

\begin{table}[h!]
\centering
\small
\begin{tabular}{lccccc}
\toprule
&  \multicolumn{3}{c}{Structural (V.U.N. $\uparrow$ / Ratio $\downarrow$)} & \multicolumn{2}{c}{Molecular (Valid.} $\uparrow$ /\,FCD $\downarrow$) \\
\cmidrule(lr){2-4}\cmidrule(lr){5-6}
DiG$^{2}$  & Planar & SBM & Tree & QM9  & ZINC \\
\midrule
\textit{Full model (\Tabref{tab:benchmark})}     & 100/2.3 & 85 / 1.8 & 97.5 / 1.56 & 99.92 / 0.29 & 95.46 / 2.7 \\
\midrule
\textit{w/o EMA prototype}     & 77.5/7.65 & 42.5 / 4.82 & 92.5 / 1.79 & 92.48 / 1.83 & 91.92 / 3.37 \\
\textit{w/o per-domain priors}      & 82.5/4.29 & 70.0 / 3.95 & 95.0 / 1.73 & 93.86 / 1.48 & 92.42 / 2.93 \\
\textit{w/o per-domain CE}      & 90.00/6.37 & 77.5 / 3.48 & 95.0 / 1.69 & 94.57 / 1.73 & 94.42 / 2.89 \\
\textit{w/o per-layer depth}      & 95.0/2.95 & 72.5 / 5.13 & 90.0 / 2.78 & 98.80 / 0.79 & 95.04 / 3.01 \\
\midrule
\bottomrule
\end{tabular}
\vskip -0.75em
\caption{Ablation Study Results. As shown, all of the following ablated mechanism are necessary for optimal multi-domain generation: per-domain corruption priors $\vq_d$, support restriction
$\upharpoonright_{\sV_d}$, multi-depth readout from \Figref{fig:ribbon}, and the EMA embedding in \Figref{fig:cnd_layer}.}
\label{tab:ablation}
\end{table}

\textbf{Results} Table~\ref{tab:ablation} demonstrates that \emph{each of DiG$^{2}$'s components are critical} for successful multi-domain graph generation. In particular, the EMA prototype has the strongest detrimental effect, with $-22.5\%$ PLANAR and $-42.5\%$ SBM V.U.N., which is indicative of the difficulty for internal model representations to separate the two, even with the remaining per-domain components. Separately, the lack of complete collapse across every test suggests that \Eqref{eq:reverse} maintains enough model capacity to resist mode collapse during the denoising process. In-depth analysis of separate components is provided in Appendix~\ref{app:extra_ablation}.

%% file: sections/scaling.tex
\subsection{Scaling the Number of Domains}
\label{sec:scaling}

A natural question is whether training on more than 5 domains aids performance. To quantify this effect, we analyze DiG$^2$ and LGGM performance after training on the 12 LGGM domains. Given LGGM's protocol is structure-only, it embeds cleanly
into DiG$^{2}$'s structural-regime vocabulary. After 250k training steps, DiG$^{2}$ performs better on 14 of the 15 domains and is within $0.04$ on Econ, leading to a $7.32$x improved average MMD ratio over LGGM ($Table~\ref{tab:lggm15}$).
The gap is widest on structural validity, where DiG$^{2}$ reaches $92.5\%$ and $97.5\%$ on Planar and Tree relative to LGGM never bumping past $0\%$.

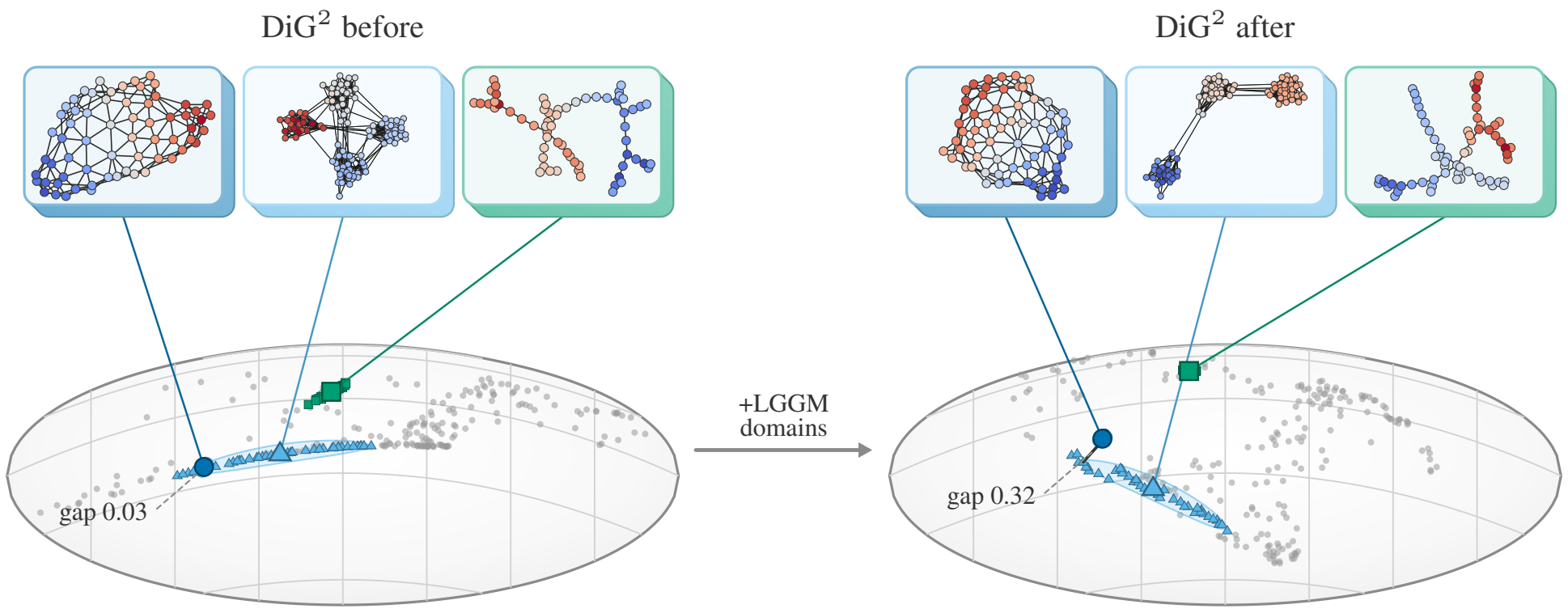
\begin{figure}[h!]
\centering
\input{figures/fig_domes_final}
\caption{ \textbf{Wider training separates conditioning signals.} DiG$^{2}$'s
structural-encoder space before and after appending LGGM domains (shown in gray) for training with the structural regime: \textcolor{digPlanar}{\rule{0.55em}{0.55em}} Planar, \textcolor{digSBM}{\rule{0.55em}{0.55em}} SBM,
\textcolor{digTree}{\rule{0.55em}{0.55em}} Tree. The 'gap' is measured between the closest Planar/SBM pair.}
\label{fig:domes}
\end{figure}

\begin{table}[h!]
\centering
\scriptsize
\setlength{\tabcolsep}{2.5pt}
\begin{tabular}{lcccccccccccccccc}
\toprule
Model & FB & ASN & Email & Web & Road & Power & Bio & Econ & RT & Col & Eco & Cit. & Planar & SBM & Tree & Avg.\ $\downarrow$ \\
\midrule
DiG$^{2}$ & \textbf{1.62} &\textbf{2.08} & \textbf{1.50} & \textbf{2.18} & \textbf{3.70} & \textbf{1.48} & \textbf{1.39} & 1.15 & \textbf{2.37} & \textbf{3.31} & \textbf{2.86} & \textbf{2.07} & \textbf{92.5}\% / \textbf{1.20} & \textbf{85}\% / \textbf{2.30} & \textbf{97.5}\% / \textbf{1.92} & \textbf{2.08} \\
LGGM & 2.15 & 5.01 & 2.27 & 3.05 & 4.93 & 3.78 & 1.63 & \textbf{1.11} & 21.94 & 13.58 & 4.52 & 2.81 & 0.0\% / 137 & 5.0\% / 11.3 & 0.0\% / 12.94 & 15.23 \\
\bottomrule
\end{tabular}
\vskip -0.5em
\caption{LGGM and DiG$^{2}$ trained on all 15 domains (LGGM's 12 plus Planar, SBM, Tree). Cells are average
MMD ratios (closer to $1.0$ is better). Planar, SBM, and Tree also provide V.U.N. (\%). The final
column is the MMD ratio averaged over all 15 domains.}
\label{tab:lggm15}
\end{table}

Figure~\ref{fig:domes} visualizes DiG$^2$'s structural encoding for all domains before and after appending the LGGM domains to the DiG$^2$ benchmarking protocol. LGGM only encodes label, so it's structure cannot be visualized. The right-side of Figure~\ref{fig:domes} indicates Planar and SBM
centroids for DiG$^2$ hardly shift, remaining within their original centroid shapes ($\times0.99$). However, the gap between SBM and Planar widens $10.7\times$. So, SBM leaves the same-axis as Planar, indicating DiG$^2$ can scale while improving between-domain separation while maintaining within-domain consistency. During sampling, we further reduce DiG$^2$'s per-domain structural prototype to a per-domain mean, SBM V.U.N. collapses by $40\%$, indicating that DiG$^2$'s structural encoding is necessary to maintain high-quality performance. Additional details are included in App.~\ref{app:repr}.

%% file: figures/fig_domes_final.tex
{\rmfamily\fontsize{6.5}{7.8}\selectfont%
\def\svgwidth{\linewidth}%
\IfFileExists{figures/fig_domes_final.pdf_tex}%
  {\input{figures/fig_domes_final.pdf_tex}}%
  {\input{fig_domes_final.pdf_tex}}%
}%

%% file: sections/interpolation.tex
\subsection{$\alpha$-Interpolation for Zero-Shot Molecular Extrapolation}
\label{sec:interp}
The previous studies test whether generation modes remain separated. However, this study is designed to answer \textbf{RQ3}. A complementary question on whether DiG$^{2}$'s learned modes can be interpolated to generate novel, yet realistic samples, for \emph{de novo} molecule generation.


\textbf{$\alpha$-Interpolation.}
In order to determine if DiG$^{2}$ can traverse the space between domains to generate unseen molecular
corpora, 
we interpolate domain conditioning via linear blending of the domain-embedding and prototype rows. The blended row are subsequently passed through a single conditional forward pass. The prior, guidance weight and union vocabulary mask are held fixed. Notably, our $\alpha$-Interpolation method does not require explicit training~\citep{zhang2023shiftddpms, choi2024simple}. For $\alpha \in [0,1]$, with $\alpha=0$ denoting QM9 and
$\alpha=1$ denoting ZINC, we construct an interpolated conditioning state,
\begin{align}
z_\alpha &= (1-\alpha)\,z_{\mathrm{QM9}}+\alpha\,z_{\mathrm{ZINC}},\\
s_{\mathrm{proto},\alpha} &= (1-\alpha)\,s_{\mathrm{proto,QM9}}+\alpha\,s_{\mathrm{proto,ZINC}},
\end{align}
and assemble $c_\alpha=[z_\alpha\Vert s_{\mathrm{proto},\alpha}]$. Sampling then uses the same closed-form reverse interpolation as Prop.~\ref{prop:bridge}, but masks logits to the union molecular support
$V_\alpha=V_{\mathrm{QM9}}\cup V_{\mathrm{ZINC}}$, held fixed across $\alpha$. No GuacaMol or MOSES graph is used for parameter updates.

The prior is held at the ZINC endpoint for every $\alpha$, so the blend acts through the conditioning alone. QM9 and ZINC provide a controlled setting, both datasets are composed of similar vocabularies, differing mostly in size and
chemical distribution. Therefore, any extrapolated molecule close to a given target tests how DiG$^{2}$'s $\alpha$-interpolation discovers viable molecules outside its training distribution.

\textbf{Baselines.}
We compare DiG$^{2}$ against three molecular generators which we treat to expose a QM9$\leftrightarrow$ZINC
axis.
LGGM~\citep{wang2025large} is given a per-domain one-hot prompt table, and $\alpha$ blends the two embeddings. GenMol~\citep{lee2025genmol}, a discrete molecule generation model, fine-tuned with QM9 and ZINC domain tokens for $12,000$ steps, and $\alpha$ blends their embeddings. DemoDiff~\citep{liu2026graph}, a 700M-parameter molecule generation foundation model integrates in-context demonstration molecules drawn from QM9 and ZINC in
proportion $(1-\alpha):\alpha$. Given space limitations, tests with LGGM and further details are provided in Appendix~\ref{app:zero_shot}.

\textbf{Zero-shot interpolation.}
Every model is evaluated on the same fixed grid,
$\alpha\in\{0,0.5,0.75,0.85,0.95,1\}$, three seeds per cell and 2{,}000 attempts,
on GuacaMol molecules of 10--14 heavy atoms. Target metrics never tune parameters or guidance.

\input{figures/caption_property_target_notes}
\begin{figure}[t!]
    \centering
    \begin{subfigure}[b]{0.5\textwidth}
        \centering
        \includegraphics[width=\textwidth]{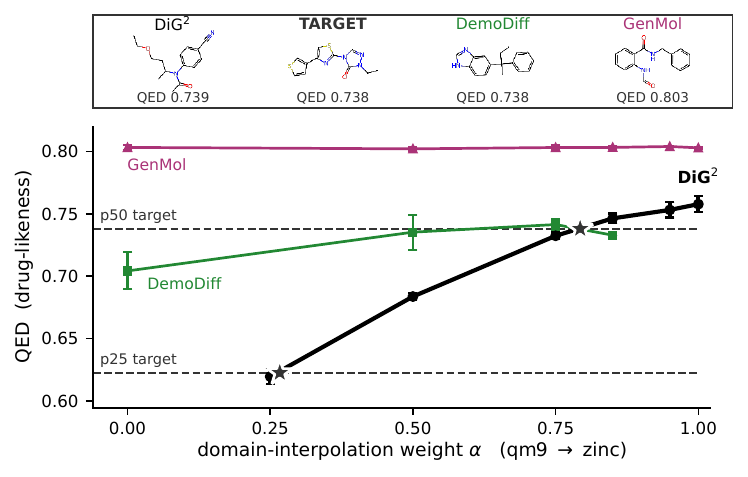}
        \caption{\textbf{QED as an interpolation target.}}
        \label{fig:property_target}
    \end{subfigure}
    \hfill
    \begin{subfigure}[b]{0.48\textwidth}
        \centering
        \includegraphics[width=\textwidth]{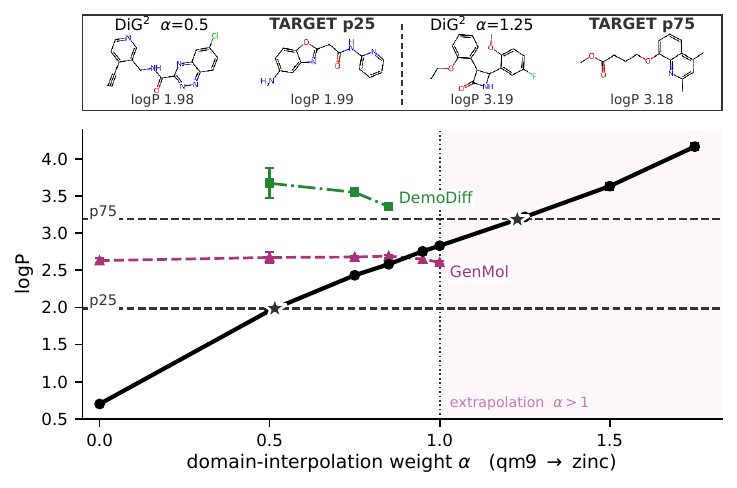}
        \caption{\textbf{logP as an extrapolation target.}}
        \label{fig:logp_target}
    \end{subfigure}
    \caption{\textbf{(\subref{fig:property_target})} GuacaMol 10-14 band. QED range covered: DiG$^2$ $= 0.138$, DemoDiff $= 0.037$, GenMol $= 0.002$, meaning GenMol's axis is inert, and DemoDiff's moves but cannot descend to p25. Per-cell values are in Table~\ref{tab:property_target_raw}. \textbf{(\subref{fig:logp_target})} MOSES 20-27 band. Each arm is swept through \emph{its own} native mechanism. logP range covered: DiG$^{2}$ $= 3.46$, DemoDiff $= 0.31$ GenMol $= 0.08$, indicating GenMol and DemoDiff are inert for p25 and p75 targets. Per-cell values are in Table~\ref{tab:logp_target_raw}.}
    \label{fig:property_logp}
\end{figure}



Despite Figure~\ref{fig:property_target} being confined to $\alpha\in[0,1]$, DiG$^{2}$ interpolates between both the $p25$ and $p50$ QED target~\citep{bickerton2012quantifying}, while DemoDiff can only reach the $p50$ target and GenMol overshoots it. As shown in App.~\ref{app:interp_test}, we observe the blend yields genuine intermediates from the endpoints, on a Fisher axis ($\Delta\mathrm{BIC}>0$ in 12/12 interior cells, all without the need for a trained classifier~\citep{deschenaux2024going}. Given DiG$^{2}$'s capacity to interpolate \emph{between} endpoints. We now consider the question, can DiG$^{2}$ \emph{extrapolate} to discover realistic samples from a wider $\alpha$ target?

\subsection{Exterior $\alpha$: interpolate-to-extrapolate}
\label{sec:extrap}

\textbf{Interpolate-to-Extrapolate.} Because $\alpha$ blends embedding rows, it extends beyond $\alpha=1$ without property labels, whereas DemoDiff's demonstration proportion is confined to $[0,1]$. GenMol's and LGGM's blended rows admit $\alpha>1$, but GenMol's moves logP by only $0.08$ inside $[0,1]$ and LGGM's degrades validity, leaving neither usable for extrapolation.  As shown in Figure~\ref{fig:logp_target}, extending the MOSES
20--27 sweep requires the interpolation to traject past ZINC's  endpoint ($a=1$). The p75 logP target is met only at interpolation values of
$\alpha>1$. DiG$^{2}$ covers $10$ times the logP interpolation range of DemoDiff, indicating capacity to extrapolate beyond ZINC's logP distribution, relative to other models. As measured against the ZINC reference set, median logP rises monotonically from $2.89$ at $\alpha=1.00$ to $4.82$ at $\alpha{=}2.00$, where $59.2\%$ of samples sit above ZINC's own $95$th percentile. Additional $\alpha$-extrapolation details are located within App.~\ref{app:extrap_test} and App.~\ref{app:zero_shot}.

%% file: figures/caption_property_target_notes.tex
\newcommand{\ptBand}{15--19}
\newcommand{\ptRefN}{7{,}860}
\newcommand{\ptDigSpan}{0.1438}
\newcommand{\ptCeiling}{0.7645}
\newcommand{\ptFloorCal}{0.0064}
\newcommand{\ptSegLo}{0.25}
\newcommand{\ptSegHi}{1.25}
\newcommand{\ptTargetPlo}{0.6225}
\newcommand{\ptAlphaStarPlo}{0.2667}
\newcommand{\ptErrPlo}{0.0051}
\newcommand{\ptTargetPmid}{0.7380}
\newcommand{\ptAlphaStarPmid}{0.7930}
\newcommand{\ptErrPmid}{0.0026}
\newcommand{\ptTargetPhi}{0.8222}
\newcommand{\ptAlphaStarPhi}{\textsc{n/a}}
\newcommand{\ptErrPhi}{\textsc{n/a}}
\newcommand{\ptGenmolSpan}{0.0489}
\newcommand{\ptGenmolBlind}{0.7746}
\newcommand{\ptSpanDigTwo}{0.1382}
\newcommand{\ptSpanDemodiff}{0.0372}
\newcommand{\ptSpanGenmol}{0.0019}
\newcommand{\ptDemodiffSpan}{\textsc{incomplete}}
\newcommand{\ptDemodiffStatus}{partial}
\newcommand{\ptDemodiffCells}{13/15}

%% file: sections/conclusion.tex
\section{Conclusion}

Within this work, we formalize the multi-domain graph generation problem as a challenging task, requiring the separation of internal representations from disjoint vocabularies and generating domain-specific structure. To address this problem, we introduce DiG$^{2}$, a graph flow-matching model integrating domain-conditioning to bridge between domain-specific priors for successful multi-domain graph generation. We also uncover DiG$^{2}$'s ability to $\alpha-$interpolate across domains, enabling the model to reach outside the scope of its original training distribution without property supervision. We believe the multi-domain graph generation problem is a promising frontier for generative modeling, enabling future work for exciting applications of graph generative systems.

%% file: sections/appendix.tex
\input{sections/appendix/notation}

\input{sections/limitations}

\input{sections/appendix/implementation}

\input{sections/appendix/baselines}

\input{sections/appendix/ablation}

\input{sections/appendix/zero_shot}
\newpage
\input{sections/appendix/initial}

\newpage
{
\section{Additional Results}
\label{app:revision}

This section consists of additional experiments and analysis which did not fit within the main text.
}

\input{sections/appendix/revision_finetune}

\input{sections/appendix/revision_baselines}

\input{sections/appendix/revision_scaling}

\input{sections/appendix/revision_repr}

\input{sections/appendix/revision_protocol}

\input{sections/appendix/revision_interp}

%% file: sections/appendix/notation.tex
\subsection{Proofs}

\label{app:proofs}

\paragraph{Roadmap.} The following proof answers four questions about the sampler of
\Secref{sec:form-bridge} and \Secref{sec:form-sample}, in order. \emph{(i) Is the reverse step exact, and
is it really shared by all domains?} Lemmas~\ref{lem:coupling}--\ref{lem:transport} build up to the proof
of Proposition~\ref{prop:bridge}. \emph{(ii) Where can the domain enter the prediction, and how much does
the noisy graph already reveal about it?} Lemma~\ref{lem:domain} and Table~\ref{tab:domain-separation}.
\emph{(iii) Are vocabulary masking and guidance safe to combine with the bridge?} Lemma~\ref{lem:mask} and
Corollary~\ref{cor:guided}, which justify how \Eqref{eq:cfg} is implemented. \emph{(iv) What does it cost
to replace the exact posterior with a network?} Proposition~\ref{prop:error}. Most proofs are short
because a single picture, the reveal coupling of Lemma~\ref{lem:coupling}, does the work; we state it first
and refer back to it throughout.

\paragraph{Setup and premises.}
A graph with $n$ nodes is the collection of classes $\gG = (x^i)_{i \in \sI_n}$ over its entries
$\sI_n = [n] \cup \{\{u,v\} : 1 \le u < v \le n\}$, of which there are $N_n = n + \binom{n}{2}$. A node
entry takes values in the global node vocabulary ($11$ classes) and a pair entry in the global edge
vocabulary ($5$ classes, including ``no edge''). The node count $n$ is drawn from domain $d$'s size
distribution before $t = 0$ and held fixed, so every statement below is conditional on $(d, n)$.
The domain reaches the sampler only through rows of tables selected by the same one-hot
$\ve_d \in \{0,1\}^{|\sD|+1}$, whose last coordinate is the guidance null token $\varnothing$
(Table~\ref{tab:domain-roles}): the prior $\vq_d^i = \mathbf{Q}_{\mathrm{pr}}^{\top}\ve_d$ (its node or edge
row), the vocabulary mask $\vm_d^i = \mathbf{M}_{\mathrm{voc}}^{\top}\ve_d$ with support
$\sV_d^i = \{x : \vm_d^i(x) = 1\}$, and the embedding $\vz_d = \mathrm{MLP}(\mE^{\top}\ve_d)$ inside
$\vc = [\vz_d \Vert \vs]$. We use three premises, all of which hold in our implementation:
\begin{itemize}
\item[(A1)] $\mathrm{supp}\,\vq_d^i \subseteq \sV_d^i$: the prior lies inside the vocabulary;
\item[(A2)] $\mathrm{supp}\,P_{\mathrm{data}}(\cdot \mid d) \subseteq \prod_i \sV_d^i$: $\sV_d$ contains every class domain $d$ uses;
\item[(A3)] both guidance branches are masked with the same $\vm_d$ before they are combined.
\end{itemize}
Each premise closes one gap. Without (A1), the initial graph $\gG_0$ could already contain a class that
domain $d$ never uses, and no later step would be obliged to remove it. Without (A2), the mask would
forbid classes that real data of domain $d$ contains, so the exact posterior would place mass where the
masked network cannot, and masking would add error instead of removing it. Without (A3), guidance can
undo or even invert a mask applied to only one branch (Lemma~\ref{lem:mask}(c)).
We write $\Pi_d(\gG_1 \mid \gG_t) = P(\gG_1 \mid \gG_t, d)$ for the exact posterior of
Definition~\ref{def:forward}, $\pi_d^i(x \mid \gG_t) = P(x_1^i = x \mid \gG_t, d)$ for its entry
marginals, and $B_{t\to s}(\cdot \mid x_t, x_1) = a\,\delta_{x_1} + (1-a)\,\delta_{x_t}$, with
$a = a(t,s) = \frac{s-t}{1-t}$, for the conditional bridge kernel of one entry. $\Pi_d$ does not
factorize over entries; the prior, the corruption given $\gG_1$, and the network's prediction do.

\begin{table}[h]
\centering
\small
\caption{Where the domain one-hot $\ve_d$ enters the sampler, and what each entry point guarantees.}
\label{tab:domain-roles}
\begin{tabular}{p{0.23\linewidth} p{0.09\linewidth} p{0.22\linewidth} p{0.3\linewidth}}
\toprule
Object & From $\ve_d$ & Enters & Guarantee \\
\midrule
prior $\vq_d = \mathbf{Q}_{\mathrm{pr}}^{\top}\ve_d$ & fixed row & $\gG_0$; the likelihood inside $\Pi_d$ & the state carries evidence of $d$ (Lemma~\ref{lem:domain}) \\
mask $\vm_d = \mathbf{M}_{\mathrm{voc}}^{\top}\ve_d$ & fixed row & support of both guidance branches & closure (Prop.~\ref{prop:bridge}(iii)); non-expansive; survives guidance (Lemma~\ref{lem:mask}) \\
embedding $\vz_d$, prototype $\vs_{\mathrm{proto},d}$ & learned row & $\vpi$, through $\vc$ & supplies $d$ without inferring it from the state (Lemma~\ref{lem:domain}(c)) \\
guidance weight $g(d)$ & lookup & tilt of $\vpi$ & tempers $P(d \mid x_1^i, \gG_t)$ (Corollary~\ref{cor:guided}) \\
bridge kernel $B_{t\to s}$ & none & every transition & shared by all domains (Prop.~\ref{prop:bridge}(i)) \\
\bottomrule
\end{tabular}
\end{table}

\paragraph{The reveal picture.} \Eqref{eq:forward} says that at time $t$ each entry shows its clean value
with probability $t$ and a prior draw otherwise. A convenient way to produce exactly this is to give every
entry its own hidden alarm clock $\tau^i$, uniform on $[0,1]$: before the clock rings the entry displays
a fixed noise class $x_0^i$ drawn from the prior, and from the moment it rings it displays its clean class
$x_1^i$. Time $t$ then simply reveals the entries whose clocks have rung. Lemma~\ref{lem:coupling} checks
that this construction reproduces \Eqref{eq:forward}; its value is that questions about the reverse step
become questions about when independent clocks ring.

\begin{lemma}[Reveal coupling]
\label{lem:coupling}
Draw $\gG_0 \sim \otimes_i \vq_d^i$, $\gG_1 \sim P_{\mathrm{data}}(\cdot \mid d)$ and reveal times
$\tau^i \sim \mathrm{Unif}(0,1)$, all independent, and for $u \in [0,1]$ set $x_u^i = x_1^i$ if
$\tau^i \le u$ and $x_u^i = x_0^i$ otherwise. Then (a) for every $u$, the law of $\gG_u$ given
$(\gG_1, d)$ is Definition~\ref{def:forward}; (b) given $x_1^i$, entry $i$ is a one-jump process from
$x_0^i$ to $x_1^i$ at time $\tau^i$; and (c) $\gG_0$ is independent of $\gG_1$.
\end{lemma}
\begin{proof}
For each entry, $P(x_u^i = x \mid x_1^i, d) = P(\tau^i \le u)\,\delta_{x_1^i}(x) + P(\tau^i > u)\,\vq_d^i(x)
= u\,\delta_{x_1^i}(x) + (1-u)\,\vq_d^i(x)$. The pairs $(x_0^i, \tau^i)$ are independent across entries
given $(\gG_1, d)$, so the entries are conditionally independent and their joint law is
\Eqref{eq:forward} at time $u$. Parts (b) and (c) hold by construction.
\end{proof}
This coupling is the mixture path of \citet{gat2024discrete} with schedule $\kappa_u = u$ and source
$\vq_d$. Its source sample $\gG_0$ is the state Algorithm~\ref{alg:sample} is initialized with.

\begin{lemma}[Conditional bridge: exact, bounded and domain-free]
\label{lem:bridge}
Let $0 \le t < s \le 1$ and $a = a(t,s)$.
\begin{itemize}
\item[(a)] $P(\gG_s \mid \gG_t, \gG_1, d) = \prod_i B_{t\to s}(x_s^i \mid x_t^i, x_1^i)$.
\item[(b)] $a \in (0,1]$, so every $B_{t\to s}(\cdot \mid x_t, x_1)$ is a probability vector; $a \to 0$
as $s \downarrow t$, and $a = 1$ if and only if $s = 1$.
\item[(c)] $B_{t\to s}$ is the exact solution on $[t,s]$ of the coupling's conditional forward equation,
whose generator is unbounded as $u \to 1$.
\item[(d)] $B_{s\to r} \circ B_{t\to s} = B_{t\to r}$ for $t < s < r \le 1$.
\item[(e)] The first-order (Euler) step of that equation from $t$, with step $s - t$, equals $B_{t\to s}$.
\item[(f)] $B_{t\to s}$ depends only on $(t, s, x_t, x_1)$; it does not depend on $d$, $\vq_d$ or $\sV_d$.
\end{itemize}
\end{lemma}
\begin{proof}
(a) Fix an entry and drop the index $i$. If $x_t \neq x_1$, then $\tau > t$ and $x_0 = x_t$, so
$x_s = x_1$ exactly when $\tau \le s$, which has probability $P(\tau \le s \mid \tau > t) =
\frac{s-t}{1-t} = a$; otherwise $x_s = x_0 = x_t$. If $x_t = x_1$, then either $\tau \le t$, or $\tau > t$
and $x_0 = x_1$; in both cases $x_u = x_1 = x_t$ for every $u \ge t$, so $x_s = x_t$, which is
$a\,\delta_{x_1} + (1-a)\,\delta_{x_t}$ because $\delta_{x_1} = \delta_{x_t}$. Given $\gG_1$, the pairs
$(x_0^i, \tau^i)$ are mutually independent, so the event that decides $x_s^i$ is independent of the
other entries' states. Hence $P(x_s^i \mid \gG_t, \gG_1, d) = B_{t\to s}(x_s^i \mid x_t^i, x_1^i)$, and the
entries evolve independently, which gives the product.

(b) $0 < s - t \le 1 - t$ gives $a \in (0,1]$, so $B_{t\to s}$ is a convex combination of point masses.

(c) Given $x_1$, an unrevealed entry jumps to $x_1$ at the hazard rate of $\tau$,
$f_\tau(u)/P(\tau > u) = 1/(1-u)$. The conditional generator is therefore
$R_u(y \to x \mid x_1) = \frac{\delta_{x_1}(x) - \delta_y(x)}{1-u}$, which vanishes for $y = x_1$ and is
unbounded as $u \to 1$. The forward equation for $p_u = P(x_u = \cdot \mid x_t, x_1)$ on $[t,s]$ is
\begin{equation}
\partial_u p_u \;=\; \sum\nolimits_y p_u(y)\,R_u(y \to \cdot \mid x_1) \;=\; \frac{\delta_{x_1} - p_u}{1-u},
\qquad p_t = \delta_{x_t}.
\end{equation}
With $r_u = p_u - \delta_{x_1}$ it reads $\partial_u r_u = -r_u/(1-u)$, so $r_s = r_t\,e^{-\Lambda(t,s)}$ with
$\Lambda(t,s) = \int_t^s \frac{du}{1-u} = \log\frac{1-t}{1-s}$, and
\begin{equation}
p_s \;=\; \big(1 - e^{-\Lambda(t,s)}\big)\,\delta_{x_1} + e^{-\Lambda(t,s)}\,\delta_{x_t},
\qquad e^{-\Lambda(t,s)} \;=\; \frac{1-s}{1-t} \;=\; 1-a.
\end{equation}
The singularity of the generator enters the finite-step solution only through the survival factor
$e^{-\Lambda(t,s)} \in [0,1)$. Every step is therefore bounded, and $e^{-\Lambda(t,s)} \to 0$ as $s \to 1$
forces full resolution at the endpoint.

(d) Because $\delta_{x_1}$ is absorbing, composing two kernels keeps the form
$a\,\delta_{x_1} + (1-a)\,\delta_{x_t}$, now with $1 - a = (1-a_{ts})(1-a_{sr}) =
\frac{1-s}{1-t}\cdot\frac{1-r}{1-s} = 1 - a_{tr}$. Given $x_1$, one step therefore equals any subdivision
of the same interval: there is no discretization error.

(e) $\delta_{x_t} + (s-t)\,R_t(x_t \to \cdot \mid x_1) = \big(1 - \frac{s-t}{1-t}\big)\delta_{x_t} +
\frac{s-t}{1-t}\,\delta_{x_1} = B_{t\to s}$, because $e^{-\Lambda(t,s)}$ is affine in $s$. The coincidence is
specific to the linear schedule $\kappa_u = u$ (a general schedule gives $1-a = \frac{1-\kappa_s}{1-\kappa_t}$).
The grid $\psi$ of Algorithm~\ref{alg:sample} only chooses which times are visited, so it is unaffected.

(f) is read off the formula.
\end{proof}

\paragraph{A worked example.} Take one edge entry whose clean class is a single bond, observed at $t = 0.5$
and stepped to $s = 0.75$, so $a = 0.25/0.5 = 0.5$. In the reveal picture, $a$ is the chance that
a clock which has not rung by $t$ rings in $(t, s]$: half of the remaining time is used up, so half of
the still-unrevealed entries are revealed. If the entry currently shows ``no edge'', it becomes a single
bond with probability $0.5$ and otherwise stays; if it already shows a single bond, it stays, because
either way it is correct. Splitting the same interval into $0.5 \to 0.625 \to 0.75$ gives $a = 0.25$ and
then $a = \frac{1}{3}$; the probability of staying unrevealed across both is $0.75 \cdot \frac{2}{3} = 0.5$,
the same as the single step. This is part (d): the step size does not matter.

\paragraph{Why this matters.} Lemma~\ref{lem:bridge} gives the sampler three properties. It is
\emph{exact}: given the clean graph, one step of any size is the true transition, so the only error a
sampler can make is in estimating the clean graph. It is \emph{bounded}: although the underlying jump rate
$1/(1-u)$ blows up near $u = 1$, the step only ever uses a probability $a \in (0,1]$, so nothing has to be
clipped. And it is \emph{domain-free}: all domains share one parameter-free step, and the domain can only
act through the prediction of the clean graph and the vocabulary it is restricted to. The final remark of
this appendix contrasts this with rate-based samplers, where the first two properties fail.

\paragraph{From one clean graph to the true posterior.} Lemma~\ref{lem:bridge} holds for a known clean
graph, but a sampler only has a prediction of it. Lemma~\ref{lem:transport} is the bridge between the two:
if we average over where an entry might be at time $t$, the step lands exactly on its time-$s$ marginal,
and the prior merely rides along.

\begin{lemma}[Conditional transport]
\label{lem:transport}
For every $x_1$, $\sum_{x_t} P_t(x_t \mid x_1, d)\,B_{t\to s}(\cdot \mid x_t, x_1) = P_s(\cdot \mid x_1, d)$.
\end{lemma}
\begin{proof}
The left side is $\sum_{x_t}\big[t\,\delta_{x_1}(x_t) + (1-t)\,\vq_d(x_t)\big]\big[a\,\delta_{x_1} +
(1-a)\,\delta_{x_t}\big] = \big[a + (1-a)t\big]\,\delta_{x_1} + (1-a)(1-t)\,\vq_d$, which equals
$s\,\delta_{x_1} + (1-s)\,\vq_d$ because $(1-a)(1-t) = 1-s$ and $a + (1-a)t = 1 - (1-a)(1-t) = s$. The
prior appears in the input and output marginals, never in the kernel.
\end{proof}

\begin{proof}[Proof of Proposition~\ref{prop:bridge}]
\label{app:bridge-proof}
(i) is Lemma~\ref{lem:bridge}(a) together with (f).

(ii) The strategy is to condition on the clean graph, where entries evolve independently and
Lemma~\ref{lem:transport} applies to each one, and then to average over the clean graph with Bayes' rule.
By the law of total probability over the clean graph and (i),
\begin{equation}
P(\gG_s \mid \gG_t, d) \;=\; \sum\nolimits_{\gG_1} \Pi_d(\gG_1 \mid \gG_t) \prod\nolimits_i
B_{t\to s}(x_s^i \mid x_t^i, x_1^i).
\label{eq:exact-kernel}
\end{equation}
Summing \Eqref{eq:exact-kernel} over every entry but $i$, each other factor sums to one, which leaves
$P(x_s^i \mid \gG_t, d) = \sum_x \pi_d^i(x \mid \gG_t)\,B_{t\to s}(x_s^i \mid x_t^i, x) =
a\,\pi_d^i(x_s^i \mid \gG_t) + (1-a)\,\delta_{x_t^i}(x_s^i)$: this is \Eqref{eq:reverse}. For transport,
Bayes' rule $P_t(\gG_t \mid d)\,\Pi_d(\gG_1 \mid \gG_t) = P_{\mathrm{data}}(\gG_1 \mid d)\,P_t(\gG_t \mid
\gG_1, d)$ and the entrywise factorization of $P_t(\gG_t \mid \gG_1, d)$ give
\begin{align}
\sum\nolimits_{\gG_t} P_t(\gG_t \mid d)\,P(\gG_s \mid \gG_t, d)
&= \sum\nolimits_{\gG_1} P_{\mathrm{data}}(\gG_1 \mid d) \prod\nolimits_i \sum\nolimits_{x_t^i}
P_t(x_t^i \mid x_1^i, d)\,B_{t\to s}(x_s^i \mid x_t^i, x_1^i) \nonumber \\
&= \sum\nolimits_{\gG_1} P_{\mathrm{data}}(\gG_1 \mid d) \prod\nolimits_i P_s(x_s^i \mid x_1^i, d)
\;=\; P_s(\gG_s \mid d),
\end{align}
with Lemma~\ref{lem:transport} in the second step. The step size never entered, so induction over any grid
$0 = t_0 < \dots < t_K = 1$ carries $P_0(\cdot \mid d) = \otimes_i \vq_d^i$ to
$P_1(\cdot \mid d) = P_{\mathrm{data}}(\cdot \mid d)$.

(iii) A step can only keep an entry's current class or replace it with a draw from $\vpi^i$, so if both
lie in $\sV_d^i$ the entry never leaves it. Formally, by (A1), $\gG_0 \in \prod_i \sV_d^i$. If $x_t^i \in \sV_d^i$, the law $a\,\vpi^i + (1-a)\,\delta_{x_t^i}$
is supported on $\mathrm{supp}\,\vpi^i \cup \{x_t^i\} \subseteq \sV_d^i$. Induction over the steps
completes the proof.
\end{proof}

\begin{remark}[The graph-level reading fails]
Reading \Eqref{eq:reverse} at the level of whole graphs, as $a\,\Pi_d(\cdot \mid \gG_t) + (1-a)\,
\delta_{\gG_t}$, resolves every entry at once with probability $a$. That kernel maps $P_t(\cdot \mid d)$ to
$a\,P_{\mathrm{data}}(\cdot \mid d) + (1-a)\,P_t(\cdot \mid d)$, which is not $P_s(\cdot \mid d)$ in
general. For instance, take two entries and data that never use a class $g$ with $\vq_d(g) > 0$: the state
$(g,g)$ has probability $(1-s)^2\,\vq_d(g)^2$ under $P_s$ but $(1-s)(1-t)\,\vq_d(g)^2$ under the
graph-level kernel. Entries must be resolved independently, as in \Eqref{eq:exact-kernel}.
\end{remark}
The graph-level reading is tempting because \Eqref{eq:reverse} has the same ``$a$ times prediction plus
$(1-a)$ times current state'' shape. The difference is how many coins are flipped. In the forward process
every entry has its own clock, so at any time some entries are revealed and others are not. The
graph-level kernel flips one coin for the whole graph, which reveals all entries together or none, and
therefore produces correlations between entries that the forward process never has. In the example, the
two entries of $(g,g)$ should each stay unrevealed through the step independently, which costs a factor
$(1-a)$ per entry; the single shared coin charges that factor only once, so $(g,g)$ is over-weighted by
$1/(1-a)$.

\paragraph{Does the noisy graph already reveal its domain?} Proposition~\ref{prop:bridge} leaves all
domain dependence to the posterior $\vpi$. Conditioning the network on $d$ is only useful to the extent that
the noisy graph does not already reveal $d$, so we next locate where $d$ enters the exact posterior and
measure how much the state alone gives away.

\begin{lemma}[Where the domain enters $\vpi$]
\label{lem:domain}
\begin{itemize}
\item[(a)] $\Pi_d(\gG_1 \mid \gG_t) \propto P_{\mathrm{data}}(\gG_1 \mid d) \prod_i \big[t\,\delta_{x_1^i}(x_t^i)
+ (1-t)\,\vq_d^i(x_t^i)\big]$, so $d$ enters only through the domain's data law and its prior
$\vq_d = \mathbf{Q}_{\mathrm{pr}}^{\top}\ve_d$.
\item[(b)] Under (A2), $\mathrm{supp}\,\pi_d^i(\cdot \mid \gG_t) \subseteq \sV_d^i$.
\item[(c)] A model that ignores $d$ can at best estimate the mixture $\Pi_\varnothing =
\sum_{d'} P(d' \mid \gG_t)\,\Pi_{d'}$, where $P(d' \mid \gG_t) \propto P(d')\,P(n \mid d')\,
P_t(\gG_t \mid d', n)$, and $\TV(\Pi_\varnothing, \Pi_d) \le 1 - P(d \mid \gG_t)$.
\item[(d)] If some $x_t^i \notin \sV_{d'}^i$, then $P(d' \mid \gG_t) = 0$.
\item[(e)] For a single entry with clean marginal $p_d$, $\pi_d(\cdot \mid x_t) = \lambda_d\,\delta_{x_t} +
(1-\lambda_d)\,p_d$ with $\lambda_d = \frac{t\,p_d(x_t)}{t\,p_d(x_t) + (1-t)\,\vq_d(x_t)}$.
\end{itemize}
\end{lemma}
\begin{proof}
(a) is Bayes' rule with the likelihood \Eqref{eq:forward}, and (e) is (a) for a single entry. (b)
$\Pi_d(\gG_1 \mid \gG_t) > 0$ requires $P_{\mathrm{data}}(\gG_1 \mid d) > 0$. (c) The mixture form is Bayes'
rule over $d'$, with the node count part of the observed state. Then $\Pi_\varnothing - \Pi_d =
\sum_{d' \neq d} P(d' \mid \gG_t)\,(\Pi_{d'} - \Pi_d)$, and each difference has total variation at most one.
(d) Under (A1) and (A2), $P_t(x_t^i \mid x_1^i, d') > 0$ only if $x_t^i \in \{x_1^i\} \cup
\mathrm{supp}\,\vq_{d'}^i \subseteq \sV_{d'}^i$.
\end{proof}
In words, (c) says that a domain-blind model is wrong only in proportion to its uncertainty about which
domain it is looking at: if the state leaves no doubt about $d$, the blind and conditional posteriors
coincide. (d) says that a single class outside a domain's vocabulary removes that domain from consideration
entirely. Table~\ref{tab:domain-separation} measures how much doubt actually remains in our five domains.

\begin{table}[h]
\centering
\small
\caption{How the noisy state itself separates the five domains at $t = 0$ (Lemma~\ref{lem:domain}(c,d)).
Evidence is the expected log-likelihood ratio of the priors; sizes and priors are measured on the
training split.}
\label{tab:domain-separation}
\begin{tabular}{p{0.2\linewidth} p{0.2\linewidth} p{0.5\linewidth}}
\toprule
Domain pair & Separated by & Evidence \\
\midrule
structural vs.\ molecular & vocabulary (exact) & disjoint node supports (one generic class vs.\ heavy atoms), and disjoint sizes ($n \ge 44$ vs.\ $n \le 38$) \\
planar vs.\ tree & prior & both always have $n = 64$; edge densities $0.088$ vs.\ $0.031$ give $53$--$73$ nats \\
SBM vs.\ planar, tree & node count, then prior & $P(n = 64 \mid \text{SBM}) = 0.031$; at $n = 64$ the prior gives $43$--$58$ nats against tree but only $0.6$ nats against planar (densities $0.081$ vs.\ $0.088$) \\
QM9 vs.\ ZINC & size, prior, partial vocabulary & sizes overlap only for $6 \le n \le 9$; $0.05$ nats per node and $0.17$ per pair; a ZINC state containing P, S, Cl, Br or I rules out QM9 exactly \\
\bottomrule
\end{tabular}
\end{table}

\paragraph{Consequences.} Because the priors are domain-specific and $n$ is drawn from the domain's size
distribution, the noisy state already identifies the domain almost surely
(Table~\ref{tab:domain-separation}). Masks and priors separate structural from molecular domains exactly,
by Lemma~\ref{lem:domain}(d) and Proposition~\ref{prop:bridge}(iii), and the only weakly separated case is
an SBM graph with $n = 64$ against planar. Under a shared prior ($\vq_d = \vq$) the prior channel would
carry no evidence. By (c), a domain-blind posterior is therefore close to the domain-conditional one; what
the conditioning $\vc$ adds is that the network does not have to infer $d$ from these channels, and $\vc$ is
also the only channel that interpolation (\Secref{sec:interp}) moves. By (e), when $\vq_d$ equals the
domain's class marginal, as with DiG$^{2}$'s marginal prior up to size effects, $\lambda_d \equiv t$ or the weight an entry places on its own observed class is then domain-free, so that entry's posterior differs across domains only through $\vq_d$ itself. Any evidence beyond the per-entry priors therefore lives in the joint configuration of $\gG_t$, which is what the size and density gaps of Table~\ref{tab:domain-separation} measure.

\paragraph{Masking and guidance.} Masking sets the logits of forbidden classes to a large negative number,
which is the same as renormalizing the prediction onto the allowed classes. Lemma~\ref{lem:mask} shows
that this is exact in floating point, that it can never move a prediction further from a target that lives
on the allowed classes, and that it survives guidance only if both branches are masked.

\begin{lemma}[Vocabulary masking]
\label{lem:mask}
Let $\vell \in \R^{C}$, let $\vell\!\upharpoonright_{\sV}$ set the logits of classes outside $\sV$ to $-M$,
and let $\vpi = \softmax(\vell)$.
\begin{itemize}
\item[(a)] As $M \to \infty$, $\softmax(\vell\!\upharpoonright_{\sV}) \to \vpi_{\sV}$, the renormalized
restriction $\vpi_{\sV}(x) = \vpi(x)\,\1[x \in \sV] / \vpi(\sV)$; with $M = 10^{9}$ the off-$\sV$
probabilities are exactly zero in single precision.
\item[(b)] For every $\vpi^\star$ supported on $\sV$, $\TV(\vpi_{\sV}, \vpi^\star) \le \TV(\vpi, \vpi^\star)$.
\item[(c)] If both guidance branches are masked, the guided logit of every class outside $\sV$ is
$-M + g\,(-M - (-M)) = -M$ for every $g \in \R$. If only $\vell^{c}$ is masked it is $(1-g)\,\ell^{u} - gM$,
which fails for $g \le 0$; if only $\vell^{u}$ is masked it is $(g-1)M + g\,\ell^{c}$, which removes the
mask at $g = 1$ and inverts it for $g > 1$.
\end{itemize}
\end{lemma}
\begin{proof}
(a) The masked logits leave the relative weights inside $\sV$ unchanged, and $e^{-M}$ vanishes. (b) For
distributions $\vp, \vr$, $\TV(\vp, \vr) = \sum_x (\vr(x) - \vp(x))^{+}$. Taking $\vr = \vpi^\star$, the terms
outside $\sV$ vanish for both arguments, and $\vpi_{\sV} \ge \vpi$ on $\sV$ makes every remaining term no
larger. (c) is direct substitution into \Eqref{eq:cfg}.
\end{proof}
By (b) and Lemma~\ref{lem:domain}(b), masking can only reduce the estimator's error against the exact
posterior. By (c), premise (A3) is necessary for the guided sampler of \Eqref{eq:cfg} at the guidance
weights we use, $g(d) \ge 2$. To see the failure concretely, take $g = 2$ and mask only the unconditional
branch: a forbidden class receives the guided logit $M + 2\,\ell^{c}$, so the mask, instead of removing the
class, makes it almost certain to be drawn. Guidance subtracts the unconditional logits, and subtracting a
one-sided $-M$ turns it into $+M$.

\begin{corollary}[Guided sampler]
\label{cor:guided}
Let $\vpi_g^i \propto (\vpi^{u,i})^{1-g}(\vpi^{c,i})^{g}$ on $\sV_d^i$, the entrywise form of \Eqref{eq:cfg}
under (A3).
\begin{itemize}
\item[(a)] $\vpi_g^i$ is a probability vector on $\sV_d^i$ for every $g \in \R$, so the kernel
$a\,\vpi_g^i + (1-a)\,\delta_{x_t^i}$ is a valid law for every $(t, s, g)$ and
Proposition~\ref{prop:bridge}(iii) applies.
\item[(b)] If $\vpi^{c,i} = \pi_d^i$ and $\vpi^{u,i} = \pi_\varnothing^i$, the entry marginal of
$\Pi_\varnothing$, then $\vpi_g^i(x) \propto \pi_\varnothing^i(x \mid \gG_t)\,P(d \mid x_1^i = x, \gG_t)^{g}$.
\item[(c)] For $g \neq 1$ the sampler no longer targets $P_{\mathrm{data}}(\cdot \mid d)$, so
Proposition~\ref{prop:bridge}(ii) is claimed for $g = 1$ only; Algorithm~\ref{alg:sample} uses $g = 1$
at its final step, where $a = 1$.
\end{itemize}
\end{corollary}
\begin{proof}
(a) follows from Lemma~\ref{lem:mask}(c): the guided logits are finite on $\sV_d^i$ and equal to $-M$ off it.
(b) By Bayes' rule, $P(d \mid x_1^i = x, \gG_t) = P(d \mid \gG_t)\,\pi_d^i(x \mid \gG_t) /
\pi_\varnothing^i(x \mid \gG_t)$, so $(\pi_\varnothing^i)^{1-g}(\pi_d^i)^{g} = \pi_\varnothing^i\,
(\pi_d^i/\pi_\varnothing^i)^{g} \propto \pi_\varnothing^i\,P(d \mid x_1^i = x, \gG_t)^{g}$. (c) At $g = 1$,
$\vpi_g = \vpi^{c}$.
\end{proof}
Guidance therefore tempers the per-entry domain posterior, pushing each entry toward the classes that are
most indicative of $d$. Read (b) as a reweighting: start from the domain-blind prediction and multiply
each class by how strongly it points to $d$, raised to the power $g$. At $g = 1$ this recovers plain
conditional sampling; at $g > 1$ it sharpens toward classes that are characteristic of $d$, at the price of
no longer sampling $P_{\mathrm{data}}(\cdot \mid d)$ exactly (part (c)). For $g > 0$ this tilt already vanishes outside $\sV_d$, because
$P(d \mid x_1^i = x, \gG_t) = 0$ there; the mask enforces on the estimator a property the exact tilt has.
Applying guidance to transition rates instead, as $R_u^{1-g}R_c^{g}$, is unbounded wherever $R_u \to 0$
faster than $R_c$ and must be clipped; the logit-level form \Eqref{eq:cfg} is bounded by construction.

\paragraph{What the network's error costs.} Everything so far assumed the exact posterior $\Pi_d$. In
practice a network predicts one independent distribution per entry, and the sampler takes $T$ steps with
it. Proposition~\ref{prop:error} bounds the resulting error in the generated distribution. The idea of the
proof is a telescoping sum: each step adds the error of its own prediction, and errors made at earlier
steps cannot grow afterwards, because applying any Markov kernel to two distributions never moves them
further apart in total variation.

\begin{aproposition}[Bounded error of the implemented sampler]
\label{prop:error}
Fix a grid $0 = t_0 < \dots < t_T = 1$ with $a_k = a(t_k, t_{k+1})$. At step $k$, let $\hat\Pi_k(\cdot \mid \gG)$
be any law over clean graphs and let $\hat K_k(\cdot \mid \gG) = \sum_{\gG_1} \hat\Pi_k(\gG_1 \mid \gG)
\prod_i B_{t_k \to t_{k+1}}(\cdot \mid x^i, x_1^i)$; the network's factorized prediction
$\hat\Pi_k = \otimes_i \hat\pi_k^i$ gives the step of Algorithm~\ref{alg:sample},
$\prod_i [a_k\,\hat\pi_k^i + (1-a_k)\,\delta_{x^i}]$. Let $\hat P_1$ be the law of the last iterate started
from $\gG_0 \sim \otimes_i \vq_d^i$, and let $\Pi_{d,S}$ denote the marginal of $\Pi_d$ on a set of
entries $S$. Then
\begin{align}
\TV\big(\hat P_1, P_{\mathrm{data}}(\cdot \mid d)\big) &\le \sum_{k=0}^{T-1}
\E_{\gG \sim P_{t_k}(\cdot \mid d)}\,\rho_k(\gG), \label{eq:error-bound} \\
\rho_k(\gG) &\le \E_{S}\,\TV\big(\hat\Pi_{k,S}, \Pi_{d,S}\big) \le
\big(1 - (1-a_k)^{N_n}\big)\,\TV\big(\hat\Pi_k, \Pi_d\big), \nonumber
\end{align}
where $\rho_k(\gG) = \TV(\hat K_k(\cdot \mid \gG), K_k(\cdot \mid \gG))$, $K_k$ is the exact kernel
\Eqref{eq:exact-kernel}, and $S$ includes each entry independently with probability $a_k$. For the
factorized prediction, $\TV(\hat\Pi_{k,S}, \Pi_{d,S}) \le \sum_{i \in S} \TV(\hat\pi_k^i, \pi_d^i) +
\TV(\otimes_{i \in S}\pi_d^i, \Pi_{d,S})$, and the last term vanishes when $|S| \le 1$.
\end{aproposition}
\begin{proof}
Write $P_k = P_{t_k}(\cdot \mid d)$ and $\hat P_k$ for the law of the $k$-th iterate, so that
$P_{k+1} = P_k K_k$ by Proposition~\ref{prop:bridge}(ii) and $\hat P_{k+1} = \hat P_k \hat K_k$. Then
$\hat P_{k+1} - P_{k+1} = (\hat P_k - P_k)\,\hat K_k + P_k\,(\hat K_k - K_k)$. A Markov kernel does not
increase total variation, so $\TV(\hat P_{k+1}, P_{k+1}) \le \TV(\hat P_k, P_k) + \E_{\gG \sim P_k}
\rho_k(\gG)$; summing from $\hat P_0 = P_0$ gives the first inequality. For the second, the product
$\prod_i [a_k\,\delta_{x_1^i} + (1-a_k)\,\delta_{x^i}]$ is the law of ``draw $S$, copy $x_1^i$ on $S$ and
keep $x^i$ elsewhere''. Both kernels are therefore mixtures over the same law of $S$, of the $S$-marginal
of $\hat\Pi_k$ or $\Pi_d$, and joint convexity of total variation bounds $\rho_k$ by
$\E_S \TV(\hat\Pi_{k,S}, \Pi_{d,S})$. The term $S = \emptyset$ vanishes and marginalization does not increase
total variation, which gives the third inequality. The last claim is the triangle inequality with
subadditivity of total variation over product measures; for $|S| \le 1$ the product of marginals equals
the marginal.
\end{proof}
Four consequences follow.
\begin{itemize}
\item \emph{No discretization error.} With the exact posterior every term vanishes for any $T$.
\item \emph{Errors add rather than compound.} Each step contributes at most the posterior error on the
entries it resolves, weighted by $a_k \le 1$. On the grid $t_k = \psi(k/T)$ with $\psi(u) = 2u - u^2$,
$1 - a_k = \big(\frac{m-1}{m}\big)^2$ with $m = T - k$, so $a_0 = \frac{2T-1}{T^2} \approx 0.01$ at
$T = 200$ and $a_{T-1} = 1$: early steps resolve few entries and so expose little of the network's error.
\item \emph{Masking helps.} It can only shrink the first term of the factorized bound
(Lemma~\ref{lem:mask}(b)).
\item \emph{Guidance is a deliberate deviation.} With guidance, the same bound measures the deviation from
$P_{\mathrm{data}}(\cdot \mid d)$ that the tilt introduces on purpose.
\end{itemize}
We do not state a numerical bound, because the entrywise sum grows with $N_n$; the value of the result is
its structure.

\begin{remark}[Relation to chain- and rate-based samplers]
Chain-based models (D3PM, DiGress) also compute their reverse step in closed form, from a fixed forward
transition matrix and a predicted clean graph; neither learns a transition object. Their forward chain
resamples the noise at every step, however, so their posterior spreads mass over other noise classes,
whereas $B_{t\to s}$ is supported on $\{x_t, x_1\}$. Rate-based samplers such as DeFoG integrate a
continuous-time generator built from the predicted clean graph with first-order steps, which must be
clipped near $t = 1$, where the rates grow as $1/(1-t)$. \Eqref{eq:reverse} is the exact integration of the
mixture-path generator (Lemma~\ref{lem:bridge}(c,e)), which is why it needs neither.
\end{remark}

%% file: sections/limitations.tex
\section{Limitations}
\label{sec:limitations}

The current benchmarking study is foundational, establishing that multi-domain graph generation is possible. Therefore, the vocabulary formulation is intentionally restricted to 11 available node-classes and 5 edge-classes. Future work with graph generative models could involve exciting applications learning with larger domain vocabularies.

%% file: sections/appendix/implementation.tex
\section{Implementation Details}

\label{app:impl}

\paragraph{Conditioning.} Every conditioning signal the denoiser receives is assembled into a
single vector $\vc$ from two sources. A learned, orthogonally-initialized domain embedding
table, followed by a residual MLP, supplies $\vz_d$, with one extra row reserved for the
classifier-free-guidance null token so that dropped-out graphs receive a dedicated learned null
vector $\vz_{\varnothing}$ rather than a zero or uniform one; a small GNN structural encoder $f_{\mathrm{struct}}(\cdot\,;\vomega)$ summarizes
the current noisy state and is fused with an exponential-moving-average estimate of that
domain's clean-graph statistics,
\begin{align}
\vs_{\mathrm{cur}} &= f_{\mathrm{struct}}(\mX_t, \tE_t\,;\vomega), \qquad
\vs = f_{\mathrm{fuse}}\big([\vs_{\mathrm{proto},d} \Vert \vs_{\mathrm{cur}}]\big), \\
\vs_{\mathrm{proto},d} &\leftarrow \lambda\, \vs_{\mathrm{proto},d}
  + (1-\lambda)\, \frac{\vm_d-\vmu}{\lVert\vm_d-\vmu\rVert}, \qquad
\vm_d = \frac{1}{|\mathcal{B}_d|}\sum_{\gG_1\in\mathcal{B}_d} f_{\mathrm{struct}}(\mX_1, \tE_1\,;\vomega),
\end{align}
where $\mathcal{B}_d$ holds the clean training graphs of domain $d$ in the current batch and
$\vmu \leftarrow \beta_\mu\vmu + (1-\beta_\mu)\,\mathrm{mean}_{d:\,\mathcal{B}_d\neq\emptyset}\,\vm_d$ is a
running center shared by all domains ($\lambda=\beta_\mu=0.99$). Only domains present in the batch
are written, and only in training steps; Algorithm~\ref{alg:proto} gives the update. Centering
removes the component common to every domain, so the prototypes differ in direction rather than
by a shared offset, and the $\ell_2$ normalization keeps each written mean on the unit sphere. Of
the two fused inputs, only $\vs_{\mathrm{cur}}$ receives gradient from the denoising loss; the
prototype is a detached running estimate over clean graphs. The two are concatenated,
$\vc = [\vz_d \Vert \vs]$.
The unconditional branch of guidance replaces the two domain-indexed terms and keeps the third,
\begin{equation}
\vc_{\varnothing} = \big[\,\vz_{\varnothing} \,\Vert\, f_{\mathrm{fuse}}([\mathbf{0} \Vert \vs_{\mathrm{cur}}])\,\big],
\label{eq:cnull}
\end{equation}
so $\vs_{\mathrm{cur}}$ still reads the noisy graph in both branches. The contrast
$\vell^{c}-\vell^{u}$ in \Eqref{eq:cfg} is therefore between domain identities \emph{given the
same observed structure}, not between a structure-aware and a structure-blind prediction.
During training each graph is independently switched from $\vc$ to $\vc_{\varnothing}$ with
probability $p_{\mathrm{drop}}$.

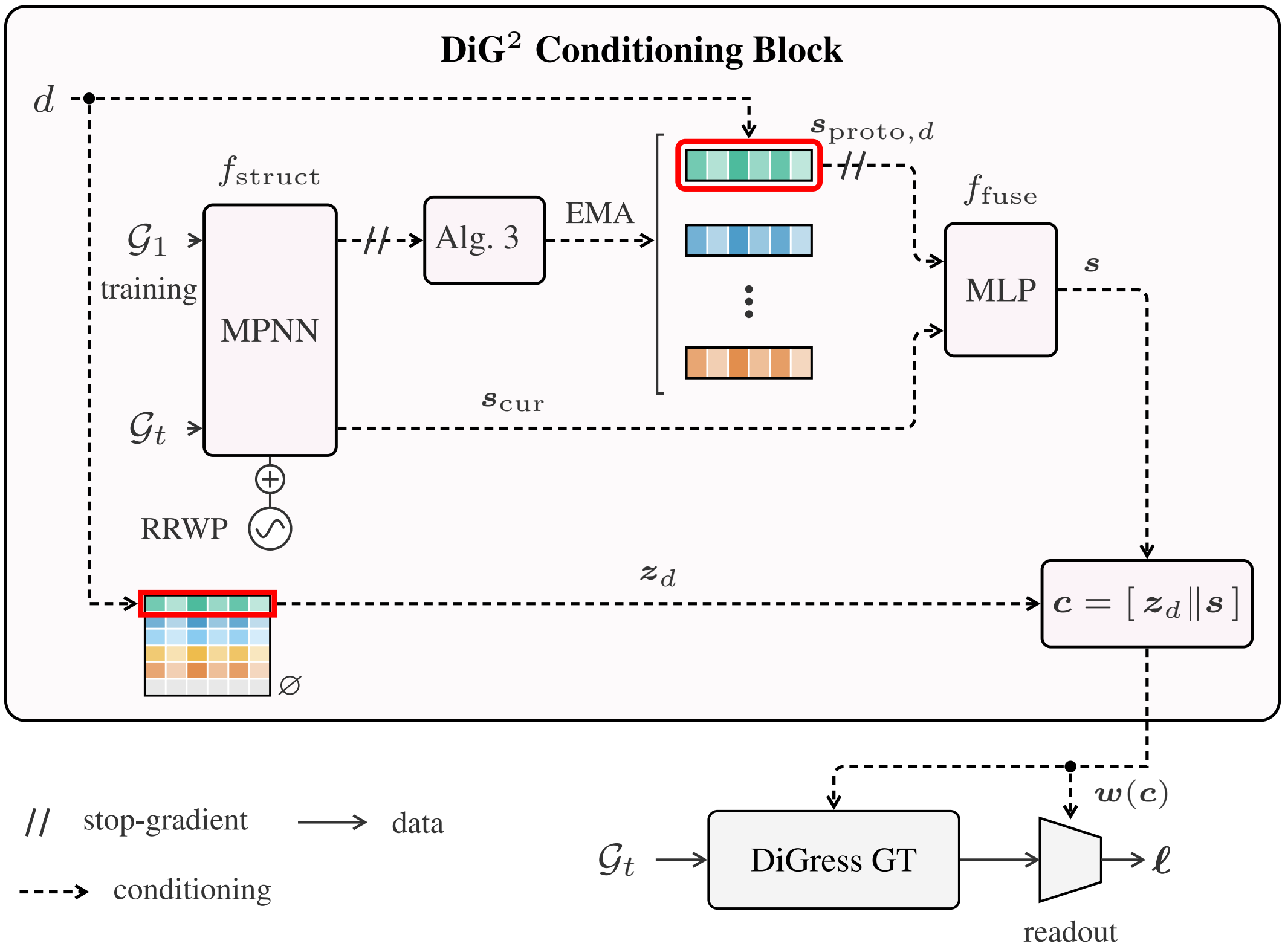
\begin{figure}[!h]
\centering
\input{figures/fig2b_layer}
\caption{\textbf{The DiG$^{2}$ conditioning block.} $\vc=[\vz_d\Vert\vs]$
joins a learned domain embedding $\vz_d$ with a structural summary $\vs$, the highlighted table row
and prototype are those of domain $d$, and the shaded card is the DiG$^{2}$ conditioning block.
Everything outside the card is the inherited transformer and readout.}
\label{fig:cnd_layer}
\end{figure}

\paragraph{Domain-conditioned attention.} Each of transformer layers injects $\vc$
through three near-identity-at-initialization adapters: a residual added to each node's
query/key/value projection, a multiplicative gate on the attention logits, and a FiLM affine
transform applied to the node, edge, and global outputs. The head gate acts on the per-feature
score tensor before the layer rewrites the edges from it, so it modulates the new edge features
as well as the attention weights. All three are scaled by $\tanh$ of learned gains initialised
small (Table~\ref{tab:gdfm-hparams}), so the layer starts close to an unconditioned transformer
and deviates from it only as training progresses. Outside the stack, $\vc$ enters at three more
places. It is concatenated into the global input before the input MLP, so $\vy$ carries it from
the first layer. The timestep embedding is FiLM-modulated by the same conditioning vector $\vc$,
with both a scale and a shift at a fixed gain,
$\vtau \leftarrow \vtau\odot(1+0.25\tanh\vgamma)+0.25\tanh\boldsymbol{\beta}$, where
$(\vgamma,\boldsymbol{\beta})$ come from a two-layer MLP on $\vc$ --- that is, from
$\vz_d$ and the structural summary $\vs$ together, not from the domain embedding alone ---
giving the network a domain-aware notion of "how far along" denoising is; it is applied once
before the stack rather than per layer. Finally, the multi-depth readout weights
$w(\vc)=\mathrm{softmax}(\mathrm{MLP}(\vc))$ are computed from $\vc$ with a zero-initialized
last layer, so they start as a uniform average over $\sK$.

\paragraph{Deep supervision.} Non-terminal readout depths ($\sK \setminus \{L\}$) each receive
their own cross-entropy term against the same target, restricted to $\sV_d$ exactly as the
terminal loss is, but with unit node and edge weights ($\CE_X + \CE_E$) rather than the terminal
loss's $(\lambda_X, \lambda_E) = (1, 5)$ (Table~\ref{tab:gdfm-hparams}). The terms are uniformly
averaged and scaled by $\beta = 1.0$; this is the $\mathcal{L}_{\mathrm{deep\text{-}sup}}$ term in
Algorithm~\ref{alg:train}.

\paragraph{Vocabulary restriction.} $\sV_d$ is the semantically declared class support for
domain $d$. The restriction operator $\upharpoonright_{\sV_d}$ used throughout is a hard mask
applied to logits before any softmax — training targets and both classifier-free-guidance
branches at every sampling step alike — implemented by setting every logit of a class outside $\sV_d$
to $-M$ with $M = 10^{9}$, so those classes receive exactly zero probability. Masking both guidance
branches to the same $\sV_d$ keeps the cond/uncond difference in \Eqref{eq:cfg} purely semantic, and it
is necessary for the mask to survive guidance at $g(d) > 1$ (Lemma~\ref{lem:mask}).

\begin{algorithm}[!h]
\caption{Prototype write (once per training step, after the loss; no gradient)}
\label{alg:proto}
\begin{algorithmic}[1]
\Require clean graphs of the batch, grouped by true domain into $\mathcal{B}_d$;\newline prototypes $\vs_{\mathrm{proto},d}$ (init.\ $\mathbf{0}$); center $\vmu$; momenta $\lambda = \beta_\mu = 0.99$
\Ensure $\vs_{\mathrm{proto},d}$ updated for the domains in the batch only
\State $\vm_d \leftarrow \mathrm{mean}_{\gG_1 \in \mathcal{B}_d}\, f_{\mathrm{struct}}(\gG_1\,;\vomega)$ for each domain $d$ in the batch \Comment{domain mean}
\State $\bar\vm \leftarrow$ mean of these $\vm_d$ \Comment{over the domains in the batch}
\State $\vmu \leftarrow \bar\vm$ on the first write, else $\vmu \leftarrow \beta_\mu\,\vmu + (1-\beta_\mu)\,\bar\vm$ \Comment{shared center}
\For{each domain $d$ in the batch}
    \State $\vs_{\mathrm{proto},d} \leftarrow \lambda\,\vs_{\mathrm{proto},d} + (1-\lambda)\,(\vm_d - \vmu)/\lVert \vm_d - \vmu \rVert$ \Comment{EMA toward centered unit mean}
\EndFor
\end{algorithmic}
\end{algorithm}

\paragraph{Guidance schedule.} $g(d)$ in \Eqref{eq:cfg} is a per-domain lookup table with a
shared fallback (Table~\ref{tab:gdfm-hparams}); classifier-free-guidance dropout uses
$p_{\mathrm{drop}} = 0.05$.

\paragraph{Time discretization.} Both training-time and sampling-time schedules reparameterize
a uniform variable through $\psi(u) = 2u - u^2$, which places more steps near $t=1$. At training
time, $t = \psi(u)$ for $u \sim \mathrm{Unif}(0,1)$; at sampling time, Algorithm~\ref{alg:sample}'s
$T=200$ steps visit the grid $t_k = \psi(k/T)$. Proposition~\ref{prop:bridge}'s resolve
probability for the step from $t_k$ to $t_{k+1}$ is then $a_k = 1 - \big(1 - \frac{1}{T-k}\big)^2$, rising from
$a_0 = \frac{2T-1}{T^2} \approx 0.01$ to $a_{T-1} = 1$.

\begin{table}[h!]
\centering
\small
\begin{tabular}{ll}
\toprule
Quantity & Value \\
\midrule
Transformer depth $L$ & 12 \\
Readout depths $\sK$ & $\{8,10,12\}$ \\
Parameters & 18M \\
Hidden dims (node, edge, global, heads) & 256, 64, 128, 8 \\
Feed-forward dims (node, edge, global) & 256, 128, 2048 \\
Domain-emb.\ dim / structural-encoder depth & 32 / 8 \\
Structural aux.\ features & RRWP, $k = 20$~\cite{ma2023graph} \\
Prototype transform & centered, $\ell_2$-normalized; momentum 0.99 \\
Corruption prior & domain-specific \\
$\mathrm{CE}$ weights $(\lambda_X, \lambda_E, \lambda_y)$ & $(1, 5, 0)$ \\
Deep-supervision weight $\beta$ (terminal excluded) & 1.0 \\
$g(d)$ & \textsc{sbm} 3.0, others 2.0 \\
Guidance dropout $p_{\mathrm{drop}}$ & 0.05 \\
\midrule
\multicolumn{2}{l}{\emph{Conditioning sites for $\vc=[\vz_d\Vert\vs]\in\mathbb{R}^{\mCondDim}$ (\Figref{fig:cnd_layer}, \S\ref{app:impl})}} \\
\quad Residual on $\mQ,\mK,\mV$, per layer & $\mQ \mathrel{+}= \tanh(\gamma_q)\,\mathrm{cond}_q(\vc)$; $\gamma$ init.\ \mGainQKV \\
\quad Attention-head gate, per layer & $\mathbf{Y} \mathrel{*}= 1+\tanh(\gamma_h)\,\mathrm{gate}(\vc)$; $\gamma_h$ init.\ \mGainGate \\
\quad FiLM on $\mX,\tE,\vy$, per layer & $u \mapsto u(1{+}\tanh(\gamma_f)a)+\tanh(\gamma_f)b$; $\gamma_f$ init.\ \mGainFiLM \\
\quad Timestep FiLM, once before the stack & scale and shift, fixed gain \mGainTime \\
\quad Global input, once before the stack & concatenated before the input MLP \\
\quad Readout gate $w(\vc)$ over $\sK$ & zero-init.\ last layer (uniform at init.) \\
\quad Unconditional branch $\vc_{\varnothing}$ & learned null row; prototype zeroed; $\vs_{\mathrm{cur}}$ kept \\
\midrule
Sample steps $T$ / time distortion & 200 / \texttt{polydec} \\
Node budget / steps per epoch / epochs & 256 nodes / 250 / 10{,}000 (\S\ref{app:dataload}) \\
Molecular hydrogen convention & \texttt{remove\_h = true} (heavy-atom only) \\
$\sD$ & planar, sbm, tree, qm9, zinc \\
\bottomrule
\end{tabular}
\caption{DiG$^{2}$ Hyperparameters.}
\label{tab:gdfm-hparams}
\end{table}

\paragraph{A note on $g(d)$.} The per-domain guidance table used by the reported runs is only
sparsely calibrated, \textsc{SBM} at 3.0 and every other domain at the shared fallback of 2.0.

\paragraph{Anchored fine-tuning.} After base training, all benchmarked models are anchor fine-tuned
to improve domain metrics without affecting other domains
($\sD_{\mathrm{prot}}$) (Alg.~\ref{alg:anchor}). In our study, only the edge pathway train: attention
projections, edge MLPs and edge readout heads, $\vtheta_E$. For parameter-matched diffusion models with the graph transformer backbone, this tunes $3.98$M of the $18$M
parameters ($22\%$). Every other weight is frozen, including the readout gate $\vphi$, the
structural encoder $\vomega$, the embedding table and the domain prototypes, which are not written
during fine-tuning. Because $\sD_{\mathrm{prot}}$ shares $\vtheta_E$, its outputs would drift, so a frozen
copy of the wave's starting network, $F_{\vtheta^{(w-1)}}$, acts as an anchor. On graphs from
$\sD_{\mathrm{prot}}$ the student is penalized by $\mathrm{KL(F_{\vtheta^{(w-1)}}\,\Vert\,\Fro)}$,
computed on the same noisy input and conditioning and with both predictions restricted to $\sV_d$,
averaged over nodes ($\mathrm{KL}_X$) and over off-diagonal node pairs ($\mathrm{KL}_E$) with
$\lambda_E=5$, matching the edge weight of the training loss. The task loss (cross-entropy and deep supervision) is applied to $\sD_{\mathrm{tgt}}$ only.
Fine-tuning proceeds in $W=3$ waves. Wave $w$ has a single target domain $d_w$, taken in the
order $(d_1,d_2,d_3)=(\textsc{Tree},\textsc{SBM},\textsc{Planar})$, so that
$\sD_{\mathrm{tgt}}=\{d_w\}$ and $\sD_{\mathrm{prot}}=\sD\setminus\{d_w\}$ holds the other four
domains. For example, the \textsc{Tree} wave has $\sD_{\mathrm{tgt}}=\{\textsc{Tree}\}$ and
$\sD_{\mathrm{prot}}=\{\textsc{SBM},\textsc{Planar},\textsc{QM9},\textsc{ZINC}\}$. Structural
domains draw from their $8{,}192$-graph pools whether they are the target or protected. The anchor
is refreshed at every wave. So, wave $w$ starts from $\vtheta^{(w-1)}$, the checkpoint selected in wave
$w-1$ (with $\vtheta^{(0)}=\vtheta_0$ the base model), and freezes a copy $F_{\vtheta^{(w-1)}}$ of
it as the teacher. A domain that was a target in an earlier wave is therefore anchored to its fine-tuned behavior in subsequent waves as opposed to the base model.
Each loss term is averaged over the entries of its own graphs in the batch, so the balance between
the two terms does not depend on the batch's mix of domains; a batch with no target graph takes an
anchor-only step. $P_{\mathrm{tune}}$ draws the domain of each graph with probability $0.25$
for the target domain and $0.1875$ for each of the four protected domains
($0.25+4\times0.1875=1$). We use $\lambda_{\mathrm{anc}}=12$, AdamW (AMSGrad) at learning rate
$2{\times}10^{-5}$ ($10\times$ below base training) with a $200$-step linear warmup, no weight decay
and gradient clipping at $1.0$. Every wave runs for the full $J=25{,}000$ steps
($75{,}000$ in total); there is no early stopping. Within a wave we keep the checkpoint with the
best average validation metric across the five domains, and that checkpoint both starts the next wave and, after the last wave, is
the model we report.

\begin{algorithm}[H]
\caption[Anchored fine-tuning]{Anchored fine-tuning\protect\newline (shaded lines differ from Alg.~\ref{alg:train}; no prototype writes)}
\label{alg:anchor}
\begin{algorithmic}[1]
\Require $\vtheta_0$ (Alg.~\ref{alg:train}); edge pathway $\vtheta_E \subset \vtheta$; wave targets $(d_1,\dots,d_W)$;\newline $\lambda_{\mathrm{anc}}, \lambda_E$; $J$
\algdiffnext\State $\vtheta^{(0)} \leftarrow \vtheta_0$
\algdiffnext\For{wave $w = 1, \dots, W$} \Comment{\textsc{Tree}, \textsc{SBM}, \textsc{Planar}}
    \algdiffnext\State $\sD_{\mathrm{tgt}} \leftarrow \{d_w\}$;\; $\sD_{\mathrm{prot}} \leftarrow \sD \setminus \{d_w\}$
    \algdiffnext\State $\vtheta \leftarrow \vtheta^{(w-1)}$; freeze a copy $F_{\vtheta^{(w-1)}}$; train only $\vtheta_E$
    \algdiffnext\For{$j = 1, \dots, J$}
        \algdiffnext\State draw a batch $B \sim P_{\mathrm{tune}}$ \Comment{reweighted}
        \For{each $(\gG_1, d) \in B$}
            \State $u \sim \mathrm{Unif}(0,1)$;\; $t \leftarrow \psi(u)$
            \State $\gG_t \sim P_t(\cdot \mid \gG_1, d)$ \Comment{noise to $\vq_d$}
            \State $\vs \leftarrow f_{\mathrm{fuse}}([\vs_{\mathrm{proto},d} \Vert f_{\mathrm{struct}}(\gG_t)])$
            \State $\vc \leftarrow [\vz_d \Vert \vs]$, or $\vc_{\varnothing}$ w.p.\ $p_{\mathrm{drop}}$
            \State $\hat\gG_1 \leftarrow \Fro(\gG_t,t,\vc) \!\upharpoonright_{\sV_d}$
            \algdiffnext\State $\hat\gG_1^{\mathrm{anc}} \leftarrow F_{\vtheta^{(w-1)}}(\gG_t,t,\vc) \!\upharpoonright_{\sV_d}$ \Comment{no gradient}
        \EndFor
        \algdiffnext\State $\mathcal{L}_{\mathrm{task}} \leftarrow \CE(\hat\gG_1, \gG_1) + \mathcal{L}_{\mathrm{deep\text{-}sup}}$ \Comment{over $\sD_{\mathrm{tgt}}$}
        \algdiffnext\State $\mathcal{L}_{\mathrm{anc}} \leftarrow \mathrm{KL}_X + \lambda_E\,\mathrm{KL}_E$ of $(\hat\gG_1^{\mathrm{anc}} \Vert \hat\gG_1)$ \Comment{over $\sD_{\mathrm{prot}}$}
        \algdiffnext\State $\vtheta_E \leftarrow$ AdamW step on $\nabla(\mathcal{L}_{\mathrm{task}} + \lambda_{\mathrm{anc}}\mathcal{L}_{\mathrm{anc}})$
    \EndFor
    \algdiffnext\State $\vtheta^{(w)} \leftarrow$ checkpoint of wave $w$ with the best validation metric
\EndFor
\State \Return $\vtheta^{(W)}$
\end{algorithmic}
\end{algorithm}

\subsection{Data Loading and Node Budgets}
\label{app:dataload}

\paragraph{Mixed-domain training batches.} Batches are filled to a node budget rather than a graph
count, so a single batch can hold graphs from every domain and size range. Each draw first picks a
domain uniformly over the five domain names, then a training graph uniformly within that domain.
The corpus is highly unbalanced ($128$ training graphs each for \textsc{Planar}, \textsc{SBM} and
\textsc{Tree}, $97{,}734$ for \textsc{QM9} and $224{,}568$ for \textsc{ZINC}), so this gives every
domain the same draw probability whatever its size. Graphs are appended while the batch's total
node count stays within $256$. The first draw that would exceed the budget closes the batch and is
discarded, not carried over. An epoch is $250$ batches, with a fresh sampler seed each epoch. The anchored fine-tuning
(Algorithm~\ref{alg:anchor}) draws domains with fixed weights and carries an overflowing
graph into the next batch rather than discarding it.

\paragraph{Independent structure within a batch.} Each graph stays its own connected component.
Collation offsets every graph's edge indices, so no edge crosses graphs. The batch is then made
dense, node features are padded to the batch's largest graph, $N_{\max}$ ($67$ nodes at the median
batch, $174$ at the $99$th percentile), and each graph gets its own $N_{\max}\times N_{\max}$ edge
block, whose padded rows and columns are removed by a per-graph node mask. Padded cells and the
diagonal carry no target class and drop out of the cross-entropy. Per-graph metadata travels with
the batch. The domain index selects each graph's vocabulary mask $\sV_d$. Classifier-free-guidance
dropout is drawn per graph. The same index routes the graph's clean embedding into its domain's
prototype (Algorithm~\ref{alg:proto}). All domains share one vocabulary of $11$ node and $5$ edge
classes. Each domain's local classes are mapped into it once, when the dataset is built. The
structural domains use one generic node class and one edge class, and the molecular domains use
atom types and bond orders. $\sV_d$ records which global classes domain $d$ can use. Padding currently has a
compute cost, on average $62\%$ of a batch's node slots and $52\%$ of its edge slots belong to a
real graph.

\paragraph{Sampling batches.} Generation batches are domain-pure. Each batch generates graphs of
a single domain under that domain's conditioning and holds $\lfloor 2048/\bar n_d\rfloor$ graphs,
where $\bar n_d$ is the domain's mean training size: $19$ graphs per batch for \textsc{SBM}, $32$ for
\textsc{Planar} and \textsc{Tree}, $89$ for \textsc{ZINC} and $227$ for \textsc{QM9}. Each graph's
node count is drawn from its domain's empirical training-size histogram. Graphs are padded to the batch's largest
node count and masked as in training, so each is denoised independently.

\paragraph{Memory and Time Computation.} Within Table~\ref{tab:mem-measure} of this section, we detail the peak memory usage and time required per-epoch of all benchmarked models during training and sampling for the benchmarking study. Within Table~\ref{tab:data-measure}, we detail the dataset statistics for reproducibility. All tests were conducted on a 1 TB RAM machine with 48GB A6000 GPUs. All models were trained for up to 2.5M steps and fine-tuned for 75K steps ($3\times 25$K), and given a 10-day training budget before stopping. Table~\ref{tab:epochs} reports the epochs each arm actually completed inside that budget.

\begin{table}[h!]
    \centering
\begin{tabular}{lrrrrr}
\toprule
 & & \multicolumn{2}{c}{Training} & \multicolumn{2}{c}{Sampling} \\
\cmidrule(lr){3-4}\cmidrule(lr){5-6}
Method & Params (M) & s/epoch & Peak GB & s/graph & Peak GB \\
\midrule
DiG² & 18 & 87.2 & 18.7 & 5.22 & 5.8 \\
G2PT & 89 & 167.4 & 4.6 & 1.52  & 4.5 \\
DeFoG† & 18 & 53.9 & 14.0 & 4.35 & 2.8 \\
DiGress† & 18 & 47.7 & 16.6 & 4.71 & 2.8 \\
HOG-Diff† & 18 & 110.2 & 16.2 & 7.79 & 2.9 \\
GraphBSI† & 18 & 46.6 & 14.0 & 2.43 & 2.8 \\
LGGM & 18 & 46.7 & 17.1 & 3.78 & 3.6 \\
Pard† & 18 & 167.0 & 36.0 & 3.74  & 2.9 \\
GraphRNN† & 18 & 20.0 & 9.4 & 2.1 & 1.9 \\
LGD† & 27 & 127.9 & 19.1 & 4.7 & 2.0 \\
SPECTRE† & 18 & 489.5 & 14.9 & 3.0 & 5.4 \\
GDSS† & 18 & 188.8 & 15.7 & 3.44  & 2.5 \\
\bottomrule
\end{tabular}
\caption{Hyperparameters, per-epoch timing and memory required for training and sampling }
\label{tab:mem-measure}
\end{table}

\begin{table}[h!]
    \centering
\begin{tabular}{lrrr}
\toprule
Dataset & Min Nodes & Max Nodes & Graphs Sampled \\
\midrule
Planar & 64 & 64 & 40 \\
SBM & 44 & 187 & 40 \\
Tree & 64 & 64 & 40  \\
QM9 & 2 & 9 & 10000 \\
ZINC & 8 & 27 & 10000 \\
\bottomrule
\end{tabular}
\caption{Dataset Statistics and Number of Graphs Sampled. }
\label{tab:data-measure}
\end{table}

%% file: figures/fig2b_layer.tex
\resizebox{\linewidth}{!}{%
\begin{tikzpicture}[
  x=1cm, y=1cm,
  lbl/.style   = {font=\rmfamily\fontsize{6.4}{7.4}\selectfont, text=digInk},
  lane/.style  = {font=\fontsize{8}{9}\selectfont, text=digInk},
  cblk/.style  = {draw=black, line width=0.6pt, fill=dig2cond!8, rounded corners=1.8pt,
                  minimum height=0.62cm, inner sep=2pt, align=center,
                  font=\rmfamily\fontsize{7}{8}\selectfont, text=digInk},
  gblk/.style  = {draw=black, line width=0.5pt, fill=digFillSoft, rounded corners=1.8pt,
                  inner sep=2pt, align=center,
                  font=\rmfamily\fontsize{7}{8.4}\selectfont, text=black},
  flow/.style  = {-{Straight Barb[length=2.4pt,width=2.8pt]}, draw=digInk, line width=0.55pt},
  cflow/.style = {-{Straight Barb[length=2.2pt,width=2.6pt]}, draw=black,
                  line width=0.65pt, dash pattern=on 1.8pt off 1.2pt},
  cwire/.style = {draw=black, line width=0.65pt, dash pattern=on 1.8pt off 1.2pt},
  cjn/.style   = {circle, fill=black, inner sep=0pt, minimum size=2.4pt},
]
\def\sg#1#2{%
  \draw[digInk, line width=0.6pt] ({#1-0.05},{#2-0.10}) -- ({#1+0.01},{#2+0.10});
  \draw[digInk, line width=0.6pt] ({#1+0.05},{#2-0.10}) -- ({#1+0.11},{#2+0.10});
}
\def\sgat#1{%
  \draw[digInk, line width=0.6pt] ($(#1)+(-0.08,-0.10)$) -- ($(#1)+(-0.02,0.10)$);
  \draw[digInk, line width=0.6pt] ($(#1)+(0.02,-0.10)$) -- ($(#1)+(0.08,0.10)$);
}
\def\cw{0.15}
\def\strip#1#2#3#4{%
  \foreach \j/\s in {0/55,1/30,2/70,3/40,4/60,5/25}{
    \filldraw[draw=white, line width=0.35pt, fill=#4!\s]
      ({#1+\j*\cw},{#2}) rectangle ({#1+(\j+1)*\cw},{#2+#3});
  }
}
\def\rT{9.42} \def\rW{8.40} \def\rS{7.05} \def\rZ{5.79} \def\rB{3.95}

\filldraw[draw=black, line width=0.6pt, fill=dig2cond!4, rounded corners=4pt]
  (-0.05,4.95) rectangle (9.10,10.08);
\node[font=\rmfamily\bfseries\fontsize{7.6}{9}\selectfont, text=black, anchor=north]
  at (4.52,10.02) {DiG$^{2}$ Conditioning Block};

\node[lane] (g1) at (0.98,\rW) {$\gG_1$};
\node[lbl, anchor=north] at (0.98,{\rW-0.14}) {training};
\node[lane] (gin) at (0.98,\rS) {$\gG_t$};
\node[cblk, minimum width=0.95cm, minimum height=1.80cm] (enc) at (1.85,{(\rW+\rS)/2+0.03}) {MPNN};
\node[lbl, anchor=south, font=\fontsize{7.5}{8.5}\selectfont] at (enc.north) {$f_{\mathrm{struct}}$};
\draw[flow] (g1.east) -- (enc.west |- 0,\rW);
\draw[flow] (gin.east) -- (enc.west |- 0,\rS);
\coordinate (pe) at (1.85,{\rS-0.72});
\draw[digInk, line width=0.5pt, fill=white] (pe) circle (0.15);
\draw[digInk, line width=0.5pt, domain=-0.10:0.10, samples=25, smooth]
  plot ({1.85+\x},{\rS-0.72+0.06*sin(\x*1800)});
\node[lbl, anchor=east] at ($(pe)+(-0.18,0)$) {RRWP};
\node[circle, draw=digInk, fill=white, line width=0.5pt, inner sep=0pt, minimum size=0.20cm]
  (plus) at ($(enc.south)+(0,-0.16)$) {};
\draw[digInk, line width=0.5pt] ($(plus)+(-0.06,0)$) -- ($(plus)+(0.06,0)$)
                                ($(plus)+(0,-0.06)$) -- ($(plus)+(0,0.06)$);
\draw[digInk, line width=0.5pt] ($(pe)+(0,0.15)$) -- (plus.south);
\draw[digInk, line width=0.5pt] (plus.north) -- (enc.south);

\node[cblk, anchor=west] (alg) at (2.95,\rW) {Alg.~\ref{alg:proto}\hspace{3pt}};
\draw[cflow] (enc.east |- 0,\rW) -- (alg.west);
\coordinate (sgw) at ($(enc.east |- 0,\rW)!0.45!(alg.west)$); \sgat{sgw}
\def\cx{4.84} \def\cX{5.74} \def\ph{0.22}
\foreach \c/\y in {digTree/8.94, digPlanar/8.40, digZINC/7.52}{
  \strip{\cx}{\y-\ph/2}{\ph}{\c}
  \draw[black, line width=0.5pt] (\cx,{\y-\ph/2}) rectangle (\cX,{\y+\ph/2});
}
\foreach \dy in {-0.09, 0, 0.09}{ \fill[digInk] ({(\cx+\cX)/2},{7.96+\dy}) circle (0.028); }
\draw[red, line width=1.1pt, rounded corners=1.5pt]
  ({\cx-0.06},{8.94-\ph/2-0.06}) rectangle ({\cX+0.06},{8.94+\ph/2+0.06});
\draw[digInk, line width=0.5pt] ({\cx-0.16},7.30) -- ({\cx-0.21},7.30) -- ({\cx-0.21},9.16)
                                -- ({\cx-0.16},9.16);
\draw[cflow] (alg.east) -- ({\cx-0.23},\rW);
\node[lbl, anchor=south] at ($(alg.east)!0.5!({\cx-0.23},\rW)$) {EMA};

\node[lane] (din) at (0.22,\rT) {$d$};
\node[cjn] (dj) at (0.55,\rT) {};
\draw[cwire] (din.east) -- (dj);
\draw[cflow] (dj) -- ({(\cx+\cX)/2},\rT) -- ({(\cx+\cX)/2},{8.94+\ph/2+0.08});

\def\tx{0.95} \def\rh{0.12}
\foreach \c [count=\i from 0] in {digTree,digPlanar,digSBM,digQM9,digZINC}{
  \strip{\tx}{\rZ+\rh/2-(\i+1)*\rh}{\rh}{\c}
}
\foreach \j in {0,...,5}{
  \filldraw[draw=white, line width=0.35pt, fill=digFill]
    ({\tx+\j*\cw},{\rZ+\rh/2-6*\rh}) rectangle ({\tx+(\j+1)*\cw},{\rZ+\rh/2-5*\rh});
}
\draw[black, line width=0.45pt] (\tx,{\rZ+\rh/2-6*\rh}) rectangle ({\tx+6*\cw},{\rZ+\rh/2});
\node[lbl, anchor=west, inner sep=1pt] at ({\tx+6*\cw},{\rZ+\rh/2-5.5*\rh}) {$\varnothing$};
\draw[red, line width=1.1pt] ({\tx-0.03},{\rZ-\rh/2-0.02}) rectangle ({\tx+6*\cw+0.03},{\rZ+\rh/2+0.02});
\draw[cflow] (dj) -- (0.55,\rZ) -- ({\tx-0.05},\rZ);
\coordinate (zout) at ({\tx+6*\cw+0.05},\rZ);

\node[cblk, minimum width=0.80cm, minimum height=0.95cm] (fuse) at (7.10,{(8.94+\rS)/2+0.05}) {MLP};
\node[lbl, anchor=south, font=\fontsize{7.5}{8.5}\selectfont] at (fuse.north) {$f_{\mathrm{fuse}}$};
\draw[cflow, rounded corners=2pt] ({\cX+0.08},8.94) -- (6.48,8.94) -- (6.48,8.25) -- (fuse.west |- 0,8.25);
\coordinate (sgp) at ({\cX+0.30},8.94); \sgat{sgp}
\node[lbl, anchor=south] at (6.18,8.97) {$\vs_{\mathrm{proto},d}$};
\draw[cflow, rounded corners=2pt] (enc.east |- 0,\rS) -- (6.48,\rS) -- (6.48,7.75) -- (fuse.west |- 0,7.75);
\node[lbl, anchor=south] at (3.60,\rS) {$\vs_{\mathrm{cur}}$};

\node[cblk, font=\fontsize{7}{8}\selectfont] (cvec) at (8.15,\rZ) {$\vc=[\,\vz_d\Vert\vs\,]$};
\draw[cflow] (zout) -- (cvec.west);
\node[lbl, anchor=south] at ($(zout)!0.5!(cvec.west)$) {$\vz_d$};
\draw[cflow, rounded corners=2pt] (fuse.east) -- (8.15,{(8.94+\rS)/2+0.05}) -- (cvec.north);
\node[lbl, anchor=south west] at ($(fuse.east)+(0.05,0.01)$) {$\vs$};

\node[gblk, minimum width=1.80cm, minimum height=0.70cm] (gt) at (5.90,\rB) {DiGress GT};
\node[lane, anchor=east] (gtin) at (4.62,\rB) {$\gG_t$};
\draw[flow] (gtin.east) -- (gt.west);
\def\rx{7.60}
\filldraw[draw=black, line width=0.5pt, fill=digFillSoft]
  ({\rx-0.22},{\rB+0.30}) -- ({\rx+0.22},{\rB+0.16}) -- ({\rx+0.22},{\rB-0.16})
  -- ({\rx-0.22},{\rB-0.30}) -- cycle;
\node[lbl, anchor=north] at (\rx,{\rB-0.31}) {readout};
\draw[flow] (gt.east) -- ({\rx-0.22},\rB);
\draw[flow] ({\rx+0.22},\rB) -- ++(0.32,0) node[lane, anchor=west, inner sep=1pt] {$\vell$};
\node[cjn] (cj) at (\rx,4.62) {};
\draw[cwire] (cvec.south) -- (8.15,4.62) -- (cj);
\draw[cflow, rounded corners=2pt] (cj) -- (gt.north |- 0,4.62) -- (gt.north);
\draw[cflow] (cj) -- (\rx,{\rB+0.29});
\node[lbl, anchor=west] at ({\rx+0.04},4.42) {$\vw(\vc)$};

\def\lgA{4.22} \def\lgB{3.72}
\sg{0.15}{\lgA}
\node[lbl, anchor=west] at (0.38,\lgA) {stop-gradient};
\draw[flow] (2.05,\lgA) -- (2.55,\lgA);
\node[lbl, anchor=west] at (2.60,\lgA) {data};
\draw[cflow] (0.05,\lgB) -- (0.55,\lgB);
\node[lbl, anchor=west] at (0.60,\lgB) {conditioning};
\end{tikzpicture}%
}

%% file: sections/appendix/baselines.tex
\section{Baseline Configurations}
\label{app:baselines}

Every arm in Table~\ref{tab:benchmark} is trained on the same corpus, splits, batching rule and
evaluation harness, so the tables below separate what is \emph{shared} from what is
\emph{architecture} and what is \emph{specific to each generative process}.
The training protocol is shared: AdamW (learning rate $2{\times}10^{-4}$, weight decay
$10^{-12}$), gradient-norm clipping at $1.0$, a budget of $256$ nodes per step, $250$ steps per
epoch for $10{,}000$ epochs, no weight EMA and seed $42$. The exceptions are the upstream
methods' own requirements or memory limits: LGD's denoiser uses learning rate $10^{-4}$ and its
autoencoder trains for $500$ epochs; SPECTRE uses separate generator and discriminator Adam
optimizers (learning rate $10^{-4}$, $(\beta_1,\beta_2)=(0.5,0.9)$), weight EMA $0.995$ and a
$192$-node budget; Pard uses at least $2$ graphs per step. Table~\ref{tab:baseline-arch}
gives each arm's neural backbone and its size relative to DiG$^2$, and
Table~\ref{tab:baseline-gen} lists the parameters that belong to each arm's generative process,
including how the domain signal enters it. Every configuration is sized on the corpus-wide
vocabulary of $11$ node and $5$ edge classes and a maximum graph size of $187$ nodes.

\begin{table}[!htbp]
\centering
\footnotesize
\setlength{\tabcolsep}{2.5pt}
\begin{tabular}{l>{\raggedright\arraybackslash}p{3.4cm}ccccc}
\toprule
Model & Backbone (components, M) & Layers & Hidden & Heads & FFN & Params (M) \\
\midrule
\multicolumn{7}{l}{\emph{Graph transformers with node\,/\,edge\,/\,global streams (widths listed $X/E/y$)}} \\
DiG$^{2}$ & graph transformer & 12 & 256/64/128 & 8 & 256/128/2048 & 18 \\
DeFoG$^{\dagger}$ & graph transformer & 12 & 256/64/128 & 8 & 256/128/2048 & 18  \\
GraphBSI$^{\dagger}$ & graph transformer & 12 & 256/64/128 & 8 & 256/128/2048 & 18 \\
DiGress$^{\dagger}$ & graph transformer & 12 & 256/64/128 & 8 & 256/128/2048 & 18 \\
LGGM & graph transformer & 12 & 256/64/128 & 8 & 256/128/2048 & 18 \\
\midrule
\multicolumn{7}{l}{\emph{Other score and diffusion networks (two-network arms listed first\,/\,second)}} \\
HOG-Diff$^{\dagger}$ & GNN $+$ attention score net per phase (7.74\,/\,10.29) & 5\,/\,5 & 320 & 8 & --- & 18 \\
GDSS$^{\dagger}$ & score nets $s_X$, GMH (6.93)\,/\,$s_A$, GCN (11.37) & 3\,/\,8 & 256 & 8 & --- & 18 \\
Pard$^{\dagger}$ & PPGN diffusion (18.34)\,/\,PPGN block-size predictor (3.46) & 8\,/\,6 & 512\,/\,256 & 8 & --- & 22 \\
LGD$^{\dagger}$ & frozen graph-transformer autoencoder (1.05)\,/\,latent denoiser (25.88) & 5\,/\,8 & 96\,/\,256 & 4\,/\,16 & --- & 27 \\
\midrule
\multicolumn{7}{l}{\emph{Autoregressive and adversarial}} \\
GraphRNN$^{\dagger}$ & graph-level GRU (16.24)\,/\,edge-level GRU (1.98)$^{e}$ & 5\,/\,5 & 736\,/\,256 & --- & --- & 18 \\
G2PT & GPT decoder & 12 & 768 & 12 & 3072 & 89 \\
SPECTRE$^{\dagger}$ & $\lambda$, SON, PPGN generators (10.16)\,/\,discriminators (7.72) & 8 & 216 & --- & --- & 18 \\
\bottomrule
\end{tabular}
\caption{Backbone architecture and measured size of every arm. $^{\dagger}$ as in
Table~\ref{tab:benchmark}. $^{\dagger}$ indicates a multi-domain aware variant. All four DiGress-family arms and DiG$^{2}$ also receive RRWP features
($k{=}20$); Pard, GraphRNN and G$^2$PT ignore them.}
\label{tab:baseline-arch}
\end{table}

\begin{table}[!htbp]
\centering
\footnotesize
\setlength{\tabcolsep}{3pt}
\begin{tabular}{l>{\raggedright\arraybackslash}p{2.5cm}>{\raggedright\arraybackslash}p{5.5cm}>{\raggedright\arraybackslash}p{3.1cm}}
\toprule
Model & Generative process & Model-specific hyperparameters & Domain conditioning \\
\midrule
DeFoG$^{\dagger}$ & discrete flow matching (CTMC) & marginal prior; stochasticity $\eta{=}0$, target guidance $\omega{=}0$; \texttt{polydec} time distortion (train and sample); $T{=}200$; edge loss weight $5$ & one-hot, broadcast to node, edge and global channels \\
GraphBSI$^{\dagger}$ & Bayesian sample inference & $\beta$: $3 \to 20$, $\beta_0{=}1$; $\gamma{=}200$ (tree $100$); $\rho{=}1$; OU sampler (EM for planar, SBM, tree); $T{=}200$; edge loss weight $5$ & one-hot, broadcast to node, edge and global channels \\
DiGress$^{\dagger}$ & discrete diffusion & marginal transitions; cosine schedule; $200$ steps (train and sample); edge loss weight $5$ & one-hot prompt embedding, projected into node and edge channels \\
LGGM & discrete diffusion & as DiGress$^{\dagger}$ & 768-d BERT text-prompt embedding, same pathway \\
\midrule
HOG-Diff$^{\dagger}$ & two-phase higher-order score SDE & phase 1 VPSDE, linear $\beta \in [0.1, 1.0]$; phase 2 OU bridge, cosine, $\sigma_{\max}{=}1.0$; $200$ scales per phase; random-walk depth $16$; spectral basis from a $512$-graph donor pool & additive bias on the time embedding and node pathway \\
GDSS$^{\dagger}$ & joint score SDE on $(\mX, \mA)$ & $\mX$: VP, $\beta \in [0.1, 1.0]$; $\mA$: VE, $\sigma \in [0.2, 1.0]$; $200$ scales; Euler predictor $+$ Langevin corrector (SNR $0.1$, scale $1.0$, $1$ step); attention channels $2/8/4$, $3$ linears & additive bias after the first layer of both score nets \\
Pard$^{\dagger}$ & block-wise autoregressive diffusion & degree-ordered blocks, max hops $1$; $50$ diffusion steps per block; two-stage training (block-size predictor separate); node count bounded by the domain's size prior & one-hot through Pard's conditioning blocks, node and edge \\
LGD$^{\dagger}$ & latent diffusion & $32$-d latent per node and edge; $x_0$-parameterized DDPM, $1000$ training steps; DDIM $200$ at sampling & domain code from the frozen autoencoder's label embedder, cross-attention (dim $96$) \\
\midrule
GraphRNN$^{\dagger}$ & autoregressive, BFS adjacency sequence & max previous nodes $186$; generation cap $200$ nodes; embedding $384$ & one-hot initializes the graph-level hidden state \\
G2PT & autoregressive, graph tokens & BFS edge-list tokenization; vocabulary $229$; context $5120$; node permutation augmentation; node count predicted; temperature $1.0$, no top-$k$ & native per-domain prefix token \\
SPECTRE$^{\dagger}$ & spectrum-first WGAN-GP & eigenvalues $\to$ eigenvectors $\to$ adjacency; $k{=}4$ eigenpairs; noise dim $100$; gradient penalty $5$; $3$ SON rotations & one-hot concatenated to every noise MLP input \\
\bottomrule
\end{tabular}
\caption{Parameters of each arm's generative process, excluding the backbone
(Table~\ref{tab:baseline-arch}) and the shared training protocol.
Values follow each method's published configuration except where noted; DiG$^{2}$'s are in
Table~\ref{tab:gdfm-hparams}. $^{\dagger}$Domain conditioning was added by us, because the
upstream method is unconditional or conditions only on a global vector.}
\label{tab:baseline-gen}
\end{table}

%% file: sections/appendix/ablation.tex
\section{Ablation Study: Extra Details}
\label{app:extra_ablation}

\textbf{EMA prototypes ensure structural separability.}
Removing the EMA prototype embedding produces the largest overall structural
degradation in the current ablation. Planar V.U.N. falls from 100.0\% to
77.5\%, while SBM drops from 85.0\% to 42.5\%. Their MMD Ratios simultaneously
increase from 2.30 to 7.65 and from 1.80 to 4.82. Molecular validity also falls
by 7.44 points on QM9 and 3.54 points on ZINC. The effect is therefore not
limited to one graph family, a dedicated prototype representation is needed for
the shared denoiser to keep heterogeneous modes separated.

\textbf{Domain-specific priors prevent early cross-domain mixing.}
Replacing the per-domain corruption endpoints with a shared prior reduces
Planar and SBM V.U.N. by 17.5 and 15.0 points and approximately doubles their
MMD Ratios. The degradation persists on QM9 and ZINC, where validity falls by
6.06 and 3.04 points. Since this intervention changes the starting distribution
rather than the denoiser architecture, the result indicates that domain
separation is already consequential at the noisy endpoint, meaning the reverse process
is easier to control when it begins from a prior matched to the requested
domain.

\textbf{Vocabulary restriction primarily protects distributional quality.}
Removing the per-domain cross-entropy mask causes smaller validity losses than
removing the EMA embedding, but sharply worsens structural distribution
matching. PLANAR retains 90.0\% V.U.N. while its MMD Ratio rises from 2.30 to
6.37, SBM similarly moves from 1.80 to 3.48. This distinction is useful. Structural validity can remain high even when probability mass leaks onto
classes outside the domain's support. Restricting the training and sampling
support therefore acts as a semantic constraint rather than merely an
optimization convenience.

\textbf{Multi-depth denoising matters most for structurally difficult domains.}
Removing the depth mixture has little effect on the current molecular validity
numbers, but degrades SBM and TREE accuracy by 12.5 and 7.5 points and raises
their MMD Ratios from 1.80 to 5.13 and from 1.56 to 2.78. This pattern is
consistent with the motivation for exposing reconstructions at multiple
transformer depths, different graph families require distinctive denoising
effort. Taken together, the ablation results indicate that DiG$^{2}$ does not
obtain its performance from a single conditioning trick. Domain-specific
endpoints, support constraints, internal conditioning, and depth adaptation
make complementary contributions to maintaining separation.

%% file: sections/appendix/zero_shot.tex
\section{$\alpha$-interpolation: Extra Details}
\label{app:zero_shot}

\textbf{Baseline Testing Details.}
DiG$^2$ and LGGM generate at node counts drawn from the target band's histogram.
GenMol controls size through its token-length pool, DemoDiff cannot condition
on size, so its molecules are filtered to the target band after sampling.
We therefore distinguish \emph{raw validity}, the fraction of all attempts that
sanitize, from \emph{in-band yield}, the fraction of attempts that are both valid and
inside the target band; for size-conditioned generators the two coincide.

\subsection{Zero-shot validity of $\alpha$-interpolated molecules}

\textbf{Zero-shot validity.}
DiG$^2$ generates valid molecules along the entire path. On GuacaMol 10--14 its
validity is 0.82 at the QM9 endpoint and 0.95--0.96 for every
$\alpha\ge 0.5$. LGGM, whose prompt blend is the closest mechanism to ours,
collapses away from ZINC: 0.17 at $\alpha=0$ and 0.32 at $\alpha=0.5$ on
GuacaMol 10--14, and 0.06--0.07 on MOSES 15--19 for $\alpha\le 0.5$.
DemoDiff's raw validity (0.85--0.93) is close to DiG$^2$'s, but because it
cannot condition on size, only 15--28\% of its attempts land in GuacaMol 10--14
and 28--39\% in MOSES 15--19.
The one region where DiG$^2$ is weaker is the QM9 endpoint on MOSES 15--19
(0.56): QM9 conditioning asked to build molecules twice QM9's size limit.

\begin{figure}[t]
    \centering
    \includegraphics[width=\linewidth]{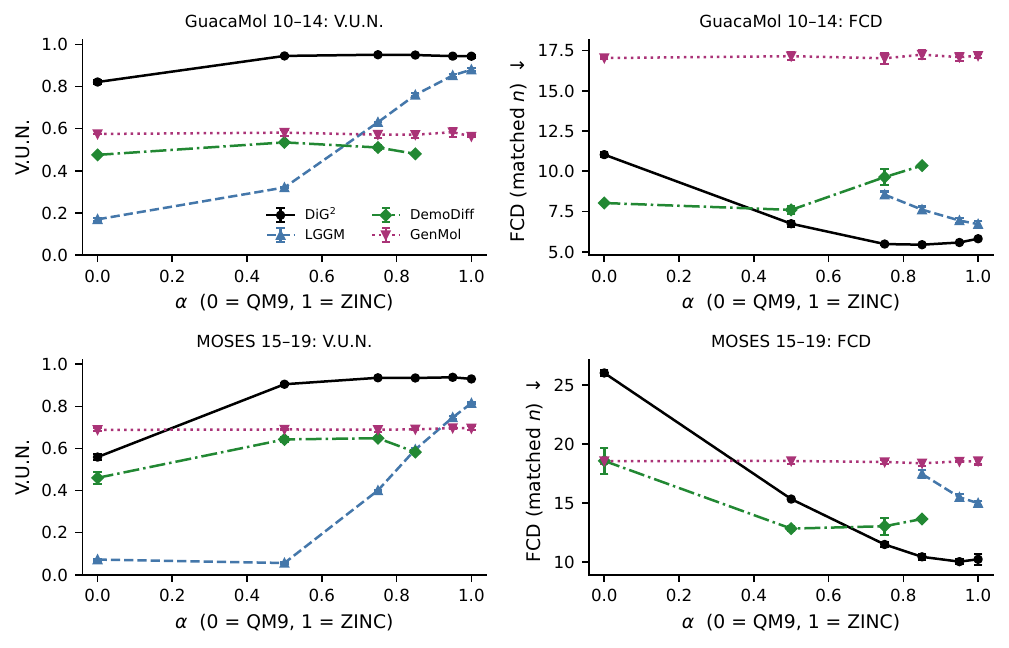}
    \vskip -1em
    \caption{
    \textbf{Zero-shot molecule generation along the QM9$\to$ZINC axis (size-conditioned protocol).}
    \emph{Left:} V.U.N. DemoDiff and GenMol reach the band by rejection, so their
    V.U.N. is measured on the unfiltered pool and is shared across bands; their
    in-band yield is given in the text and Appendix~\ref{app:zero_shot}.
    \emph{Right:} FCD to the target band. Baselines are subsampled to DiG$^2$'s
    valid-molecule count in each cell. Cells with fewer than 1{,}000 valid molecules
    are omitted (LGGM at low $\alpha$).
    Mean $\pm$ s.d.\ over seeds. No target molecule is used for training or tuning.
    }
    \label{fig:zero_shot_interpolation}
\end{figure}

\input{figures/caption_logp_target_notes}

\textbf{Secondary metrics.}
Tables~below list every cell. DemoDiff's uniqueness is computed over all
generated SMILES rather than its deduplicated output.

\input{figures/table_zero_shot_metrics}

\textbf{Property-dial cells.} Tables~\ref{tab:property_target_raw} and~\ref{tab:logp_target_raw}
list every cell behind Figures~\ref{fig:property_target} and~\ref{fig:logp_target}. The plotted
values are taken from each figure's build record, so they match the curves exactly.

\begin{table}[h]
\centering\scriptsize
\caption{Per-cell values behind Figure~\ref{fig:property_target} (GuacaMol band 15--19). QED is the
mean over a cell's molecules, then mean $\pm$ s.d.\ over seeds, as plotted (a single-seed cell shows
its value alone). $n$\,/\,seed: molecules
scored per seed. DiG$^2$ writes $500$ generations per seed; DemoDiff's cells are resampled to the
reference heavy-atom composition and GenMol's are filtered to $15$--$19$ heavy atoms, so their $n$ is
what remains. The $\alpha^*$ rows are the calibrated DiG$^2$ cells for the p25 ($0.6225$) and p50
($0.738$) targets. p75 ($0.8222$) is out of reach. The figure draws $0\le\alpha\le1$ (DiG$^2$'s
$\alpha=0$ lies below its axis); the $\alpha>1$ cells are listed for completeness.}
\label{tab:property_target_raw}
\input{figures/table_property_target_raw}
\end{table}

\begin{table}[h]
\centering\scriptsize
\caption{Per-cell values behind Figure~\ref{fig:logp_target} (MOSES band 20--27). QED and logP are
means over a cell's molecules, then mean $\pm$ s.d.\ over seeds (a single-seed cell shows its value
alone). DiG$^2$: $n$ is the valid molecules
per seed out of $2{,}000$ generated, and Valid is their fraction. $^{\dagger}$below the $0.70$
validity floor, generated but not drawn. DemoDiff: $n$ after resampling to the reference heavy-atom
composition. GenMol: $n$ after filtering to $20$--$27$ heavy atoms.}
\label{tab:logp_target_raw}
\input{figures/table_logp_target_raw}
\end{table}

\newpage

\subsection{DiG$^{2}$ QED Generated Samples}
\label{app:property_grid}

\begin{figure}[h]
    \centering
    \includegraphics[width=\linewidth]{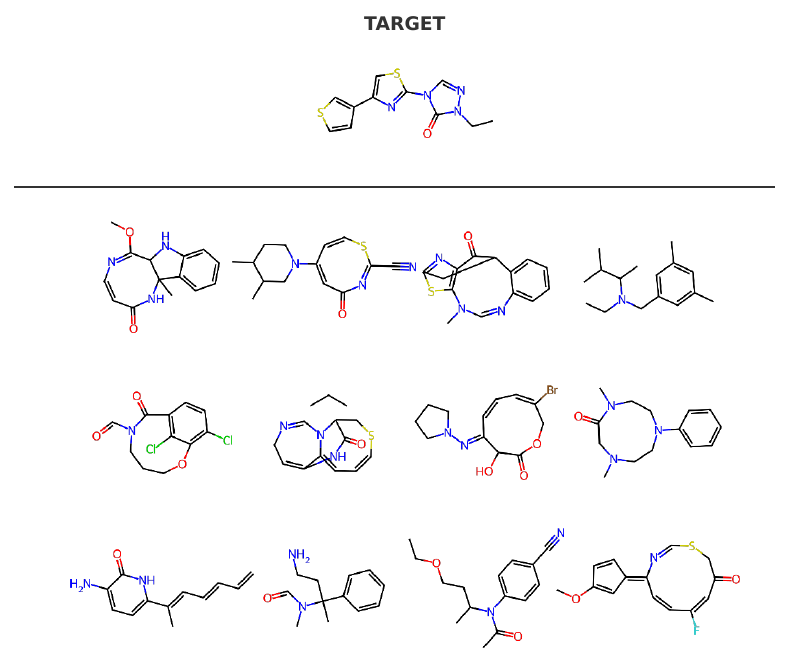}
    \caption{
    \textbf{QED target - DiG$^{2}$ Samples.} Top: the real GuacaMol band-15--19
    molecule at the p50 target. Below the rule: DiG$^2$ generations at $\alpha^*=0.793$, the twelve
    closest to the target.
    }
    \label{fig:property_target_grid}
\end{figure}

\newpage

\subsection{Multi-domain samples}
\label{app:domain_grid}

\begin{figure}[h]
    \centering
    \includegraphics[width=\linewidth]{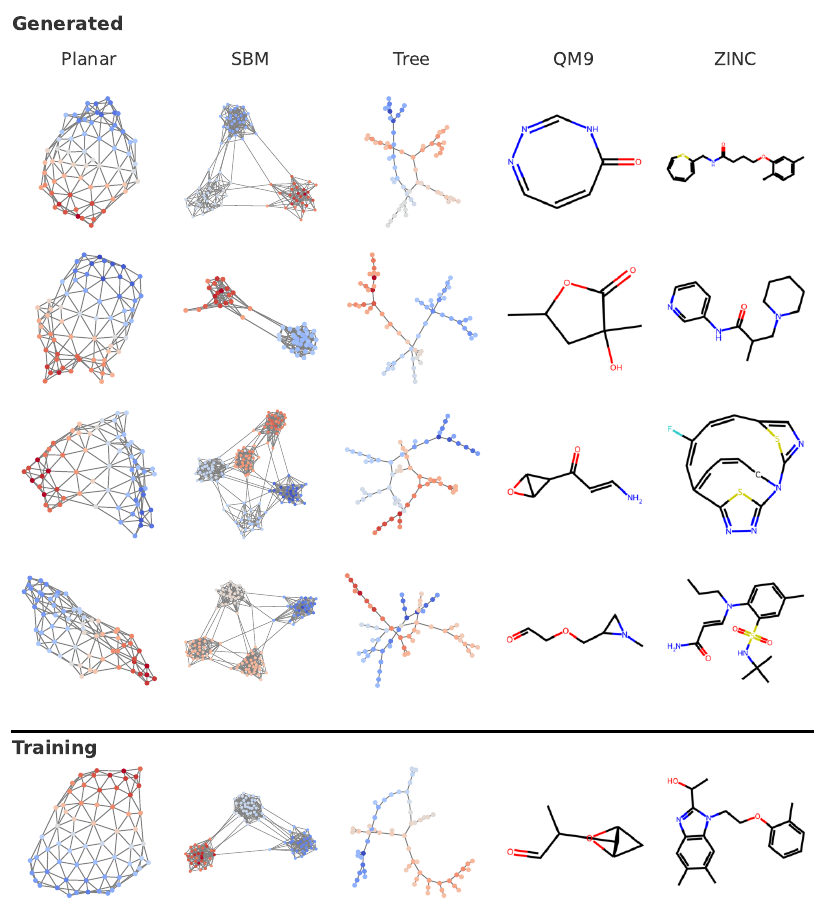}
    \caption{
    \textbf{DiG$^2$ generations across all five domains, over real training graphs.} One column per
    domain. The top four rows are generations from a single DiG$^{2}$ checkpoint, the bottom row
    (below the rule) is a reference graph from the corresponding domain.
    }
    \label{fig:domain_grid}
\end{figure}

%% file: figures/caption_logp_target_notes.tex
\newcommand{\lpBand}{20--27}
\newcommand{\lpSrc}{moses}
\newcommand{\lpRefN}{147{,}449}
\newcommand{\lpRefScored}{20{,}000}
\newcommand{\lpFloor}{0.70}
\newcommand{\lpTargetPlo}{1.9862}
\newcommand{\lpAlphaStarPlo}{0.5150}
\newcommand{\lpTargetPmid}{2.6367}
\newcommand{\lpAlphaStarPmid}{0.8818}
\newcommand{\lpTargetPhi}{3.1837}
\newcommand{\lpAlphaStarPhi}{1.2282}
\newcommand{\lpSpanDigTwo}{3.4625}
\newcommand{\lpSpanGenmol}{0.0847}
\newcommand{\lpSpanDemodiff}{0.3133}
\newcommand{\lpAlphaMax}{2}
\newcommand{\lpExtPassing}{1.25, 1.5, 1.75}
\newcommand{\lpExtExcluded}{2}
\newcommand{\lpLogpAtOne}{2.830}
\newcommand{\lpLogpMax}{4.163}
\newcommand{\lpTpsaAtOne}{60.25}
\newcommand{\lpTpsaMin}{46.28}
\newcommand{\lpQedAtOne}{0.791}
\newcommand{\lpQedMaxExt}{0.790}
\newcommand{\lpQedAtAlphaMax}{0.652}
\newcommand{\lpProjPeakAlpha}{1.5}
\newcommand{\lpProjPeak}{1.121}
\newcommand{\lpProjMoses}{1.051}
\newcommand{\lpOffAxisMoses}{1.48}
\newcommand{\lpProjGuacamol}{0.913}
\newcommand{\lpOffAxisGuacamol}{2.54}
\newcommand{\lpGenmolAlphaMax}{1}
\newcommand{\lpDemodiffAlphaMax}{0.85}

%% file: figures/table_zero_shot_metrics.tex
\begin{table}[h]
\centering\scriptsize
\caption{Zero-shot cells, GuacaMol 10–14 (mean over seeds). QED/SA $W_1$ are Wasserstein distances to the target's property distribution (lower is closer).}
\begin{tabular}{lcccccccccc}
\toprule
Model & $\alpha$ & seeds & $n_{\mathrm{valid}}$ & Valid & Uniq & Novel & Filters & QED $W_1$ & SA $W_1$ & SNN \\
\midrule
DiG$^2$ & 0.00 & 3 & 1643 & 0.822 & 1.000 & 1.000 & 0.584 & 0.083 & 2.219 & 0.235 \\
DiG$^2$ & 0.50 & 3 & 1896 & 0.948 & 1.000 & 0.997 & 0.654 & 0.032 & 1.266 & 0.268 \\
DiG$^2$ & 0.75 & 3 & 1916 & 0.958 & 1.000 & 0.993 & 0.705 & 0.040 & 0.798 & 0.289 \\
DiG$^2$ & 0.85 & 3 & 1915 & 0.958 & 1.000 & 0.992 & 0.720 & 0.050 & 0.680 & 0.295 \\
DiG$^2$ & 0.95 & 3 & 1908 & 0.954 & 1.000 & 0.990 & 0.735 & 0.056 & 0.574 & 0.300 \\
DiG$^2$ & 1.00 & 3 & 1905 & 0.953 & 0.999 & 0.991 & 0.738 & 0.061 & 0.531 & 0.301 \\
LGGM & 0.00 & 3 & 337 & 0.169 & 1.000 & 1.000 & 0.770 & 0.073 & 2.209 & 0.252 \\
LGGM & 0.50 & 3 & 640 & 0.320 & 1.000 & 0.999 & 0.758 & 0.056 & 2.341 & 0.236 \\
LGGM & 0.75 & 3 & 1262 & 0.631 & 1.000 & 0.998 & 0.725 & 0.040 & 1.528 & 0.256 \\
LGGM & 0.85 & 3 & 1525 & 0.763 & 1.000 & 0.997 & 0.725 & 0.037 & 1.282 & 0.265 \\
LGGM & 0.95 & 3 & 1711 & 0.856 & 1.000 & 0.996 & 0.725 & 0.037 & 1.011 & 0.275 \\
LGGM & 1.00 & 3 & 1766 & 0.883 & 1.000 & 0.996 & 0.730 & 0.040 & 0.905 & 0.279 \\
DemoDiff & 0.00 & 1 & 2007 & 0.850 & 0.602 & 0.928 & 0.791 & 0.035 & 0.297 & 0.387 \\
DemoDiff & 0.50 & 2 & 2005 & 0.892 & 0.659 & 0.909 & 0.773 & 0.042 & 0.394 & 0.399 \\
DemoDiff & 0.75 & 2 & 2005 & 0.915 & 0.615 & 0.907 & 0.817 & 0.066 & 0.421 & 0.399 \\
DemoDiff & 0.85 & 1 & 2002 & 0.930 & 0.576 & 0.895 & 0.791 & 0.060 & 0.404 & 0.398 \\
GenMol & 0.00 & 3 & 1035 & 0.999 & 0.639 & 0.900 & 0.777 & 0.088 & 0.851 & 0.449 \\
GenMol & 0.50 & 3 & 1025 & 0.999 & 0.646 & 0.900 & 0.771 & 0.088 & 0.834 & 0.446 \\
GenMol & 0.75 & 3 & 1034 & 0.999 & 0.643 & 0.889 & 0.773 & 0.087 & 0.829 & 0.444 \\
GenMol & 0.85 & 3 & 1022 & 0.999 & 0.640 & 0.893 & 0.765 & 0.086 & 0.840 & 0.448 \\
GenMol & 0.95 & 3 & 1033 & 0.999 & 0.647 & 0.902 & 0.786 & 0.088 & 0.839 & 0.446 \\
GenMol & 1.00 & 3 & 1018 & 0.999 & 0.633 & 0.889 & 0.785 & 0.089 & 0.842 & 0.449 \\
\bottomrule
\end{tabular}
\end{table}

\begin{table}[h]
\centering\scriptsize
\caption{Zero-shot cells, MOSES 15–19 (mean over seeds). QED/SA $W_1$ are Wasserstein distances to the target's property distribution (lower is closer).}
\begin{tabular}{lcccccccccc}
\toprule
Model & $\alpha$ & seeds & $n_{\mathrm{valid}}$ & Valid & Uniq & Novel & Filters & QED $W_1$ & SA $W_1$ & SNN \\
\midrule
DiG$^2$ & 0.00 & 3 & 1118 & 0.559 & 1.000 & 1.000 & 0.355 & 0.249 & 3.287 & 0.230 \\
DiG$^2$ & 0.50 & 3 & 1811 & 0.905 & 1.000 & 0.999 & 0.489 & 0.125 & 1.957 & 0.291 \\
DiG$^2$ & 0.75 & 3 & 1874 & 0.937 & 1.000 & 0.999 & 0.601 & 0.075 & 1.370 & 0.328 \\
DiG$^2$ & 0.85 & 3 & 1877 & 0.939 & 1.000 & 0.996 & 0.635 & 0.065 & 1.169 & 0.340 \\
DiG$^2$ & 0.95 & 3 & 1884 & 0.942 & 1.000 & 0.995 & 0.666 & 0.058 & 1.004 & 0.350 \\
DiG$^2$ & 1.00 & 3 & 1872 & 0.936 & 1.000 & 0.994 & 0.672 & 0.056 & 0.973 & 0.353 \\
LGGM & 0.00 & 3 & 144 & 0.072 & 1.000 & 1.000 & 0.556 & 0.155 & 3.406 & 0.223 \\
LGGM & 0.50 & 3 & 111 & 0.056 & 1.000 & 1.000 & 0.538 & 0.143 & 3.729 & 0.230 \\
LGGM & 0.75 & 3 & 801 & 0.401 & 1.000 & 1.000 & 0.460 & 0.097 & 2.513 & 0.268 \\
LGGM & 0.85 & 3 & 1190 & 0.595 & 1.000 & 1.000 & 0.479 & 0.088 & 2.180 & 0.279 \\
LGGM & 0.95 & 3 & 1495 & 0.748 & 1.000 & 0.999 & 0.526 & 0.085 & 1.809 & 0.293 \\
LGGM & 1.00 & 3 & 1630 & 0.815 & 1.000 & 1.000 & 0.551 & 0.084 & 1.659 & 0.298 \\
DemoDiff & 0.00 & 3 & 2000 & 0.858 & 0.540 & 0.993 & 0.636 & 0.116 & 0.324 & 0.372 \\
DemoDiff & 0.50 & 3 & 2000 & 0.896 & 0.728 & 0.987 & 0.687 & 0.092 & 0.231 & 0.426 \\
DemoDiff & 0.75 & 2 & 2000 & 0.915 & 0.718 & 0.987 & 0.690 & 0.086 & 0.208 & 0.424 \\
DemoDiff & 0.85 & 1 & 2000 & 0.930 & 0.634 & 0.987 & 0.661 & 0.094 & 0.248 & 0.427 \\
GenMol & 0.00 & 3 & 1261 & 0.996 & 0.744 & 0.929 & 0.700 & 0.031 & 0.593 & 0.503 \\
GenMol & 0.50 & 3 & 1252 & 0.996 & 0.746 & 0.929 & 0.691 & 0.031 & 0.583 & 0.505 \\
GenMol & 0.75 & 3 & 1264 & 0.996 & 0.743 & 0.931 & 0.694 & 0.031 & 0.577 & 0.504 \\
GenMol & 0.85 & 3 & 1272 & 0.995 & 0.746 & 0.930 & 0.698 & 0.031 & 0.577 & 0.504 \\
GenMol & 0.95 & 3 & 1257 & 0.997 & 0.742 & 0.943 & 0.695 & 0.030 & 0.583 & 0.503 \\
GenMol & 1.00 & 3 & 1245 & 0.995 & 0.748 & 0.934 & 0.692 & 0.030 & 0.583 & 0.505 \\
\bottomrule
\end{tabular}
\end{table}

%% file: figures/table_property_target_raw.tex
\begin{tabular}{lcrrrc}
\toprule
Model & $\alpha$ & seeds & $n$ / seed & heavy atoms & QED \\
\midrule
DiG$^2$ & 0.00 & 5 & 500 & 17.4 & $0.559\pm0.006$ \\
DiG$^2$ & 0.25 & 5 & 500 & 17.5 & $0.620\pm0.006$ \\
DiG$^2$ & 0.50 & 5 & 500 & 17.5 & $0.684\pm0.003$ \\
DiG$^2$ & 0.75 & 5 & 500 & 17.5 & $0.732\pm0.003$ \\
DiG$^2$ & 0.85 & 5 & 500 & 17.5 & $0.746\pm0.004$ \\
DiG$^2$ & 0.95 & 5 & 500 & 17.5 & $0.753\pm0.006$ \\
DiG$^2$ & 1.00 & 5 & 500 & 17.5 & $0.758\pm0.007$ \\
DiG$^2$ & 1.25 & 5 & 500 & 17.5 & $0.763\pm0.006$ \\
DiG$^2$ & 1.50 & 5 & 500 & 17.5 & $0.762\pm0.003$ \\
DiG$^2$ & 2.00 & 5 & 500 & 17.5 & $0.715\pm0.002$ \\
DiG$^2$, $\alpha^*$ (target 0.6225) & 0.267 & 3 & 500 & 17.5 & $0.627\pm0.005$ \\
DiG$^2$, $\alpha^*$ (target 0.738) & 0.793 & 3 & 500 & 17.5 & $0.739\pm0.003$ \\
\midrule
DemoDiff & 0.00 & 3 & 1051 & 17.5 & $0.704\pm0.015$ \\
DemoDiff & 0.50 & 3 & 1527 & 17.5 & $0.735\pm0.014$ \\
DemoDiff & 0.75 & 2 & 1471 & 17.5 & $0.741\pm0.004$ \\
DemoDiff & 0.85 & 1 & 1457 & 17.5 & $0.733$ \\
\midrule
GenMol & 0.00 & 3 & 1476 & 17.2 & $0.803\pm0.002$ \\
GenMol & 0.50 & 3 & 1502 & 17.2 & $0.802\pm0.001$ \\
GenMol & 0.75 & 3 & 1500 & 17.2 & $0.803\pm0.002$ \\
GenMol & 0.85 & 3 & 1458 & 17.2 & $0.803\pm0.001$ \\
GenMol & 0.95 & 3 & 1475 & 17.2 & $0.804\pm0.001$ \\
GenMol & 1.00 & 3 & 1454 & 17.2 & $0.803\pm0.001$ \\
\bottomrule
\end{tabular}

%% file: figures/table_logp_target_raw.tex
\begin{tabular}{lcrrrrcc}
\toprule
Model & $\alpha$ & seeds & $n$ / seed & Valid & heavy atoms & QED & logP \\
\midrule
DiG$^2$ & 0.00 & 3 & 1604 & 0.802 & 22.6 & $0.580\pm0.003$ & $0.701\pm0.028$ \\
DiG$^2$ & 0.50 & 3 & 1819 & 0.909 & 22.6 & $0.738\pm0.004$ & $1.958\pm0.015$ \\
DiG$^2$ & 0.75 & 3 & 1851 & 0.925 & 22.6 & $0.776\pm0.005$ & $2.432\pm0.022$ \\
DiG$^2$ & 0.85 & 3 & 1863 & 0.931 & 22.6 & $0.780\pm0.001$ & $2.581\pm0.022$ \\
DiG$^2$ & 0.95 & 3 & 1853 & 0.926 & 22.6 & $0.789\pm0.002$ & $2.756\pm0.023$ \\
DiG$^2$ & 1.00 & 3 & 1860 & 0.930 & 22.6 & $0.791\pm0.001$ & $2.830\pm0.026$ \\
DiG$^2$ & 1.25 & 3 & 1817 & 0.908 & 22.6 & $0.790\pm0.003$ & $3.217\pm0.018$ \\
DiG$^2$ & 1.50 & 3 & 1779 & 0.889 & 22.6 & $0.775\pm0.002$ & $3.632\pm0.046$ \\
DiG$^2$ & 1.75 & 3 & 1625 & 0.812 & 22.6 & $0.733\pm0.002$ & $4.163\pm0.050$ \\
DiG$^2$ & 2.00$^{\dagger}$ & 3 & 1335 & 0.668 & 22.6 & $0.652\pm0.003$ & $4.734\pm0.049$ \\
\midrule
DemoDiff & 0.50 & 3 & 851 &  & 22.6 & $0.673\pm0.018$ & $3.670\pm0.200$ \\
DemoDiff & 0.75 & 2 & 809 &  & 22.6 & $0.677\pm0.018$ & $3.549\pm0.015$ \\
DemoDiff & 0.85 & 1 & 946 &  & 22.6 & $0.686$ & $3.357$ \\
\midrule
GenMol & 0.00 & 3 & 1192 &  & 23.3 & $0.689\pm0.007$ & $2.630\pm0.031$ \\
GenMol & 0.50 & 3 & 1178 &  & 23.3 & $0.693\pm0.012$ & $2.672\pm0.077$ \\
GenMol & 0.75 & 3 & 1136 &  & 23.3 & $0.691\pm0.006$ & $2.678\pm0.014$ \\
GenMol & 0.85 & 3 & 1173 &  & 23.3 & $0.693\pm0.001$ & $2.690\pm0.020$ \\
GenMol & 0.95 & 3 & 1160 &  & 23.2 & $0.694\pm0.003$ & $2.651\pm0.055$ \\
GenMol & 1.00 & 3 & 1161 &  & 23.2 & $0.697\pm0.000$ & $2.606\pm0.025$ \\
\bottomrule
\end{tabular}

%% file: sections/appendix/initial.tex
\section{Initial Investigation: Graph-Level Conditioning}
\label{app:initial}

This section details results from our initial investigation into promoting multi-domain awareness in single-domain graph generation models. Within Table~\ref{tab:conditioning}, we demonstrate these architectures carry conditioning
in a single graph-level vector $\vy$. Perturbing the domain one-hot in $\vy$ moves the network logit
output ($\hat{\mX}$) by at most, $\max|\Delta\hat{\mX}| = 0.06$. Whereas, perturbing the node features $\mX$ moves it by $5.14$. The
conditioning pathway is roughly $85\times$ weaker than the data pathway, making it too weak to override an $839{:}1$ prior. LGGM, on identical
data, does not suffer this, given its native pathway broadcasts the conditioning vector onto every node
and edge before the first layer.

\begin{table}[h]
\centering
\small
\begin{tabular}{lrrl}
\toprule
Domain & $n$ (mean) & Avg.\ degree & Node-class histogram \\
\midrule
\multicolumn{4}{l}{\emph{Graph-level conditioning only}} \\
planar & 64.0 & 15.61 & C 0.87, N 0.08, S 0.04, O 0.02 \\
sbm    & 115.4 & 50.38 & C 0.63, \texttt{generic} 0.24, N 0.06, O 0.03 \\
tree   & 64.0 & 15.41 & C 0.85, N 0.10, O 0.03, S 0.03 \\
\midrule
\multicolumn{4}{l}{\emph{Broadcast to node and edge features}} \\
planar & 64.0 & 4.77 & \texttt{generic} \textbf{1.00} \\
sbm    & 97.6 & 9.75 & \texttt{generic} 0.29, C 0.21, N 0.16, O 0.16 \\
tree   & 64.0 & 2.15 & \texttt{generic} \textbf{1.00} \\
\bottomrule
\end{tabular}
\caption{\emph{Why do baseline models fail at multi-domain graph generation?} The following table shows the emitted node classes from a baseline model, before and after the
conditioning signal is broadcast to node and edge features. The structural domains use only the
\texttt{generic} class, so any atom mass is conditioning leakage from the pooled molecular prior.
Node counts come from a histogram prior in both settings and are correct throughout, which
localizes the failure to graph content. Here we see the graph-level conditioning signal unable to prevent molecular vocabulary (i.e. C,N,S,O) from leaking into the \texttt{generic} class which PLANAR, SBM, and TREE require.}
\label{tab:conditioning}
\end{table}

%% file: sections/appendix/revision_finetune.tex
{
\subsection{Fine-tuning: pools, parity and measured effect}
\label{app:finetune}

\paragraph{Why fine-tune at all.} Fine-tuning is applied to every benchmarked arm to
improve performance on the structural domains, which are the hardest cells of
Table~\ref{tab:benchmark}. The reason each model uses the \emph{anchored} variant of
Algorithm~\ref{alg:anchor} is that performance is often unstable in graph generation, even on well-tuned models. The anchored KL term exists to bound that drift meaning
Table~\ref{tab:finetune} reports what effect it helps for each domain. However, performance with anchored fine-tuning is still subject to drift, as shown with the anchor fine-tuned DiGress and LGGM lowering Planar V.U.N by $12.5$ and
$7.5$ points, respectively.

\paragraph{Fine-tuning pools and baseline parity.} The structural datasets supply only $128$
training graphs each. Empirically, we found this is too small a signal for a $25$K-step fine-tune. We therefore draw an
additional pool of $8{,}192$ graphs per structural domain from each dataset's \emph{own random
generator}, the same generator that produced the original training split. No model samples are
used at any point. Every benchmarked arm is fine-tuned on these same pools, so no arm sees data
another arm does not. Novelty in Table~\ref{tab:novelty} is measured against the union of the
training split and the pool, not the training split alone. Although QM9 and ZINC are never fine-tuning targets, both improve after AFT (Table~\ref{tab:aft-frag}): the share of valid single-fragment molecules rises by $0.92$ (QM9) and $2.73$ (ZINC) points, whereas $25$K steps of unanchored fine-tuning yields no gain, consistent with the shared edge pathway improving connectivity.}

\begin{table}[h!]
\centering
\small
\begin{tabular}{llccccc}
\toprule
Model & & Planar & SBM & Tree & QM9 & ZINC \\
\midrule
\multirow{2}{*}{DiG$^{2}$}
 & Before & 100\% / 2.6 & 77.5\% / 4.1 & 95.0\% / 1.8 & 98.36\% / 0.41 & 91.94\% / 2.9 \\
 & After  & 100\% / 2.3 & 85.0\% / 1.8 & 97.5\% / 1.56 & 99.92\% / 0.29 & 95.46\% / 2.7 \\
\midrule
\multirow{2}{*}{DeFoG$^{\dagger}$}
 & Before & 77.5\% / 7.6 & 12.5\% / 8.4 & 57.5\% / 2.7 & 88.57\% / 0.83 & 90.41\% / 3.7 \\
 & After  & 95.0\% / 4.3 & 5.0\% / 6.9 & 62.5\% / 2.35 & 90.21\% / 0.49 & 92.93\% / 3.4 \\
\midrule
\multirow{2}{*}{DiGress$^{\dagger}$}
 & Before & 22.5\% / 12.9 & 27.5\% / 4.2 & 45.0\% / 2.30 & 92.79\% / 0.70 & 86.8\% / 5.1 \\
 & After  & 10.0\% / 7.8 & 37.5\% / 3.04 & 67.5\% / 1.81 & 95.43\% / 0.4 & 87.1\% / 4.7 \\
\midrule
\multirow{2}{*}{LGGM}
 & Before & 7.5\% / 20.1 & 10.0\% / 9.5 & 0.0\% / 9.6 & 85.9\% / 0.92 & 77.4\% / 9.07 \\
 & After  & 0.0\% / 76 & 12.5\% / 9.1 & 0.0\% / 8.78 & 86.3\% / 0.75 & 85.2\% / 8.55 \\
\bottomrule
\end{tabular}
\vskip -0.5em
\caption{Validity / MMD Ratio or FCD, before and after fine-tuning, on the same pools
for every arm. The ``After'' rows are the values reported in Table~\ref{tab:benchmark}. Fine-tuning
helps DiG$^{2}$ on every domain, but is not uniformly positive for the baselines: DiGress loses
$12.5$ points of Planar V.U.N. and LGGM loses $7.5$, which is the drift the anchored KL term of
Algorithm~\ref{alg:anchor} is designed to prevent.}
\label{tab:finetune}
\end{table}

\begin{table}[h!]
\centering
\small
\begin{tabular}{lccccc}
\toprule
DiG$^{2}$ & Planar & SBM & Tree & QM9 & ZINC \\
\midrule
Validity & 100\% & 85\% & 97.5\% & 99.92\% & 95.46\% \\
Uniqueness & 100\% & 100\% & 100\% & 97.28\% & 99.95\% \\
Novelty & 100\% & 100\% & 100\% & 50.15\% & 99.95\% \\
\bottomrule
\end{tabular}
\vskip -0.5em
\caption{ Uniqueness and Novelty alongside the headline metric of Table~\ref{tab:benchmark}. The
structural domains are perfectly unique and perfectly novel, measured against both the training
split and the $8{,}192$-graph fine-tuning pools, so none of those samples reproduce an input graph.
QM9's $50.15\%$ Novelty is expected: QM9 is an enumeration of a closed chemical space, and the
standard protocol scores novelty against the full enumeration rather than a held-out region.}
\label{tab:novelty}
\end{table}

\begin{table}[h!]
\centering
\small
\begin{tabular}{lcccc}
\toprule
AFT Stage & QM9/ZINC frag. w/ V. &	QM9/ZINC V., single frag. & QM9 V.U. & ZINC V.U.N \\
\midrule
Before &	0.41\%  /	17.0\% &	98.70\% /	74.9\% & 96.1\% & 91.9\% \\
After & 0.11\%  /	14.5\% &	99.82\% / 	81.0\% & 97.2\% & 95.4\% \\
w/o AFT	 & 0.47\%  /	17.6\% &	98.24\% / 75.1\% & 96.0\% & 92.7\% \\
\bottomrule
\end{tabular}
\vskip -0.5em
\caption{The percentage of fragmented yet valid molecules for QM9 and ZINC (frag/ w/ V.), the percentage of single fragment yet valid molecules for QM9 and ZINC, and then the respective V.U. and V.U.N. scores for QM9 and ZINC. \emph{w/o AFT}: $25$K steps of fine-tuning without the KL anchor. \emph{Note:} We separate QM9 and ZINC results with a '/' in the first two columns to conserve space.}
\label{tab:aft-frag}
\end{table}

\begin{table}[h!]
\centering
\small
\begin{tabular}{lccccc}
\toprule
DiG$^{2}$ & Planar & SBM & Tree & QM9 & ZINC \\
\midrule
V.     & 100\% & 85\% & 97.5\% & 99.92\% & 95.46\% \\
V.U.   & 100\% & 85\% & 97.5\% & 97.20\% & 95.41\% \\
V.U.N. & 100\% & 85\% & 97.5\% & 48.75\% & 95.36\% \\
\bottomrule
\end{tabular}
\vskip -0.5em
\caption{Valid (V.), valid and unique (V.U.), and valid, unique and novel (V.U.N.) samples
as a percentage of all generated samples, for the DiG$^{2}$ row of Table~\ref{tab:benchmark}. On
the structural domains every sample is unique and novel, so the three rows coincide. We found the saturated Uniqueness and Novelty pattern to be consistent across all tested baselines. \emph{Note:} QM9 enumerates the chemical training space, so QM9's V.U.N.\ is low given how Novelty penalizes V.U.N.}
\label{tab:mol-vun}
\end{table}


\begin{table}[h!]
\centering
\small
\begin{tabular}{lcccccc}
\toprule
 & \multicolumn{3}{c}{QM9} & \multicolumn{3}{c}{ZINC} \\
\cmidrule(lr){2-4}\cmidrule(lr){5-7}
Model & V. & V.U. & V.U.N. & V. & V.U. & V.U.N. \\
\midrule
\multicolumn{7}{l}{\emph{Table~\ref{tab:benchmark} evaluation with added Uniqueness and Novelty}} \\
DiG$^{2}$            & 99.9\% & 97.2\% & 48.8\% & 95.5\% & 95.4\% & 95.4\% \\
G2PT                 & 90.6\%  & $89.8\%$  & $39.7\%$  & $72.6\%$  & $71.7\%$  & $70.8\%$ \\
DeFoG$^{\dagger}$    & 90.2\% & $90.0\%$ & $47.9\%$ & 92.9\% & $ 91.7\%$ & $90.9\%$ \\
DiGress$^{\dagger}$  & 95.4\% & $93.8\%$ & $45.4\%$ & 87.1\%  & $86.4\%$  & $86.0\%$ \\
HOG-Diff$^{\dagger}$ & 93.2\% & $91.8\%$ & $33.2\%$ & 86.4\% & $82.8\%$ & $80.9\%$ \\
GraphBSI$^{\dagger}$ & 43.5\% & $33.1\%$ & $31.2\%$ & 42.2\% & $42.2\%$ & $42.2\%$ \\
LGGM                 & 86.3\%  & $86.2\%$  & $39.0\%$  & 85.2\%  & $85.0\%$  & $84.9\%$ \\
Pard$^{\dagger}$     & 47.5\% & $47.5\%$ & $42.6\%$ & 31.4\%  & $31.4\%$  & $31.4\%$ \\
GraphRNN$^{\dagger}$ & 47.5\% & $47.5\%$ & $47.5\%$ & 32.8\% & $32.8\%$ & $32.8\%$ \\
LGD$^{\dagger}$      & 34.2\%  & $34.2\%$  & $34.2\%$  & 37.9\% & $12.9\%$ & $11.4\%$ \\
SPECTRE$^{\dagger}$  & 17.3\%  & $ 15.7\%$  & $14.9\%$  & 9.5\%  & $9.3\%$  & $9.2\%$ \\
\midrule
\multicolumn{7}{l}{\emph{Table~\ref{tab:lggm-priors} evaluation with added Uniqueness and Novelty}} \\
LGGM$^{\dagger}$     & 91.8\% & 89.7\% & 45.2\% & 77.5\% & 77.4\% & 77.4\% \\
\bottomrule
\end{tabular}
\vskip -0.5em
\caption{Molecular V.U.\ and V.U.N.\ as a percentage of all generated molecules. \emph{Note:} QM9 enumerates the chemical training space, so QM9's V.U.N.\ is low given how Novelty penalizes V.U.N.}
\label{tab:mol-vun-baselines}
\end{table}

\subsection{ Guidance weight ablation}
\label{app:cfg}

Table~\ref{tab:cfg} isolates how much of DiG$^{2}$'s performance comes from classifier-free
guidance as opposed to its architecture. Removing guidance entirely ($g = 1$, the conditional
prediction alone) costs between $0$ and $5$ validity points, so the per-domain priors, prototype
conditioning and masked support carry the result on their own. Dropping the conditional branch as
well (unconditional sampling) is what actually collapses the model, and SBM is the domain most
exposed: it loses $40$ points relative to the calibrated weight. This matches the role guidance is
given in \Secref{sec:form-sample}, which is to separate domains that share a vocabulary and a size
range rather than to supply the separation itself.

\begin{table}[h!]
\centering
\small
\begin{tabular}{lccccc}
\toprule
CFG setting & Planar & SBM & Tree & QM9 & ZINC \\
\midrule
Calibrated $g(d)$ (Table~\ref{tab:benchmark}) & 100 / 2.3 & 85.0 / 1.8 & 97.5 / 1.56 & 99.92 / 0.29 & 95.46 / 2.67 \\
$g = 1$ (no guidance)                         & 97.5 / 2.4 & 82.5 / 3.5 & 92.5 / 1.8 & 99.74 / 0.31 & 95.94 / 2.36 \\
Unconditional                                 & 92.5 / 2.7 & 45.0 / 10.9 & 87.5 / 2.0 & 99.21 / 0.33 & 95.72 / 3.13 \\
\bottomrule
\end{tabular}
\vskip -0.5em
\caption{Validity / MMD Ratio or FCD across guidance settings. Guidance is a
refinement, not the mechanism, ablating it costs at most $5$ validity points, while removing
domain conditioning altogether costs $40$ on SBM.}
\label{tab:cfg}
\end{table}

\subsection{What the multi-depth gate learns}
\label{app:readout}

The gate of \Eqref{eq:combine} mixes readouts taken at depths $8$, $10$ and $12$.
To determine the effect of gated readout predictions on the benchmarked DiG$^2$ model within Table~\ref{tab:benchmark}, we run a post-hoc analysis on the DiG$^2$ checkpoint with separate readout prediction configurations. Table~\ref{tab:readout} compares the learned mixture against each fixed depth and against a uniform
average of the three. No single depth is best everywhere: depth $12$ is the strongest fixed choice
on SBM but the weakest on Tree and QM9, while depth $10$ is best on Tree and ZINC but weak on
Planar and SBM. The uniform average is dominated by the learned gate on every domain, which is the
behaviour the per-depth mechanism was introduced for.

\begin{table}[h!]
\centering
\small
\begin{tabular}{lccccc}
\toprule
Readout & Planar & SBM & Tree & QM9 & ZINC \\
\midrule
Learned gate (ours)          & 100\% & 85\% & 97.5\% & 99.92\% & 95.46\% \\
Depth 8 only                 & 65\% & 60\% & 95\% & 99.6\% & 95.1\% \\
Depth 10 only                & 80\% & 67.5\% & 100\% & 99.7\% & 96.0\% \\
Depth 12 only (final layer)  & 82.5\% & 80\% & 72.5\% & 91.7\% & 90.8\% \\
Uniform mean of depths       & 80\% & 80\% & 85\% & 98.0\% & 92.8\% \\
\bottomrule
\end{tabular}
\vskip -0.5em
\caption{ V.U.N. (structural) or Validity (molecular) under fixed readout depths, their uniform
mean, and the learned gate. Which depth matters is domain-dependent, and only the gate is strong
on all five domains at once.}
\label{tab:readout}
\end{table}

%% file: sections/appendix/revision_baselines.tex
{
\subsection{Training budget actually consumed}
\label{app:budget}

Evaluating every arm at a fixed 10-day wall-clock window ensures comparison on one standard, but
the arms differ in cost per epoch, so they do not all reach the same epoch count.
Table~\ref{tab:epochs} reports what each arm completed. Five arms (DeFoG$^{\dagger}$,
DiGress$^{\dagger}$, LGGM, GraphBSI$^{\dagger}$, GraphRNN$^{\dagger}$) reached the $10{,}000$-epoch
cap alongside DiG$^{2}$, so the headline comparison is not decided by the cutoff. Additional training with the slower arms did not indicate improved performance or convergence.

\begin{table}[h!]
\centering
\small
\begin{tabular}{lccc}
\toprule
Model & s / epoch & Max epochs by cutoff & Max epochs reached \\
\midrule
DiG$^{2}$              & 87.2  & 10,000 & 10,000 \\
DeFoG$^{\dagger}$      & 53.9  & 10,000 & 10,000 \\
DiGress$^{\dagger}$    & 47.7  & 10,000 & 10,000 \\
LGGM                   & 46.7  & 10,000 & 10,000 \\
GraphBSI$^{\dagger}$   & 46.6  & 10,000 & 10,000 \\
GraphRNN$^{\dagger}$   & 20.0  & 10,000 & 10,000 \\
G2PT                   & 167.4 & 5,167  & 7,199 \\
Pard$^{\dagger}$       & 167.0 & 5,274  & 7,199 \\
HOG-Diff$^{\dagger}$   & 110.2 & 7,849  & 7,849 \\
GDSS$^{\dagger}$       & 188.8 & 4,581  & 6,699 \\
LGD$^{\dagger}$        & 127.9 & 6,765  & 8,799 \\
SPECTRE$^{\dagger}$    & 489.5 & 1,770  & 2,699 \\
\bottomrule
\end{tabular}
\vskip -0.5em
\caption{Epochs each arm completed within the 10-day budget, and the furthest epoch reached when
training was extended beyond it.}
\label{tab:epochs}
\end{table}

\paragraph{SPECTRE Timing Investigation}

SPECTRE is the slowest arm and reached only
epoch $1{,}770$ inside the $10-$day window, where generated samples already collapsed onto hub structures. Training
SPECTRE for five more days to epoch $2{,}699$ and find that SPECTRE's structural accuracy stays at $0$ and FCD remains unmeasurable for the molecular regime. Therefore, indicating SPECTRE does not recover a multi-domain signal under this study's training setting (Table~\ref{tab:spectre}).

\begin{table}[h!]
\centering
\small
\begin{tabular}{lccccc}
\toprule
SPECTRE$^{\dagger}$ & Planar & SBM & Tree & QM9 & ZINC \\
\midrule
Table~\ref{tab:benchmark} (epoch 1,770) & 0.0 / 2766 & 0.0 / 374 & 0.0 / 461 & 17.3 / -- & 9.51 / -- \\
Extended (epoch 2,699)                  & 0.0 / 2961 & 0.0 / 365 & 0.0 / 351 & 14.6 / -- & 8.91 / -- \\
\bottomrule
\end{tabular}
\vskip -0.5em
\caption{SPECTRE after five additional days of training.}
\label{tab:spectre}
\end{table}

\paragraph{LGD Encoding Investigation}

LGD generates in the latent space of a
pre-trained autoencoder, so a weak autoencoder would by itself explain its low scores. We therefore
measured that autoencoder's reconstruction directly. It exceeds $95\%$ latent reconstruction
accuracy by epoch $400$ and is exact by epoch $500$, the checkpoint used for all LGD results
(Table~\ref{tab:lgd}). The trivial floor in that table is the accuracy of always predicting the
majority class, which the autoencoder clears by a wide margin on every domain. LGD's difficulty is
therefore in the latent diffusion stage for multi-domain generation instead of encoding failures.

\begin{table}[h!]
\centering
\small
\begin{tabular}{lcccccc}
\toprule
Domain & Node acc. & Real-edge acc. & All-pair acc. & Trivial floor & Nodes & Real edges \\
\midrule
Planar & 100\% & 100\% & 100\% & 91.3\% & 2,048 & 11,376 \\
SBM    & 100\% & 100\% & 100\% & 91.1\% & 2,082 & 19,460 \\
Tree   & 100\% & 100\% & 100\% & 96.9\% & 2,048 & 4,032 \\
QM9    & 100\% & 100\% & 100\% & 85.7\% & 2,285 & 4,468 \\
ZINC   & 100\% & 100\% & 100\% & 91.0\% & 2,149 & 4,618 \\
\bottomrule
\end{tabular}
\vskip -0.5em
\caption{ LGD autoencoder reconstruction at epoch 500, the checkpoint used for all reported LGD
results. All-pair accuracy includes the ``no edge'' class.}
\label{tab:lgd}
\end{table}

\subsection{Single-domain control: does the pipeline reproduce published results?}
\label{app:singledomain}

A multi-domain failure is only informative if the same pipeline can train the same architecture
successfully on one domain. We therefore retrained DeFoG and DiGress inside this paper's pipeline
on a single dataset at a time, and compared against each method's published numbers
(Table~\ref{tab:singledomain}). DeFoG lands within margin of its published V.U.N. and MMD ratios
on every domain, and DiGress exceeds its published accuracies. The pipeline is therefore not
suppressing these methods; the gap in Table~\ref{tab:benchmark} comes from the multi-domain setting
itself. The same table also places DiG$^{2}$ against single-domain specialists: trained on one
domain at a time it is competitive with both, and its multi-domain model matches or exceeds its own
single-domain variant on four of five domains, so the multi-domain result is not bought by
sacrificing per-domain quality.

\begin{table}[h!]
\centering
\small
\begin{tabular}{llccccc}
\toprule
Model & Setting & Planar & SBM & Tree & QM9 & ZINC \\
\midrule
\multirow{2}{*}{DeFoG}
 & single-domain (ours)   & 95.0 / 1.6 & 85 / 3.1 & 95 / 1.8 & 99.23 / 0.27 & 95.35 / 2.2 \\
 & published              & 99.5 / 1.6 & 90 / 4.9 & 96.5 / 1.6 & 98.9 / 0.26 & 96.65 / 2.1 \\
\midrule
\multirow{2}{*}{DiGress}
 & single-domain (ours)   & 85 / 1.2 & 80 / 2.3 & 95 / 2.0 & 99.4 / 0.3 & 94.16 / 2.5 \\
 & published              & 77.5 / 5.1 & 60.0 / 1.7 & 90.0 / 1.6 & 99.0 / -- & -- / -- \\
\midrule
\multirow{2}{*}{DiG$^{2}$}
 & single-domain          & 97.5 / 1.8 & 80 / 2.2 & 100 / 1.9 & 99.45 / 0.27 & 95.80 / 2.5 \\
 & multi-domain (Table~\ref{tab:benchmark}) & 100 / 2.3 & 85 / 1.8 & 97.5 / 1.56 & 99.92 / 0.29 & 95.46 / 2.7 \\
\bottomrule
\end{tabular}
\vskip -0.5em
\caption{Single-domain control. Accuracy or Validity / MMD Ratio or FCD. Published values are each
method's own reported results; cells its paper does not report are marked ``--''.}
\label{tab:singledomain}
\end{table}

\paragraph{Training cost.} Multi-domain capability does not cost extra compute per step. Timed to
batch saturation of 48GB of GPU memory on one domain at a time, DiG$^{2}$ runs at $47$/$51$/$43$/$44$/$43$
seconds per epoch on Planar/SBM/Tree/QM9/ZINC, against $45$/$52$/$45$/$42$/$44$ for DiGress and
$47$/$53$/$44$/$43$/$44$ for DeFoG. The per-arm costs under the multi-domain protocol are in
Table~\ref{tab:mem-measure}.

\subsection{Comparison with molecular foundation models}
\label{app:molfm}

GenMol~\citep{lee2025genmol} and DemoDiff~\citep{liu2026graph} are molecule-specific generators
rather than multi-domain graph models, so they cannot enter Table~\ref{tab:benchmark}. To compare
them on comparable terms, we run the zero-shot extrapolation protocol of \Secref{sec:extrap} on
GuacaMol's 2--27 node bands with $2{,}000$ generated samples per arm, and give each baseline an
embedding adapter fine-tuned on QM9 and ZINC for $250$K steps, providing explicit endpoint
awareness. Table~\ref{tab:molfm} reports the result. DiG$^{2}$'s V.U.N. is $21.8$ points above
GenMol and $39.0$ above DemoDiff, driven by uniqueness rather than raw validity, and its FCD is
$7.22$ and $3.59$ points lower. Because DiG$^{2}$ does not rely on rejection sampling, DiG$^{2}$
generates a valid molecule $1.9\times$ faster than GenMol and roughly $200\times$ faster than
DemoDiff.

\begin{table}[h!]
\centering
\small
\begin{tabular}{lccc}
\toprule
Metric & DiG$^{2}$ & GenMol & DemoDiff \\
\midrule
Validity              & 94.7\% & 95.4\% & 89.8\% \\
Uniqueness            & 100.0\% & 81.6\% & 63.9\% \\
Novelty               & 99.8\% & 94.2\% & 96.8\% \\
V.U.N.                & 94.6\% & 72.8\% & 55.6\% \\
FCD $\downarrow$      & 10.60 & 17.82 & 14.19 \\
Avg.\ s / valid sample $\downarrow$ & 0.40 & 0.76 & 80.1 \\
\bottomrule
\end{tabular}
\vskip -0.5em
\caption{DiG$^{2}$ against two molecular foundation models under one protocol, GuacaMol 2--27,
$2{,}000$ samples. Both baselines receive a QM9/ZINC embedding adapter fine-tuned for $250$K steps.}
\label{tab:molfm}
\end{table}

\subsection{DiG$^2$ Benchmarking Protocol: Motivation and Mechanisms}
\label{app:protocol}

Within the following subsection, we describe the mechanisms and motivations behind DiG$^2$'s benchmarking protocol. Figure~\ref{fig:protocol} compares the mechanism between the original setting (shown left) and the proposed DiG$^2$ benchmarking protocol (shown right). As shown in the left of Figure~\ref{fig:protocol}, the single domain setting forces each domain to be benchmarked alone with its own local node- and edge-features (circles are node classes, named by element on the two molecular cards, bars are edge classes, the hollow bar is no-edge), batches of one domain sized by graph count, its own model $\theta_{k}$, and one test on its samples. Independent dataset validity and MMD tests are indicated with check-marks. The right side of Figure~\ref{fig:protocol} indicates the uplift provided by the DiG$^2$ benchmarking protocol. At the top of Figure~\ref{fig:protocol}, where a global vocabulary for the benchmarking study consists of $11$ node and $5$ edge classes ($\ast$ is the generic node of the structural domains, followed by the atom element-types. The edge classes are: no-edge, generic single-edge, single, double and triple). Blank columns with one row per domain marking the domains respective mask $\sV_d$. Structural and molecular rows share only the no-edge class. We note that the H-atom is included within the node-vocabulary for compatibility with the QM9 dataset, no tests within this study included Hydrogen atoms. The natural benefit of the global vocabulary is that it can be extended to include further individual node types to suit practitioner needs. In the middle of Figure~\ref{fig:protocol}, we include an example of a training batch on the shared node budget with mixed-domains. Where one segment per graph is provided width equal to the respective graphs node count. The batching follows a strict node-ceiling limit to ensure consistent memory pressure, a mechanism which had not been previously-guaranteed with single-domain graph generation. In the bottom of Figure~\ref{fig:protocol}, a dispatcher asks for the domain label $d$ so that a benchmarked model $\theta$ is isolated to that domain's own validity tests, drawn as an icon: planarity (triangulation), block structure (SPECTRE validity), acyclicity (tree), or chemical validity (ring). This also enables further routing of MMD reference testing for each given domain. Altogether, DiG$^2$'s benchmarking protocol enables evaluation of multiple graph generation domains at-once, without requiring the loading of multiple models.

\begin{figure}[h!]
\centering
\input{figures/fig_protocol}
\caption{
\textbf{Single-domain benchmarking versus the multi-domain protocol.} \textcolor{digPlanar}{\rule{0.55em}{0.55em}}\,Planar, \textcolor{digSBM}{\rule{0.55em}{0.55em}}\,SBM, \textcolor{digTree}{\rule{0.55em}{0.55em}}\,Tree, \textcolor{digQM9}{\rule{0.55em}{0.55em}}\,QM9, \textcolor{digZINC}{\rule{0.55em}{0.55em}}\,ZINC.
\textbf{(1)}~The original setting, each domain remains isolated.
\textbf{(2)}~The DiG$^2$ protocol. \emph{Top:} the global vocabulary of $11$ node and $5$ edge classes. Blank columns reflect the vocabulary mask, $\sV_d$.
\emph{Middle:} a mixed-domain batch on the shared node budget.
\emph{Bottom:} A dispatcher routes $d$ to isolate model ($\theta$) predictions on a given domain's test, drawn as an icon: planarity (triangulation), block structure (SPECTRE validity), acyclicity (tree), or chemical validity (ring).}
\label{fig:protocol}
\end{figure}

\paragraph{Why five domains.} The five-domain protocol extends the setting that state-of-the-art
single-domain models are evaluated in, that way comparison in
Table~\ref{tab:benchmark} is measured against the same reference graphs as preceding literature. It also targets structural distinctiveness rather than structural coverage,
successful generation requires Planar samples to retain a Delaunay triangulation, SBM samples to
retain community structure, and QM9 samples to retain valid atom and bond types.

\label{app:benchmarking_protocol}

\begin{figure}[h]
    \centering
    \includegraphics[width=\linewidth]{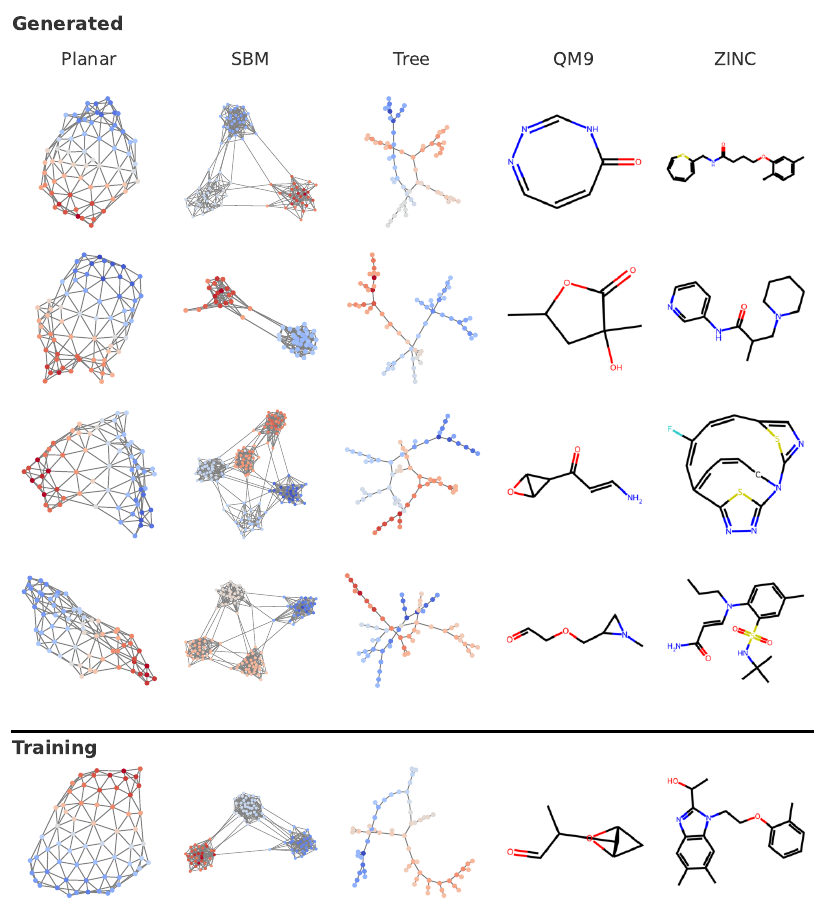}
    \caption{
    \textbf{DiG$^2$ generations across all five domains, over real training graphs.} One column per
    domain. The top four rows are generations from a single DiG$^{2}$ checkpoint, the bottom row
    (below the rule) is a reference graph from the corresponding domain.
    }
    \label{fig:domain_grid_h}
\end{figure}

\paragraph{Why not LGGM's pre-training protocol directly.} LGGM extracts 2- and 3-hop subgraphs
from its domains. At 3 hops an SBM sample is capped at 71 of its 188 nodes, which severs the
inter-community edges that the SBM validity test measures, so the resulting samples are not
comparable to the SBM reference graphs used within single-domain literature. LGGM's
protocol is also adjacency-only, which carries no node features, so atom types cannot be modeled without the node- and edge-vocabularies introduced within this study's benchmarking protocol. DiG$^{2}$'s vocabulary-aware protocol
can therefore be seen as an extension of LGGM's structure-only setting rather than a different choice of
datasets.

%% file: figures/fig_protocol.tex
{\rmfamily\fontsize{6.5}{7.8}\selectfont%
\def\svgwidth{\linewidth}%
\IfFileExists{figures/fig_protocol.pdf_tex}%
  {\input{figures/fig_protocol.pdf_tex}}%
  {\input{fig_protocol.pdf_tex}}%
}%

%% file: figures/fig_protocol.pdf_tex
\begingroup%
  \makeatletter%
  \providecommand\color[2][]{%
    \errmessage{(Inkscape) Color is used for the text in Inkscape, but the package 'color.sty' is not loaded}%
    \renewcommand\color[2][]{}%
  }%
  \providecommand\transparent[1]{%
    \errmessage{(Inkscape) Transparency is used (non-zero) for the text in Inkscape, but the package 'transparent.sty' is not loaded}%
    \renewcommand\transparent[1]{}%
  }%
  \providecommand\rotatebox[2]{#2}%
  \newcommand*\fsize{\dimexpr\f@size pt\relax}%
  \newcommand*\lineheight[1]{\fontsize{\fsize}{#1\fsize}\selectfont}%
  \ifx\svgwidth\undefined%
    \setlength{\unitlength}{396bp}%
    \ifx\svgscale\undefined%
      \relax%
    \else%
      \setlength{\unitlength}{\unitlength * \real{\svgscale}}%
    \fi%
  \else%
    \setlength{\unitlength}{\svgwidth}%
  \fi%
  \global\let\svgwidth\undefined%
  \global\let\svgscale\undefined%
  \makeatother%
  \begin{picture}(1,0.68686869)%
    \lineheight{1}%
    \setlength\tabcolsep{0pt}%
    \put(0,0){\includegraphics[width=\unitlength,page=1]{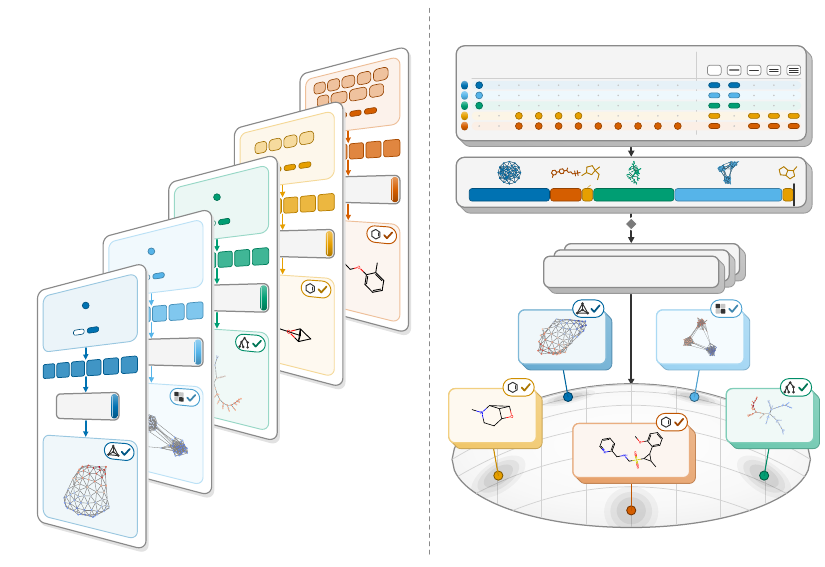}}%
    \put(0.38740309,0.57647445){\color[rgb]{0.2000,0.2000,0.2000}\scalebox{0.7512}[0.8361]{\makebox(0,0)[t]{\lineheight{1.25}\smash{\begin{tabular}[t]{c}{\fontsize{5.2}{6.2}\selectfont C}\end{tabular}}}}}%
    \put(0.40421505,0.58012544){\color[rgb]{0.2000,0.2000,0.2000}\scalebox{0.7979}[0.8617]{\makebox(0,0)[t]{\lineheight{1.25}\smash{\begin{tabular}[t]{c}{\fontsize{5.2}{6.2}\selectfont N}\end{tabular}}}}}%
    \put(0.42208754,0.58400673){\color[rgb]{0.2000,0.2000,0.2000}\scalebox{0.8490}[0.8889]{\makebox(0,0)[t]{\lineheight{1.25}\smash{\begin{tabular}[t]{c}{\fontsize{5.2}{6.2}\selectfont O}\end{tabular}}}}}%
    \put(0.44112420,0.58814085){\color[rgb]{0.2000,0.2000,0.2000}\scalebox{0.9052}[0.9178]{\makebox(0,0)[t]{\lineheight{1.25}\smash{\begin{tabular}[t]{c}{\fontsize{5.2}{6.2}\selectfont F}\end{tabular}}}}}%
    \put(0.46144258,0.59255331){\color[rgb]{0.2000,0.2000,0.2000}\scalebox{0.9672}[0.9487]{\makebox(0,0)[t]{\lineheight{1.25}\smash{\begin{tabular}[t]{c}{\fontsize{5.2}{6.2}\selectfont P}\end{tabular}}}}}%
    \put(0.39151186,0.56077418){\color[rgb]{0.2000,0.2000,0.2000}\scalebox{0.7581}[0.8424]{\makebox(0,0)[t]{\lineheight{1.25}\smash{\begin{tabular}[t]{c}{\fontsize{5.2}{6.2}\selectfont S}\end{tabular}}}}}%
    \put(0.41078767,0.56438284){\color[rgb]{0.2000,0.2000,0.2000}\scalebox{0.8118}[0.8717]{\makebox(0,0)[t]{\lineheight{1.25}\smash{\begin{tabular}[t]{c}{\fontsize{5.2}{6.2}\selectfont Cl}\end{tabular}}}}}%
    \put(0.43384189,0.56869887){\color[rgb]{0.2000,0.2000,0.2000}\scalebox{0.8784}[0.9068]{\makebox(0,0)[t]{\lineheight{1.25}\smash{\begin{tabular}[t]{c}{\fontsize{5.2}{6.2}\selectfont Br}\end{tabular}}}}}%
    \put(0.45623580,0.57289128){\color[rgb]{0.2000,0.2000,0.2000}\scalebox{0.9456}[0.9408]{\makebox(0,0)[t]{\lineheight{1.25}\smash{\begin{tabular}[t]{c}{\fontsize{5.2}{6.2}\selectfont I}\end{tabular}}}}}%
    \put(0.46966408,0.44930311){\color[rgb]{0.2000,0.2000,0.2000}\scalebox{0.9703}[0.9612]{\makebox(0,0)[rt]{\lineheight{1.25}\smash{\begin{tabular}[t]{r}{\fontsize{9}{10.5}\selectfont $\theta_{5}$}\end{tabular}}}}}%
    \put(0.31613704,0.50425715){\color[rgb]{0.2000,0.2000,0.2000}\scalebox{0.7717}[0.8487]{\makebox(0,0)[t]{\lineheight{1.25}\smash{\begin{tabular}[t]{c}{\fontsize{5.2}{6.2}\selectfont C}\end{tabular}}}}}%
    \put(0.33346710,0.50776110){\color[rgb]{0.2000,0.2000,0.2000}\scalebox{0.8204}[0.8751]{\makebox(0,0)[t]{\lineheight{1.25}\smash{\begin{tabular}[t]{c}{\fontsize{5.2}{6.2}\selectfont N}\end{tabular}}}}}%
    \put(0.35190790,0.51148962){\color[rgb]{0.2000,0.2000,0.2000}\scalebox{0.8738}[0.9031]{\makebox(0,0)[t]{\lineheight{1.25}\smash{\begin{tabular}[t]{c}{\fontsize{5.2}{6.2}\selectfont O}\end{tabular}}}}}%
    \put(0.37156976,0.51546503){\color[rgb]{0.2000,0.2000,0.2000}\scalebox{0.9326}[0.9330]{\makebox(0,0)[t]{\lineheight{1.25}\smash{\begin{tabular}[t]{c}{\fontsize{5.2}{6.2}\selectfont F}\end{tabular}}}}}%
    \put(0.39011863,0.38364654){\color[rgb]{0.2000,0.2000,0.2000}\scalebox{0.9703}[0.9612]{\makebox(0,0)[rt]{\lineheight{1.25}\smash{\begin{tabular}[t]{r}{\fontsize{9}{10.5}\selectfont $\theta_{4}$}\end{tabular}}}}}%
    \put(0.31057317,0.31798998){\color[rgb]{0.2000,0.2000,0.2000}\scalebox{0.9703}[0.9612]{\makebox(0,0)[rt]{\lineheight{1.25}\smash{\begin{tabular}[t]{r}{\fontsize{9}{10.5}\selectfont $\theta_{3}$}\end{tabular}}}}}%
    \put(0.23102772,0.25233341){\color[rgb]{0.2000,0.2000,0.2000}\scalebox{0.9703}[0.9612]{\makebox(0,0)[rt]{\lineheight{1.25}\smash{\begin{tabular}[t]{r}{\fontsize{9}{10.5}\selectfont $\theta_{2}$}\end{tabular}}}}}%
    \put(0.09974550,0.18731263){\color[rgb]{0.2000,0.2000,0.2000}\scalebox{0.8180}[0.8826]{\makebox(0,0)[t]{\lineheight{1.25}\smash{\begin{tabular}[t]{c}{\fontsize{9}{10.5}\selectfont $\theta_{1}$}\end{tabular}}}}}%
    \put(0.03535354,0.64898990){\color[rgb]{0.2000,0.2000,0.2000}\makebox(0,0)[lt]{\lineheight{1.25}\smash{\begin{tabular}[t]{l}\fontsize{11}{13}\selectfont\textbf{(1)}\end{tabular}}}}%
    \put(0.58080808,0.59873737){\color[rgb]{0.2000,0.2000,0.2000}\makebox(0,0)[t]{\lineheight{1.25}\smash{\begin{tabular}[t]{c}{\fontsize{5.6}{6.6}\selectfont $\ast$}\end{tabular}}}}%
    \put(0.60488215,0.59696970){\color[rgb]{0.2000,0.2000,0.2000}\makebox(0,0)[t]{\lineheight{1.25}\smash{\begin{tabular}[t]{c}{\fontsize{5.6}{6.6}\selectfont H}\end{tabular}}}}%
    \put(0.62895623,0.59696970){\color[rgb]{0.2000,0.2000,0.2000}\makebox(0,0)[t]{\lineheight{1.25}\smash{\begin{tabular}[t]{c}{\fontsize{5.6}{6.6}\selectfont C}\end{tabular}}}}%
    \put(0.65303030,0.59696970){\color[rgb]{0.2000,0.2000,0.2000}\makebox(0,0)[t]{\lineheight{1.25}\smash{\begin{tabular}[t]{c}{\fontsize{5.6}{6.6}\selectfont N}\end{tabular}}}}%
    \put(0.67710438,0.59696970){\color[rgb]{0.2000,0.2000,0.2000}\makebox(0,0)[t]{\lineheight{1.25}\smash{\begin{tabular}[t]{c}{\fontsize{5.6}{6.6}\selectfont O}\end{tabular}}}}%
    \put(0.70117845,0.59696970){\color[rgb]{0.2000,0.2000,0.2000}\makebox(0,0)[t]{\lineheight{1.25}\smash{\begin{tabular}[t]{c}{\fontsize{5.6}{6.6}\selectfont F}\end{tabular}}}}%
    \put(0.72525253,0.59696970){\color[rgb]{0.2000,0.2000,0.2000}\makebox(0,0)[t]{\lineheight{1.25}\smash{\begin{tabular}[t]{c}{\fontsize{5.6}{6.6}\selectfont P}\end{tabular}}}}%
    \put(0.74932660,0.59696970){\color[rgb]{0.2000,0.2000,0.2000}\makebox(0,0)[t]{\lineheight{1.25}\smash{\begin{tabular}[t]{c}{\fontsize{5.6}{6.6}\selectfont S}\end{tabular}}}}%
    \put(0.77340067,0.59696970){\color[rgb]{0.2000,0.2000,0.2000}\makebox(0,0)[t]{\lineheight{1.25}\smash{\begin{tabular}[t]{c}{\fontsize{5.6}{6.6}\selectfont Cl}\end{tabular}}}}%
    \put(0.79747475,0.59696970){\color[rgb]{0.2000,0.2000,0.2000}\makebox(0,0)[t]{\lineheight{1.25}\smash{\begin{tabular}[t]{c}{\fontsize{5.6}{6.6}\selectfont Br}\end{tabular}}}}%
    \put(0.82154882,0.59696970){\color[rgb]{0.2000,0.2000,0.2000}\makebox(0,0)[t]{\lineheight{1.25}\smash{\begin{tabular}[t]{c}{\fontsize{5.6}{6.6}\selectfont I}\end{tabular}}}}%
    \put(0.57676768,0.61540404){\color[rgb]{0.4784,0.4784,0.4784}\makebox(0,0)[lt]{\lineheight{1.25}\smash{\begin{tabular}[t]{l}{\fontsize{5.6}{6.6}\selectfont node vocabulary}\end{tabular}}}}%
    \put(0.85749158,0.61540404){\color[rgb]{0.4784,0.4784,0.4784}\makebox(0,0)[lt]{\lineheight{1.25}\smash{\begin{tabular}[t]{l}{\fontsize{5.6}{6.6}\selectfont edge vocabulary}\end{tabular}}}}%
    \put(0.77878788,0.40782828){\color[rgb]{0.2000,0.2000,0.2000}\makebox(0,0)[lt]{\lineheight{1.25}\smash{\begin{tabular}[t]{l}{\fontsize{9}{10.5}\selectfont $d$}\end{tabular}}}}%
    \put(0.76515152,0.34949495){\color[rgb]{0.2000,0.2000,0.2000}\makebox(0,0)[t]{\lineheight{1.25}\smash{\begin{tabular}[t]{c}{\fontsize{9.5}{11}\selectfont $\theta$}\end{tabular}}}}%
    \put(0.54545455,0.64898990){\color[rgb]{0.2000,0.2000,0.2000}\makebox(0,0)[lt]{\lineheight{1.25}\smash{\begin{tabular}[t]{l}\fontsize{11}{13}\selectfont\textbf{(2)}\end{tabular}}}}%
  \end{picture}%
\endgroup%

%% file: sections/appendix/revision_scaling.tex
{
\subsection{Node versus edge vocabulary masking}
\label{app:maskablation}

The ablation of \Secref{sec:ablation} removes vocabulary masking as a single component.
Table~\ref{tab:maskablation} separates it into its node and edge halves. Removing node masking
costs little on Planar and Tree, given that masking the node channel removes almost no probability mass. Edge masking
is the load-bearing half, removing it costs $35.0$ points on Planar and $12.5$ on SBM, both well
outside the roughly $\pm8.0$ noise band of this evaluation. This is reflective on the asymmetry for the structural domains,
since what separates a planar graph from a molecule in the edge channel is often which bond
classes are admissible given current sampled structure.

\begin{table}[h!]
\centering
\small
\begin{tabular}{lccc}
\toprule
Condition & Planar & SBM & Tree \\
\midrule
Both (Table~\ref{tab:benchmark}) & 100\% & 85.0\% & 97.5\% \\
w/o node masking & 95.0\% & 77.5\% & 95.0\% \\
w/o edge masking & 65.0\% & 72.5\% & 90.0\% \\
Neither & 72.5\% & 70.0\% & 92.5\% \\
\bottomrule
\end{tabular}
\vskip -0.5em
\caption{Structural V.U.N. when vocabulary masking is removed from the node channel, the edge
channel, or both. Edge masking carries the effect.}
\label{tab:maskablation}
\end{table}

\subsection{Scaling the number of training domains}
\label{app:scaling}

Table~\ref{tab:scaling} trains DiG$^{2}$ under six different domain mixtures, holding the
architecture and schedule fixed. Three observations follow. First, performance does not degrade as
domains are added, going from four to five to six domains leaves every shared cell within a few
points, and SBM improves monotonically across that range ($85 \rightarrow 85 \rightarrow 87.5$).
Second, adding MOSES as a sixth domain costs nothing on the structural domains, which share no
vocabulary with it. Third, the family-restricted arms show that DiG$^{2}$ does not depend on having
both families present. So, the molecular-regime and structural-regime models land within a few points of
the five-domain model on their own domains.

\begin{table}[h!]
\centering
\small
\begin{tabular}{lcccccc}
\toprule
Training mixture & Planar & SBM & Tree & QM9 & ZINC & MOSES \\
\midrule
4 domains                  & 97.5 / 2.1 & 85 / 2.0 & 95 / 1.5 & 99.29 / 0.28 & -- & -- \\
5 domains (Table~\ref{tab:benchmark}) & 100 / 2.3 & 85 / 1.8 & 97.5 / 1.56 & 99.92 / 0.29 & 95.46 / 2.7 & -- \\
6 domains                  & 95.0 / 2.5 & 87.5 / 1.7 & 95.0 / 1.4 & 99.46 / 0.31 & 94.38 / 2.4 & 93.0 / 2.1 \\
Molecular only             & -- & -- & -- & 99.38 / 0.22 & 95.26 / 2.4 & 94.4 / 1.9 \\
Structural only            & 97.5 / 1.9 & 82.5 / 2.1 & 100 / 1.5 & -- & -- & -- \\
15 domains (Table~\ref{tab:lggm15}) & 92.5 / 1.20 & 85 / 2.30 & 97.5 / 1.92 & -- & -- & -- \\
\bottomrule
\end{tabular}
\vskip -0.5em
\caption{DiG$^{2}$ under different training mixtures. Validity / MMD Ratio or FCD. A
``--'' marks a domain absent from that mixture.}
\label{tab:scaling}
\end{table}

%% file: sections/appendix/revision_repr.tex
{
\subsection{What the internal representations separate}
\label{app:repr}

This section looks inside two representations, the structural encoder which produces DiG$^{2}$'s
conditioning signal, and the denoising transformer's final layer. Throughout, the same $40$
held-out graphs per structural domain are used, and each model is given its own conditioning and
its own \emph{clean} end of the time axis. That last point is easy to get wrong, DiG$^{2}$ runs
flow matching, where $t=1$ is the clean graph (Definition~\ref{def:forward}), while LGGM is
DDPM-style, where $\bar\alpha$ decreases in $t$ so the clean end is $t \to 0$. Feeding both
$t=1$ asks LGGM to encode pure noise, which collapses its representation on its own.

\paragraph{The conditioning encoder: DiG$^{2}$ clusters stop touching.} As shown in the left-half of Figure~\ref{fig:domes}, before wider training, Planar
and SBM are all but inseparable in the encoder. This is measured directly where their closest members lie $0.070$ apart against
a centroid distance of $0.871$, a relative gap of $0.08$. Afterwards, the gap between Planar and SBM is $0.284$, or
$0.33$ relative to one another. This equates to a $4\times$ wider gap in the full $32$-d space and roughly $10.7\times$ increase in the projection ($0.03\rightarrow 0.32$ in the 2-D projection of Figure~\ref{fig:domes}. The centroid distance is measurably unchanged ($\times 1.00$). However, the shape of the embedding is altered, where SBM's principal-axis ratio falls from $89{:}1$ to $17{:}1$, as shown by the long axis within Figure~\ref{fig:domes} no longer
pointing at Planar.

\paragraph{The denoiser's final layer: LGGM is label-aware but structure-blind.} Given that LGGM has no structural encoder, we can only make comparisons across sampled domains at it's final denoising layer. This is notable given that Table~\ref{tab:repr} shows that the application of LGGM labels separates the three domains for a Silhouette score of $0.906$ relative to
DiG$^{2}$ which encodes structure for a Silhouette score of $0.731$. Despite this, LGGM's representation collapses every graph onto label clusters so we observe that LGGM reaches $4.2\%$ structural accuracy
against DiG$^{2}$'s $94.2\%$.

\begin{table}[h!]
\centering
\small
\begin{tabular}{lccc}
\toprule
 & DiG$^{2}$ (5 domains) & DiG$^{2}$ (15 domains) & LGGM \\
\midrule
\multicolumn{4}{l}{\emph{Structural Encoder (LGGM relies on Text)}} \\
Planar--SBM Closest-Pair Gap & 0.070 & \textbf{0.284} & -- \\
Planar--SBM Centroid Distance & 0.871 & 0.868 & -- \\
SBM Elongation & 89.5 & \textbf{16.6} & -- \\
\midrule

\midrule
\multicolumn{4}{l}{\emph{Denoiser Final Layer}} \\
Silhouette (Planar, SBM, Tree)    & 0.731 & 0.710 & \textbf{0.906} \\
\midrule
Mean Structural V.U.N. & 94.2\% & 91.7\% & 4.2\% \\
\bottomrule
\end{tabular}
\vskip -0.5em
\caption{Separation in both internal spaces, on the same held-out graphs. In the
encoder, wider training opens the Planar/SBM gap $4\times$ without moving the centroids. In the
denoiser, silhouette score ranks LGGM first but mean structural accuracy indicates this does not contribute to improved downstream performance. Bold marks the better value in
each row.}
\label{tab:repr}
\end{table}
}

%% file: sections/appendix/revision_protocol.tex
{
\subsection{Benchmark protocol: data, splits, preprocessing, baselines and evaluation}
\label{app:protocol-doc}

The contribution is an evaluation protocol over existing public datasets, not a new dataset: every graph comes from the sources in Table~\ref{tab:splits}, and the released code rebuilds the splits deterministically.

Within the following section, we detail how the five datasets within the DiG$^2$ benchmarking protocol, with: one shared vocabulary with a per-domain mask
$\sV_d$, mixed-domain batches under a node budget, baselines retrained under that setting, and one
scoring pipeline with a structural test per domain (Figure~\ref{fig:protocol}).

\paragraph{Sources and splits.} Table~\ref{tab:splits} lists each domain's source, split rule and
size. \textsc{Planar} and \textsc{SBM} are the datasets of \citet{martinkus2022spectre} and
\textsc{Tree} that of \citet{bergmeister2024efficient}, \textsc{QM9}~\citep{ramakrishnan2014qm9} and
\textsc{ZINC}250k~\citep{sterling2015zinc} are the molecular datasets. The structural domains use the split stored in the
upstream files. \textsc{QM9} uses the split of the DiGress codebase, the $133{,}885$ molecules are
shuffled with seed $42$ and divided into $100{,}000$ training molecules, $10\%$ test and the
remainder validation; after RDKit sanitization and deduplication the split sizes are those of Table~\ref{tab:splits}. \textsc{ZINC}
uses the held-out index file of the GruM release~\citep{grum2024pmlr}. That release provides a single held-out set of
$24{,}887$ molecules, which serves as both validation and test set, for every arm. The sixth-domain experiment
(Table~\ref{tab:scaling}) adds \textsc{MOSES}~\citep{polykovskiy2020moses}, which is kekulized and hydrogen-free, using seven atom types that are already in the vocabulary. The
$15$-domain experiment (Table~\ref{tab:lggm15}) adds twelve domains from the 2-hop/3-hop extracted subgraphs from LGGM's public
release~\citep{wang2025large} with that release's splits, as symmetric adjacency matrices without
self-loops, encoded with the generic node and edge classes. LGGM obtained their graphs from the Network Data Repository~\citep{nr16kdd}. The $\alpha$-interpolation experiemnts uses both GuacaMol~\citep{brown2019guacamol} and MOSES within subsampled noise bands, as described within Appendix~\ref{app:zero_shot} The released
code downloads each source and rebuilds these splits deterministically, so no split files are
distributed separately.

\begin{table}[h!]
\centering
\small
\begin{tabular}{lllrrr}
\toprule
Domain & Source & Source Split Rule & Train & Val. & Test \\
\midrule
Planar & \citet{martinkus2022spectre} & Random & 128 & 32 & 40 \\
SBM    & \citet{martinkus2022spectre} & Random & 128 & 32 & 40 \\
Tree   & \citet{bergmeister2024efficient} & Random & 128 & 32 & 40 \\
QM9    & \citet{ramakrishnan2014qm9} & Random & 97{,}734 & 20{,}042 & 13{,}055 \\
ZINC250k & \citet{sterling2015zinc} & Random & 224{,}568 & 24{,}887 &  24{,}887\\
MOSES & \citet{polykovskiy2020molecular} & Scaffold & 1{,}584{,}663 & 176{,}225 &  176{,}074\\
GuacaMol & \citet{brown2019guacamol} & Random & 1{,}273{,}104 & 79{,}568 &  238{,}706\\
FB & \citet{nr16kdd} & Random & 403 & 50 & 51\\
ASN & \citet{nr16kdd} & Random & 213 &	27 &27 \\
Email & \citet{nr16kdd} & Random & 403	& 50 & 51 \\
Web & \citet{nr16kdd} & Random & 403 & 50 & 51 \\
Road & \citet{nr16kdd} & Random & 403 & 50 & 51 \\
Power & \citet{nr16kdd} & Random & 409 & 51 & 52 \\
Bio & \citet{nr16kdd} & Random & 403 & 50 & 51 \\
Econ & \citet{nr16kdd} & Random & 403 & 50 & 51 \\
RT & \citet{nr16kdd} & Random & 446 & 56 & 56 \\
Col & \citet{nr16kdd} & Random & 403 & 50 & 51 \\
Eco & \citet{nr16kdd} & Random & 200 & 50 & 1 \\
Cit. & \citet{nr16kdd} & Random & 403 & 50 & 51 \\

\bottomrule
\end{tabular}
\vskip -0.5em
\caption{Source, split rule and number of graphs per split for the five benchmark domains, MOSES and GuacaMol (Sections~\ref{sec:interp}--\ref{sec:extrap}), and the 12 LGGM domains (Table~\ref{tab:lggm15}).}
\label{tab:splits}
\end{table}

\paragraph{Preprocessing.} Every graph is encoded once, when the dataset is built, into the
shared vocabulary. The $11$ node classes are a generic structural node and the atoms H, C, N, O,
F, P, S, Cl, Br and I. The $5$ edge classes are no edge, a generic structural edge, and single,
double and triple bonds. Structural domains (i.e. Planar, SBM, Tree) use the generic node and generic edge classes only;
\textsc{QM9} uses C, N, O, F and \textsc{ZINC} uses C, N, O, F, P, S, Cl, Br, I, both with the
three bond orders. Molecules are kekulized, so there is no aromatic bond class. The protocol is
hydrogen-suppressed: \textsc{QM9} is reduced to heavy atoms and \textsc{ZINC} molecules are parsed
without hydrogens, which is the released configuration (Table~\ref{tab:gdfm-hparams}). Batching and node budgets
are described in App.~\ref{app:dataload}, and node-count ranges in Table~\ref{tab:data-measure}.

\paragraph{Baselines.} Each baseline is run from its authors' public implementation, wrapped to
read the shared data and vocabulary: G2PT~\citep{g2pt2025}, DeFoG~\citep{qin2024defog},
HOG-Diff~\citep{huang2026hog}, LGGM~\citep{wang2025large}, Pard~\citep{zhao2024pard},
GraphRNN~\citep{you2018graphrnn}, LGD~\citep{zhou2024unifying}
and SPECTRE~\citep{martinkus2022spectre}. DiGress$^{\dagger}$ uses the
DiGress implementation~\citep{vignac2023digress} distributed with LGGM, with one-hot domain
labels. GraphBSI$^{\dagger}$~\citep{petersen2026discrete} is our reimplementation from the paper. The shared training settings, architectures and
method-specific hyperparameters are in App.~\ref{app:baselines}
(Tables~\ref{tab:baseline-arch} and~\ref{tab:baseline-gen}), the epochs each arm completed in
Table~\ref{tab:epochs}, and the conditioning each arm received in \Secref{sec:benchmark}.

\paragraph{Evaluation.} Each model generates $40$ graphs per structural domain and $10{,}000$
molecules per molecular domain (Table~\ref{tab:data-measure}), one domain at a time, with node
counts drawn from that domain's training-size histogram. Each domain samples from its own seeded
random stream, so its samples do not depend on the other domains' budgets.
\begin{itemize}
\itemsep0em
\item \emph{Structural accuracy/validity} applies the domain's test to each generated graph (planarity,
block structure or acyclicity; \Secref{sec:benchmark}).
\item \emph{MMD ratio.} Degree, clustering, orbit, spectral and wavelet MMDs are computed between
the generated graphs and the domain's test split. Each is divided by the same MMD measured
between the training and test splits, and the five ratios are averaged. Orbit counts use ORCA.
\item \emph{Uniqueness and novelty} are computed by isomorphism against the training split, and
in Table~\ref{tab:novelty} also against the fine-tuning pools. The DiG$^2$ benchmarking protocol applies this for all structural and molecular domains.
\item \emph{Molecular validity} follows RDKit sanitization; \emph{FCD} is computed against the
test-split SMILES (for \textsc{ZINC}, the held-out set); novelty is measured against the training
SMILES.
\end{itemize}

All resources for the multi-domain benchmarking protocol, $\alpha$-interpolation, and the DiG$^2$ model code are included within the \href{\coderepo}{code repository}.

%% file: sections/appendix/revision_interp.tex
{
\subsection{Are individual molecules interpolated, or is each set a mixture?}
\label{app:interp_test}

A sweep over $\alpha$ could produce intermediate \emph{aggregate} statistics in two very different
ways. Either each generated molecule is itself intermediate between QM9 and ZINC, or each sampled
set is a mixture of QM9-like and ZINC-like molecules in proportion to $\alpha$, with nothing in
between. 

To separate these points directly, each generated molecule is projected onto the Fisher direction
between the QM9 and ZINC endpoint populations, giving a single axis score per molecule, scaled so
that the QM9 endpoint sits at $0$ and the ZINC endpoint at $1$. If the samples are blended together into a mixture~\citep{reynolds2009gaussian}, the
per-set spread should \emph{widen} at intermediate $\alpha$, since the set would contain two
separated clusters to signify two separation sample populations. If the two samples are truly interpolating, then DiG$^2$'s distribution spread should move with the provided endpoint values. Table~\ref{tab:interp} reports both the observed spread with the spread of a two-component
Gaussian mixture of the $\alpha=0$ and $\alpha=1$ endpoint populations with weight $\alpha$, $\sigma_{\mathrm{mix}=\sqrt{(1-\alpha)s_0^2+\alpha s_1^2+\alpha(1-\alpha)}}$ (endpoint means are $0$ and $1$ on this axis), computed from the endpoint rows alone}. We also compare the two
explanations by Bayesian information criterion ($\Delta\mathrm{BIC} > 0$ favors the single
interpolated population).

The mean axis score tracks $\alpha$ monotonically, and at every intermediate $\alpha$ the observed
spread is tighter than the corresponding mixture, by $0.27$ at $\alpha=0.5$. $\Delta\mathrm{BIC}$
is positive at every intermediate $\alpha$. We therefore observed the conditioning blend moving individual
molecules along the axis rather than re-weighting two endpoint populations.

\begin{table}[h!]
\centering
\small
\begin{tabular}{lccccl}
\toprule
$\alpha$ & Molecules & Mean axis score & Observed spread (sd) & Mixture spread (sd) & $\Delta\mathrm{BIC}$ \\
\midrule
0 (QM9)  & 2,583 & 0.00 & 0.40 & -- & reference \\
0.50 & 5,596 & 0.58 & 0.34 & 0.62 & $+4044$ \\
0.75 & 5,655 & 0.81 & 0.32 & 0.55 & $+1557$ \\
0.85 & 5,654 & 0.89 & 0.32 & 0.49 & $+559$ \\
0.95 & 5,640 & 0.97 & 0.31 & 0.39 & $+40$ \\
1 (ZINC) & 2,829 & 1.00 & 0.31 & -- & reference \\
\bottomrule
\end{tabular}
\vskip -0.5em
\caption{Interpolation versus mixture along the QM9$\rightarrow$ZINC Fisher direction. The observed
spread stays within the range of the endpoint spreads ($0.31$--$0.40$) instead of widening, and $\Delta\mathrm{BIC}$ favors
a single interpolated population at every intermediate $\alpha$. The mixture null is given its best case---its weight is fitted on a $201$-point grid rather than fixed at $\alpha$, and it carries one free parameter against the single population's two, so it receives the lighter complexity penalty---and it still loses in every interior cell.}
\label{tab:interp}
\end{table}

\begin{figure}[h]
    \centering
    \includegraphics[width=\linewidth]{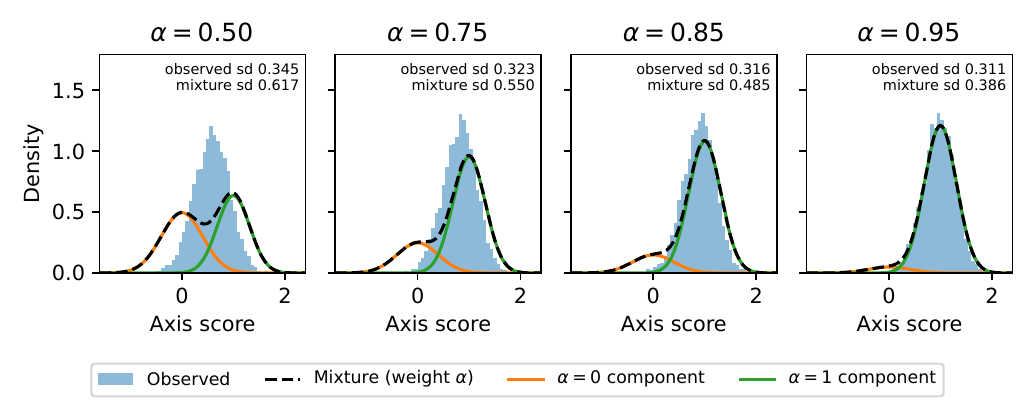}
    \caption{
    \textbf{The interpolated distributions for DiG$^2$ on the QM9 to ZINC $\alpha$-endpoints, as shown in Table~\ref{tab:interp}.}
    }
    \label{fig:bic_interp}
\end{figure}

We also visualize DiG$^2$'s interpolation between the QM9 and ZINC $\alpha$-interpolation endpoints directly within Figure~\ref{fig:bic_interp}. At $\alpha=0.95$, the observed distribution from DiG$^2$ (blue histogram) follows closely with the replicated ZINC distribution (green). Given that DiG$^2$ relies on the $\alpha$-interpolation mechanism, the observed spread of it's standard deviation is $0.311$ relative to the QM9-ZINC components spread of $0.386$. We observe the spread for the mixed QM9-ZINC components grows drastically with $\alpha=0.5$, reaching a standard-deviation spread of $0.617$ and indicating bimodal structure, whereas the DiG$^2$ distribution (observed) retains a standard deviation of $0.345$ and maintains a singular mode. Therefore, indicating that DiG$^2$ interpolates between given $\alpha$-endpoints instead of simple generating a mixture of QM9 and ZINC graphs.

\subsection{Extrapolation measured against ZINC, not MOSES}
\label{app:extrap_test}

Given that MOSES filters out molecules above roughly $\log P = 3.5$, a sample could sit outside MOSES while remaining ordinary for ZINC, the
dataset DiG$^{2}$ actually trained on. We therefore conduct additional testing re-measureing the $\alpha > 1$
sweep against ZINC's own $\log P$ distribution, as shown within Table~\ref{tab:extrap}.

The median logP rises monotonically past the $\alpha = 1$ ZINC endpoint, from $2.89$ to $4.82$,
against a ZINC reference median of $2.92$. At the last cell above our validity floor, median logP rises monotonically from $2.89$ at $\alpha{=}1.00$ to $4.21$. The fraction of samples above ZINC's own $95$th
percentile grows from $5.4\%$ to $59.2\%$, and the Mahalanobis distance from the ZINC centroid
grows by $2.77$ over the $\alpha = 1$ distance. Novelty remains $100\%$ at every $\alpha$,
including $\alpha > 1$, so the extrapolated molecules are not drawn from either training set. Therefore indicating that $\alpha > 1$ moves generation beyond ZINC's own $\log P$
distribution (\Secref{sec:extrap}).

\begin{table}[h!]
\centering
\small
\begin{tabular}{lccccc}
\toprule
Set & $n$ & $\log P$ median & \% above ZINC p95 & Novelty & Mahalanobis vs ZINC \\
\midrule
ZINC reference & 38,354 & 2.92 & -- & -- & -- \\
$\alpha = 1.00$ & 5,580 & 2.89 & 5.4\% & 100\% & 0.90 \\
$\alpha = 1.25$ & 5,451 & 3.25 & 11.2\% & 100\% & 0.81 \\
$\alpha = 1.50$ & 5,337 & 3.66 & 22.8\% & 100\% & 1.34 \\
$\alpha = 1.75$ & 4,875 & 4.21 & 41.2\% & 100\% & 2.30 \\
$\alpha = 2.00$ & 4,006 & 4.82 & 59.2\% & 100\% & 3.67 \\
\bottomrule
\end{tabular}
\vskip -0.5em
\caption{$\alpha > 1$ extrapolation measured against the ZINC training distribution rather than the
MOSES band.}
\label{tab:extrap}
\end{table}

\subsection{LGGM with and without per-domain priors}
\label{app:lggm_w_wo_priors}

\begin{table}[h!]
\centering
\small
\begin{tabular}{lccccc}
\toprule
 & \multicolumn{3}{c}{Structural (V.U.N.\ $\uparrow$ / Ratio $\downarrow$)} & \multicolumn{2}{c}{Molecular (Valid.\ $\uparrow$ / FCD $\downarrow$)} \\
\cmidrule(lr){2-4}\cmidrule(lr){5-6}
Model & Planar & SBM & Tree & QM9 & ZINC \\
\midrule
LGGM (Table~\ref{tab:benchmark}) & 0.0 / 76 & 12.5 / 9.1 & 0.0 / 8.8 & 86.3 / 0.75 & 85.2 / 8.55 \\
LGGM$^{\dagger}$ & 0.0 / 131 & 35.0 / 5.38 & 65.0 / 1.3 & 91.8 / 0.78 & 77.53 / 7.71 \\
\bottomrule
\end{tabular}
\vskip -0.5em
\caption{LGGM with per-domain priors (LGGM$^{\dagger}$) against the LGGM row of
Table~\ref{tab:benchmark}. Per-domain priors raise \textsc{SBM} and \textsc{Tree} V.U.N.\ by $22.5$
and $65$ points and QM9 validity by $5.5$ points. They do not help everywhere: \textsc{Planar}
stays at zero with a larger MMD ratio, and ZINC validity falls by $7.7$ points while its FCD
improves.}
\label{tab:lggm-priors}
\end{table}